\documentclass{article}

\usepackage[final]{colm2025_conference}

\usepackage{microtype}
\usepackage{newpxtext}
\usepackage{hyperref}
\usepackage{url}
\usepackage{booktabs}

\usepackage{lineno}

\definecolor{darkblue}{rgb}{0, 0, 0.5}
\hypersetup{colorlinks=true, citecolor=darkblue, linkcolor=darkblue, urlcolor=darkblue}
\definecolor{darkgreen}{rgb}{0,0.5,0}
\definecolor{darkred}{rgb}{0.7,0,0}  

\usepackage{amsmath}
\usepackage{enumitem}
\usepackage{amssymb}
\usepackage{tocloft}
\usepackage{subcaption}
\usepackage{multicol}
\usepackage{multirow}
\usepackage{pifont}  % For \ding
\usepackage{arydshln}

\newcommand{\name}{\textsc{VaPR}\xspace}

\newcommand{\cmark}{\textcolor{darkgreen}{\ding{51}}}
\newcommand{\xmark}{\textcolor{darkred}{\ding{55}}}

\usepackage{tcolorbox}

\definecolor{myboxcolor}{RGB}{245,245,245} 
\definecolor{myframe}{RGB}{0,0,128} 

\newtcolorbox{mybanner}{
  colback=myboxcolor,
  colframe=myframe,
  boxrule=1pt, % Adjust the border thickness
  left=1pt,
  right=1pt,
  top=1pt,
  bottom=1pt,
}

\newtcolorbox{mybody}{
  colback=myboxcolor,
  colframe=myframe,
  boxrule=1pt, % Adjust the border thickness
  left=1pt,
  right=1pt,
  top=1pt,
  bottom=1pt,
}

\usepackage{wrapfig}
\usepackage{xspace}
\usepackage[utf8]{inputenc} % allow utf-8 input
\usepackage[T1]{fontenc}    % use 8-bit T1 fonts

\usepackage{nicefrac}       % compact symbols for 1/2, etc.
\usepackage{xcolor}         % colors
\usepackage[normalem]{ulem}
\usepackage{pifont}
\definecolor{darkgreen}{rgb}{0,0.5,0}
\definecolor{darkblue}{rgb}{0,0,0.7}
\definecolor{darkred}{rgb}{0.7,0,0}

\title{\name – Vision-language Preference alignment for Reasoning}

\author{Rohan Wadhawan\textsuperscript{1}\thanks{Correspondance to: rohanwadhawan7@gmail.com}, Fabrice Harel-Canada\textsuperscript{1}, Zi-Yi Dou\textsuperscript{1}, Suhaila Shakiah\textsuperscript{2}, \\ \textbf{Robinson Piramuthum\textsuperscript{2}, Nanyun Peng\textsuperscript{1}} \\ \textsuperscript{1}Department of Computer Science, University of California Los Angeles, USA \\ 
\textsuperscript{2}Amazon.com, Inc., USA
}

\begin{document}

\ifcolmsubmission
\linenumbers
\fi

\maketitle

\begin{abstract}
Preference finetuning methods like Direct Preference Optimization (DPO) with AI-generated feedback have shown promise in aligning Large Vision-Language Models (LVLMs) with human preferences. However, existing techniques overlook the prevalence of noise in synthetic preference annotations in the form of stylistic and length biases. To this end, we introduce a hard-negative response generation framework based on LLM-guided response editing, that produces rejected responses with targeted errors, maintaining stylistic and length similarity to the accepted ones. Using this framework, we develop the \name dataset, comprising 30K high-quality samples, to finetune three LVLM families: LLaVA-V1.5, Qwen2VL \& Qwen2.5VL (2B-13B sizes). Our \name models deliver significant performance improvements across ten benchmarks, achieving average gains of 6.5\% (LLaVA), 4.0\% (Qwen2VL), and 1.5\% (Qwen2.5VL), with notable improvements on reasoning tasks. A scaling analysis shows that performance consistently improves with data size, with LLaVA models benefiting even at smaller scales. Moreover, \name reduces the tendency to answer "Yes" in binary questions - addressing a common failure mode in LVLMs like LLaVA. Lastly, we show that the framework generalizes to open-source LLMs as editors, with models trained on \name-OS achieving ~99\% of the performance of models trained on \name, which is synthesized using GPT-4o. Our data, models, and code can be found on the project page
\href{https://vap-r.github.io/}{https://vap-r.github.io/}
\end{abstract}    

\section{Introduction}

Recent advances in Large Vision-Language Models (LVLMs) have greatly enhanced their ability to perceive, reason, and generate open-ended responses to instructions \citep{llava-v1, llava-v15, llava-next, mplugowl2, instruct-blip, open-flamingo, cambrian, internvl-v15, internvl-v2, sharegpt4v, minigpt4, qwen2vl}. Despite the development, LVLMs often face challenges in vision-language alignment and reasoning, 
resulting in linguistically plausible texts that either contradict the visual context or lack logical reasoning. Several studies \citep{llava-rlhf, rlhfv, rlaifv, povid, csr, stic, silkie, ha-dpo, volcano, halva} have focused on improving the modality alignment capabilities of instruction-tuned LVLMs through preference fine-tuning methods, including reinforcement learning from human feedback (RLHF) \citep{instruct-gpt, ppo}, AI feedback (RLAIF) \citep{rlaif}, and direct preference optimization (DPO) \citep{dpo}.

Early preference finetuning methods for LVLMs primarily incorporated human feedback \citep{llava-rlhf, rlhfv}, but the high costs and limited scalability of this approach have led to a growing use of AI-generated feedback and synthetic preference dataset creation \citep{povid, stic, rlaifv, csr, volcano, ha-dpo}. These synthetic preference datasets have demonstrated advantages over human-annotated versions, particularly in maintaining truthfulness \citep{ppo-dpo-reco} and enhancing modality alignment \citep{povid, stic, rlaifv, csr, volcano, ha-dpo}. However, our analysis reveals that these methods often overlook inadvertently injected ``noises'', including biases in length and style \citep{ha-dpo} (see Table~\ref{tab:preference_dataset_cmp}). As DPO has been found to exploit both length \citep{length-dpo, length-bias-dpo} and stylistic biases \citep{rlhfv, ha-dpo, orpo, rlaifv}, this noise in the dataset can consequently undermine the alignment process.

\begin{figure*}[h!]
\centering
\begin{subfigure}[b]{0.44\textwidth}
    \fontsize{8.0pt}{\baselineskip}\selectfont
    \linespread{0.9}\selectfont
    \begin{mybody}
        \centering
        \includegraphics[width=0.34\textwidth,height=0.34\textwidth]{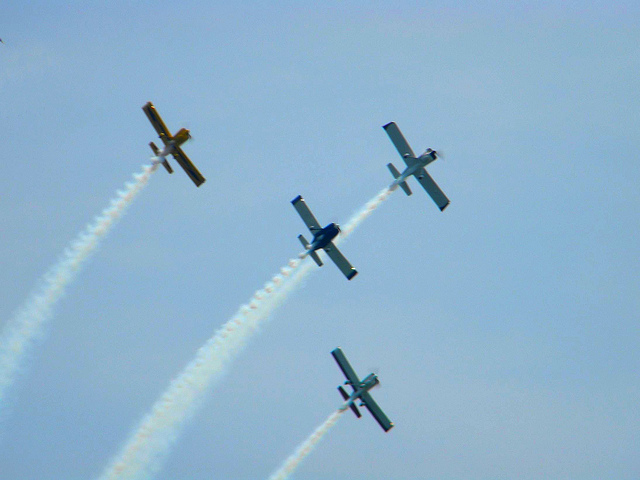}
        \\ \raggedright
        \textbf{Instruction:} How many planes are visible in the image? \\
        \textbf{Chosen Response:} 
        There are \textcolor{darkblue}{four} planes visible in the image. \\
        \textbf{\name Response:} There are \textcolor{darkgreen}{six} planes visible in the image.\\
        \textbf{Response with length bias:} 
        \textcolor{darkred}{There are seven planes visible in the sky in the image, each leaving a bright white trail behind.} \\
        \textbf{Response with style bias:} 
        \textcolor{darkred}{Five planes can be seen in the image.}
    \end{mybody}
    \label{subfig:first}
\end{subfigure}
\hfill
\begin{subfigure}[b]{0.54\textwidth}
    \fontsize{8.0pt}{\baselineskip}\selectfont
    \linespread{0.9}\selectfont
    \begin{mybody}
        \centering
        \includegraphics[width=0.23\textwidth,height=0.28\textwidth]{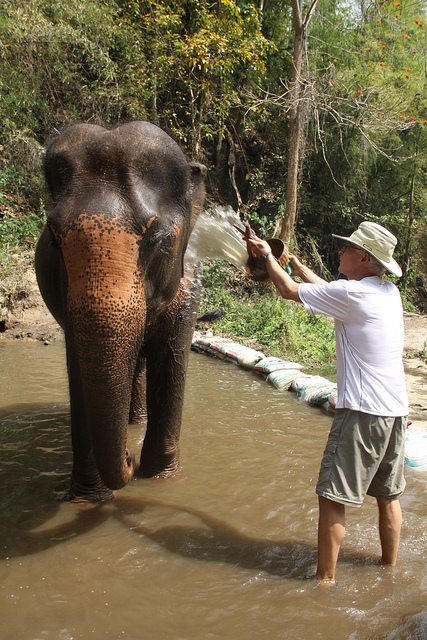}
        \\ \raggedright
        \textbf{Instruction:} Where is the man standing in relation to the baby elephant? \\
        \textbf{Chosen Response:} 
        The man is standing \textcolor{darkblue}{next to} the baby elephant \textcolor{darkblue}{in} the water. \\
        \textbf{\name Response:} The man is standing \textcolor{darkgreen}{far away from} the baby elephant, \textcolor{darkgreen}{near the edge of} the water. \\
        \textbf{Response with length bias:} 
        \textcolor{darkred}{The man is standing far from the baby elephant, attempting to throw water using a bucket while standing at the edge of the water.} \\
        \textbf{Response with style bias:} 
        \textcolor{darkred}{The elephant is standing in the water, away from the man, who is on the shore.}
    \end{mybody}
    \label{subfig:second}
\end{subfigure}
\caption{\small Examples from the \name hard-negative generation framework show instruction, image, chosen response, and three rejected variants. (a) fine-grained perception and counting capability, and (b) spatial reasoning. \name introduces targeted error - modifying only task-relevant spans - while length-biased rejections add verbose description, and style-biased ones alter style and content. \textcolor{darkblue}{Blue} highlights relevant spans in chosen response, \textcolor{darkgreen}{green} shows \name perturbations, and \textcolor{darkred}{red} indicates stylistic or length-biased edits.}
\label{fig:teaser_fig}
\end{figure*}

To improve LVLMs' alignment and reasoning, we propose a hard-negative preference data generation framework and create \name, a dataset of 30K high-quality samples derived from the LLaVA-665K SFT dataset \citep{llava-v15}. Each preference sample in \name pairs a ground truth response with a generated hard-negative response that preserves style and length while introducing targeted perturbations to ensure deliberate misalignment (see Fig.~\ref{fig:teaser_fig}). While existing approaches \citep{povid, sima, csr, rlaifv} rely on VLMs (the same or a large oracle VLM) to generate and/or score sampled responses, we frame hard-negative response generation as ground truth response editing using an LLM - implemented via constrained data generation. This editing approach leverages LLMs’ semantic understanding of text, which is typically more reliable than the image-text comprehension of VLMs \citep{hallusion-bench}. We provide the LLM with the instruction, ground truth response, and VL task-specific information (e.g., spatial reasoning, object attributes), and prompt it to perturb specific spans of the ground truth response. Unlike prior works, we use task-specific information to guide the editor in minimally editing the ground-truth response, injecting semantic errors that make the rejected response incorrect for the task while preserving style and length (see Table~\ref{tab:preference_dataset_cmp}). This prevents models from exploiting superficial cues during preference optimization (\S\ref{sec:main_results}, \& \S\ref{sec:dataset_cmp}). We also ensure that the \name corpus covers diverse instructions - perception (e.g., object existence, attribute, size, color, environment), reasoning (e.g., counting, spatial, comparison), and captioning - to help models learn generalizable preferences. For our main results, we use GPT-4o \citep{gpt-4o} as the LLM editor and refer to the approach as \name. Lastly, to evaluate the generalizability of the pipeline across different LLMs, we also conduct an ablation experiment using Qwen3-32b \citep{qwen3} as the editor, referred to as \name-OS (open-source), to distinguish it from the GPT-4o–based \name dataset.

To assess the effectiveness of the proposed framework, we fine-tune models from the LLaVA-V1.5-Instruct(7B, 13B), Qwen2VL-Instruct (2B, 7B), and Qwen2.5VL-Instruct (3B, 7B) families on the \name dataset and evaluate them across ten diverse benchmarks. These span open-ended instruction following, vision-centric reasoning, academic and mathematical tasks, hallucination detection, and adversarial robustness. \name models outperform baselines on 8 out of 10 benchmarks, with average gains of 6.5\% for LLaVA-V1.5, 4\% for Qwen2VL, and 1.5\% for Qwen2.5VL. Notably, \name yields consistent improvements on comprehensive benchmarks like SEED \citep{seedbench} and MMStar \citep{mmstar}, vision-centric reasoning benchmarks such as CV-Bench \citep{cambrian} (counting and spatial reasoning), and adversarial reasoning benchmarks like NaturalBench \citep{naturalbench}, which test compositional Visio-linguistic reasoning. Interestingly, improvements are also observed in textual and math reasoning despite no explicit training on such data. A scaling analysis reveals that performance scales with dataset size, with LLaVA-v1.5 models benefiting even with small preference tuning data budget, and Qwen2VL \& Qwen2.5VL variants improving at larger scales.

We further compare \name to other preference datasets and empirically analyze how they function under DPO optimization. Our findings highlight the influence of stylistic and length differences in preference data, which can lead to spurious learning signals during training. \name mitigates this issue by generating hard-negative rejected responses, which are stylistically and length-wise similar to the chosen ones. We also observe that \name reduces the tendency to answer "Yes" in binary questions, addressing a common issue in LVLMs like LLaVA \citep{pope, gavie, hallusion-bench, naturalbench}. Lastly, our ablation using Qwen3-32b to generate \name-OS show that open-source models can follow the same prompting strategy, produce targeted perturbations, and yield models that outperform the base instruct model and achieve ~99\% performance of the \name models.

The contributions of this work are fourfold:
\begin{itemize}
\item We propose \name, a hard-negative generation framework based on LLM-guided response editing that constructs synthetic preference data with reduced stylistic and length biases. Using this framework, we create a 30K sample high-quality dataset.
\item We fine-tune LLaVA-V1.5, Qwen2VL, and Qwen2.5VL on \name and evaluate them across ten benchmarks. \name outperforms SFT and preference-tuned baselines on 8 out of 10 benchmarks, with average gains of 6.5\% (LLaVA), 4.0\% (Qwen2VL), and 1.5\% (Qwen2.5VL). Scaling analysis shows LLaVA benefits from smaller-scale data, while Qwen models require larger-scale data to improve.
\item Our analysis shows that \name enhances performance on reasoning tasks like adversarial, spatial, counting, \& textual, reduces the overuse of "Yes" in binary questions, and avoids spurious learning signals by mitigating stylistic and length biases present in prior preference datasets.

\item Our ablation with Qwen3-32b as the LLM-editor, demonstrates that the framework generalizes well to open-source LLMs, producing similarly effective hard negatives (\name-OS) and preference-tuned models achieving ~99\% of the performance of models trained on \name. Our data, models, and code can be found on the project page \href{https://vap-r.github.io/}{https://vap-r.github.io/}
\end{itemize}

\section{Preliminaries}

%\subsection{Large Vision Language Models}
%\zd{i think you don't have to introduce LVLMs. may change the subsection name to ``Notations''. or just remove the subsection name. }

Large Vision Language Models (LVLMs) enhance the capabilities of Large Language Models (LLMs) by adapting them to multimodal tasks. This enables the model to predict the probability distribution for the next token in a sequence given multimodal inputs. Specifically, given an input pair $x$ $\langle x_v, x_t \rangle$ comprising an image $x_v$ and instruction text $x_t$, the LVLM generates a text response $y$. Preference optimization has emerged as a promising technique for fine-tuning language models to align their outputs with desired outcomes, which we briefly overview in this section. 

\subsection{Direct Preference Optimization}
%\zd{instead of just introducing DPO, the more important part may be pointing out the importance of the preference dataset.}
Given a prompt $x$  $\langle x_v, x_t \rangle$, a large vision language model governed by policy $\pi_\theta$ can yield a conditional distribution $\pi_\theta(y \mid x)$, where $y$ is the generated text response. To train using preference data, we define the dataset $\mathcal{D} = \{(x^{(i)}, y_\text{w}^{(i)}, y_\text{l}^{(i)})\}_{i=1}^N$, where $y_\text{w}^{(i)}$ and $y_\text{l}^{(i)}$ represent the more and less preferred responses, in our case, chosen and hard-negative rejected response, respectively, for a given input $x^{(i)}$. Preference optimization leverages this dataset to fine-tune models effectively.

\paragraph{Objective} Direct Preference Optimization (DPO) calculates the probability of preferring $y_\text{w}$ over $y_\text{l}$ as:
\[
p(y_\text{w} \succ y_\text{l}) = \sigma(r(x, y_\text{w}) - r(x, y_\text{l})),
\]
where $\sigma(\cdot)$ is the sigmoid function. The loss function for DPO can be written as follows: \\
\begin{multline}
\mathcal{L}_{\text{DPO}}(\pi_\theta; \pi_\text{ref}) = 
-\mathbb{E}_{(x, y_\text{w}, y_\text{l}) \sim \mathcal{D}} \left[ \log \sigma \left( \alpha \log \frac{\pi_\theta(y_\text{w} \mid x)}{\pi_\text{ref}(y_\text{w} \mid x)} \right. \right. \left. \left. - \alpha \log \frac{\pi_\theta(y_\text{l} \mid x)}{\pi_\text{ref}(y_\text{l} \mid x)} \right) \right],
\label{eq:dpo_loss}
\end{multline}

where $\pi_\text{ref}(y \mid x)$ represents the reference policy, typically the model after SFT.

\paragraph{Preference Dataset} There are different ways to construct the preference dataset. For example, \citet{llava-rlhf} employs human annotations, which are of high quality but is difficult to scale. On the other hand, AI annotation with LVLMs like GPT-4V \citep{gpt-4v} scales better but risks inconsistencies and hallucinations \citep{ha-dpo, orpo}. Our focus is to construct a scalable framework for preference dataset construction while ensuring its quality.

\section{\name Dataset Construction}
\label{sec:dataset_construction}
\begin{figure*}[!h]
    \centering
    \includegraphics[scale=0.36]{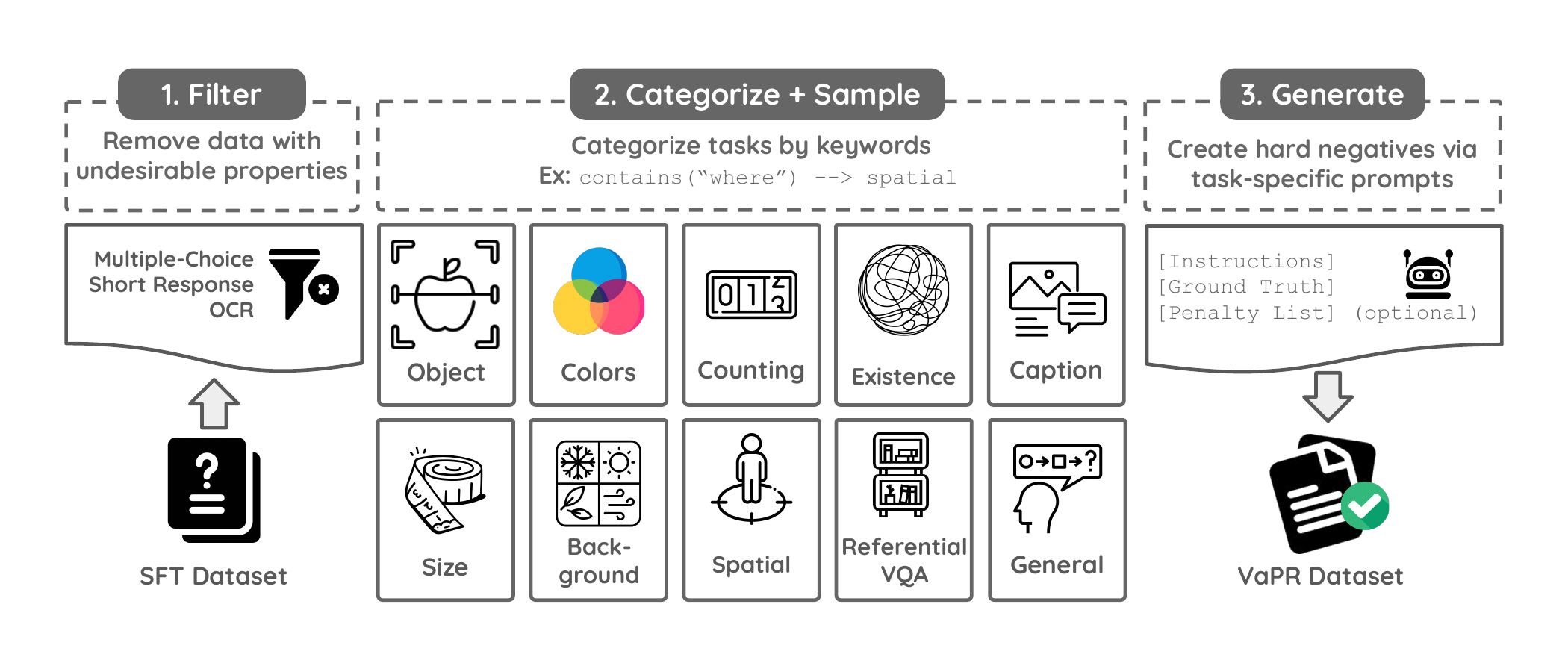}
    \caption{\small \name: A three-stage pipeline that generates 30K hard-negative preference pairs from LLaVA-v1.5-665K SFT~\citep{llava-v15}. \textbf{Stage 1:} Filter out irrelevant samples (e.g., MCQs). \textbf{Stage 2:} Categorize remaining samples based on task. \textbf{Stage 3:} Use task-specific prompts (with optional penalty lists) to produce stylistically and length-wise similar but content-distinct negative responses.}
    \label{fig:dataset_pipeline}
\end{figure*}

%\zd{need to connect to the previous section, i.e., say that you are constructing the preference dataset for DPO first.}

We detail the design principles and construction processes of our preference dataset. Our main idea is to construct hard negatives that are stylistically and length-wise similar to the ground-truth samples.

\subsection{Design Principles}
We adhered to two key principles for constructing the \name dataset:

\paragraph{Task Diversity} The dataset ensures a near-balanced distribution across a wide range of tasks (see Fig.~\ref{fig:dataset_piechart}). It consists of foundational vision-language capabilities, including perceptual (e.g., object recognition, object attribute recognition - color, size, etc., and background understanding) and reasoning (e.g., spatial, counting, comparative). We also incorporate tasks combining perception and reasoning like captioning, world knowledge reasoning, and referential VQA (eg. region-level perception or reasoning). This diversity is designed to improve the model's comprehensive capabilities.

\paragraph{Hard Negative Rejected Responses} Inspired by the role of hard negatives in enhancing vision-language compositionality in image-text pretraining \citep{openai-clip, neg-clip, sugarcrepe}, we generate hard-negative responses. These are synthesized by introducing targeted perturbations to high-quality SFT ground truth responses while maintaining stylistic and structural similarity (e.g., length), thereby resulting in incorrect responses given a task.

\subsection{Sourcing \& Processing}

\paragraph{Sourcing} We build preference samples using the high-quality LLaVA-665K SFT dataset~\citep{llava-v15}, which has broad task coverage. 

\paragraph{Filtering} The data undergoes careful processing to filter tasks unsuitable for hard-negative generation. We exclude tasks that lack sufficient training signals to address vision-language misalignments, such as text-only tasks and simple response types (e.g., MCQ or bounding box prediction). However, we retain "Yes/No" response instruction on the existence of objects, attributes, and reasoning (e.g., count, spatial, comparative), as they correspond to the key vision-language capabilities, and convert them to extended responses to better suit preference optimization. We also filter out OCR instructions, as prior works emphasize the need for larger input image resolutions and fine-grained visual perception as driving factors in improving OCR performance \citep{contextual, monkey, texthawk2}.

\paragraph{Categorization} We categorize the filtered corpus into ten task categories using task-specific keywords (see Fig.~\ref{fig:dataset_pipeline}). We subsample a portion of the SFT dataset from each category for hard-negative response generation. Notably, for binary response instructions ("Yes"/"No"), we enforce an equal distribution of "Yes" and "No" responses to mitigate bias towards "Yes," which arises from the predominance of affirmative instructions in the original SFT dataset \citep{pope, hallusion-bench, gavie}. See Appendix~\S\ref{sec:datagen_categories_appendix} for details.

\subsection{Generation Pipeline}
The \name framework generates hard-negative responses by editing ground truth (instruction, response) pairs from our filtered \& categorized SFT subset using GPT-4o. This process is guided by two key components: conditioning information and perturbation diversity.

\paragraph{Conditioning Information}
 Task-specific prompts (e.g., attributes, spatial reasoning, counting) guide edits to ensure responses remain fluent yet incorrect (see Fig.~\ref{fig:dataset_pipeline}). These were preferred over general prompts, which often introduced irrelevant changes given the task or added verbosity or style changes, contributing to noise in the dataset.

\paragraph{Diversity of Perturbations}
 Most categories (e.g., object type, size, existence) produced diverse outputs with zero-shot prompting, while for others (e.g., color, counting, captioning), we introduced a penalty list, periodically updating it with commonly used perturbation values to prevent reuse. In captioning, we first extracted dimensions (e.g., color, spatial relation) and applied random perturbations across them, using a penalty list to encourage a variety of dimensions and perturbation values. See Appendix~\S\ref{sec:datagen_prompts_appendix} for details.

\begin{figure}[!ht]
    \centering
    \begin{minipage}[c]{0.45\textwidth}
        \centering
        \includegraphics[width=\linewidth]{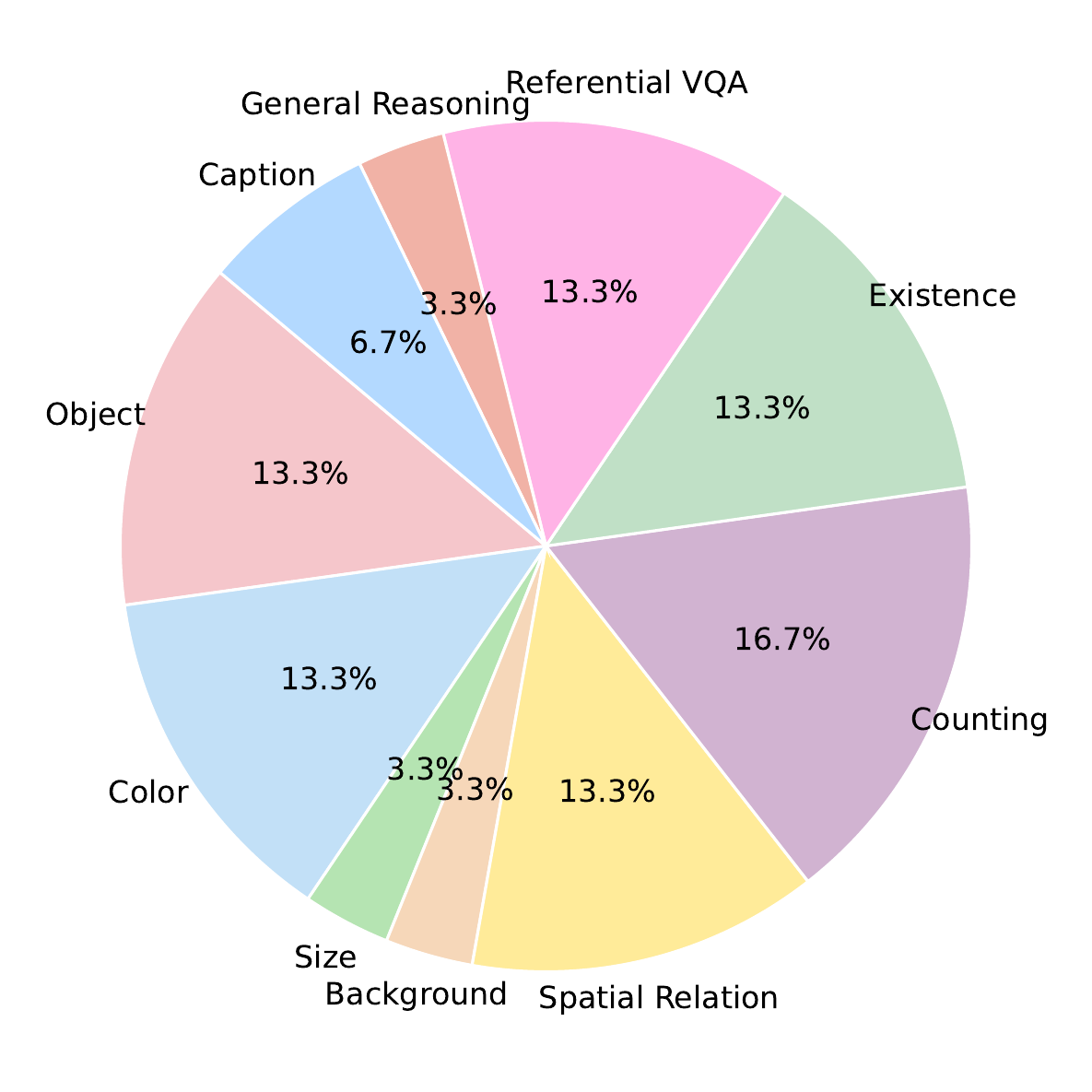}
        \caption{\small \name task distribution}
        \label{fig:dataset_piechart}
    \end{minipage}%
    \hfill
    \begin{minipage}[c]{0.53\textwidth}
        \centering
        \captionof{table}{\small Comparison of the \name Preference Dataset with Related Works: We compare \name to HA-DPO~\citep{ha-dpo}, POVID~\citep{povid}, RLAIF-V~\citep{rlaifv}, and CSR~\citep{csr} based on stylistic and length analysis, using linguistic similarity via word-level Levenshtein distance (LD) and token-level sequence length differences, respectively. We report the percentage of samples where chosen responses are longer \textcolor{darkgreen}{$(chosen>rejected)$} or shorter \textcolor{darkred}{$(rejected>chosen)$}, with corresponding token differences following the same color notation. A lower LD stands for higher stylistic similarity, and lower token-level sequence length differences stand for higher length similarity.}
        \label{tab:preference_dataset_cmp}
        \setlength{\tabcolsep}{0.8mm}
        \resizebox{\linewidth}{!}{
        \begin{tabular}{lccccc}
        \hline
         & \textbf{Ours} & HA-DPO & POVID & RLAIF-V & CSR \\\hline
        Overall Samples (\%) & (\textcolor{darkgreen}{21}, \textcolor{darkred}{79}) & (\textcolor{darkgreen}{41}, \textcolor{darkred}{59}) & (\textcolor{darkgreen}{24}, \textcolor{darkred}{76}) & (\textcolor{darkgreen}{45}, \textcolor{darkred}{55}) & (\textcolor{darkgreen}{38}, \textcolor{darkred}{62}) \\
        - Linguistic Similarity & 6 & 49 & 30 & 62 & 97 \\
        - Avg. Token Length Difference & 3 & 18 & 16 & 15 & 27 \\
        - Token Length Difference & (\textcolor{darkgreen}{10}, \textcolor{darkred}{1}) & (\textcolor{darkgreen}{15}, \textcolor{darkred}{20}) & (\textcolor{darkgreen}{27}, \textcolor{darkred}{13}) & (\textcolor{darkgreen}{17}, \textcolor{darkred}{14}) & (\textcolor{darkgreen}{23}, \textcolor{darkred}{29}) \\
        \hline
        \end{tabular}
        }
    \end{minipage}
\end{figure}

\subsection{Dataset Analysis}
\label{sec:dataset_analysis}

We perform statistical analysis of the \name dataset, with task-wise distribution shown in Fig.~\ref{fig:dataset_piechart}. To assess annotation quality, we conducted a human evaluation on 500 stratified samples across task categories, finding 97\% alignment with hard-negative criteria and 86\% inter-annotator agreement (IAA) using Fleiss' kappa (see Appendix~\S\ref{sec:datagen_human_study_appendix}).

To quantify stylistic and length similarity, we compare \name against prior datasets - HA-DPO~\citep{ha-dpo}, POVID~\citep{povid}, RLAIF-V~\citep{rlaifv}, and CSR~\citep{csr}. Linguistic similarity is measured via word-level Levenshtein distance, where lower values indicate targeted content edits; higher values reflect broader stylistic variation. Length similarity is computed as token-level sequence length differences. We also report the proportion of samples where chosen or rejected responses are longer, for both short-form (e.g., VQA) and long-form (e.g., captioning) tasks (see Appendix~\S\ref{sec:fine_grained_data_cmp_appendix}). Table~\ref{tab:preference_dataset_cmp} shows that \name has the lowest Levenshtein distance and token length difference, reinforcing the hard-negative nature of its rejections. In Section~\S\ref{sec:dataset_cmp}, we show that larger stylistic and length discrepancies in other datasets lead to premature saturation in reward accuracy, suggesting reward hacking behavior that is mitigated with \name.

\section{Experiments}
\subsection{Experimental Setup}
\subsubsection{Models \& Baselines}
We preference-tune LLaVA-1.5~\citep{llava-v15} (7B and 13B), Qwen2VL-Instruct~\citep{qwen2vl} (2B and 7B), and Qwen2.5VL-Instruct (3B and 7B) using Direct Preference Optimization (DPO)~\citep{dpo}. LoRA finetuning~\citep{lora} is employed during the preference learning phase. Models preference-tuned on the \name dataset, which comprises 30K samples, are referred to as LLaVA-\name, Qwen2VL-\name, and Qwen2.5VL-\name, respectively. Additional details on the training setup are provided in Appendix~\S\ref{sec:training_setup_appendix}.

We compare the performance of the \name models with two baselines: base instruct models and Supervised Finetuned (SFT) variants of base instruct models on the \name dataset. We also compare the \name models with prior works involving different preference dataset generation techniques - Human (LLaVA-RLHF~\citep{llava-rlhf}), AI annotations using a large closed-source model (HA-DPO~\citep{ha-dpo}, POVID~\citep{povid}), self-generated AI feedback (SIMA ~\citep{sima}, RLAIF-V~\citep{rlaifv}, and CSR~\citep{csr}) - using models made publicly available by the respective works. 

\begin{table*}[t]
\caption{
\small Performance comparison of LLaVA-v1.5-Instruct, Qwen2VL-Instruct, Qwen2.5VL-Instruct, SFT and DPO models finetuned on \name, and other preference datasets across 2B, 3B, 7B, and 13B parameter sizes on ten benchmarks. Higher scores indicate better performance across all benchmarks, with the highest score for each benchmark highlighted in \textbf{bold}. All models are evaluated using publicly available checkpoints, adhering to evaluation parameters prescribed by the benchmarks. $^{\dagger}$ represents results that show statistically significant improvement via bootstrap resampling (95\% CI).
}
\label{tab:main_results}
\centering
\setlength{\tabcolsep}{4pt}
\scriptsize
\begin{tabular}{llcccccccccccc}
\toprule
\textsc{Row} & \textsc{Method} & \textsc{LLaVA\textsuperscript{W}} & \textsc{ConT} & \textsc{MMV} & \textsc{SEED\textsuperscript{I}} & \textsc{CV} & \textsc{MV} & \textsc{MMMU} & \textsc{MMS} & \textsc{POPE} & \textsc{NB}
 \\
\midrule 
1 & LLaVA-1.5-7B & 64.8 & 16.8 & 30.9 & 66.2 & 62.1 & 30.1 & 35.4 & 32.6 & 85.9 & 12.7  \\
2 & + Fact-RLHF & 58.7 & 10.5 & 30.2 & 55.7 & 35.1 & 26.4 & 28.0 & 30.7 & 79.8 & 7.8 \\
3 & + HA-DPO & 69.8 & 14.2 & 30.9 & 64.6 & 62.0 & 29.7 & 35.5 & 32.9 & 83.5 & 13.5  \\
4 & + POVID & 70.1 & 18.6 & 31.8 & 66.3 & 62.3 &  30.6 & 35.6 & 33.3 & 86.0 & 13.1  \\
5 & + SIMA & 66.4 & 18.0 & 31.4 & 66.1 & 60.1 & 30.1 & 35.6 & 32.5 & 85.8 & 12.6  \\
6 & + RLAIF-V & 73.1 & 14.1 & 29.3 & 65.8 & 59.8 &  30.3 & 34.2 & 33.7 & 78.9 & 12.9  \\ 
7 & + CSR  & 69.5 & 15.8 & 30.6 & 65.9 & 61.1 &  30.5 & 35.6 & 32.2 & \textbf{86.2} & 13.4  \\
8 & + \name SFT (ours) & 64.0 & 16.0 & 30.0 & 65.6 & 59.5 &  29.5  & 34.9 & 31.7 & 83.1 & 12.2  \\
9 & + \name DPO (ours) & \textbf{76.2$^{\dagger}$} & \textbf{20.6$^{\dagger}$} & \textbf{32.9} & \textbf{66.7$^{\dagger}$} & \textbf{62.9$^{\dagger}$} & \textbf{30.8} &  \textbf{35.7} & \textbf{34.7$^{\dagger}$} & 85.4 & \textbf{14.5$^{\dagger}$}\\
\midrule
10 & LLaVA-1.5-13B & 72.3 & 18.6 & 36.7 & 68.2 & 62.5  & 30.7 & \textbf{36.1} & 33.8 & 86.0 & 14.9\\
11 & + Fact-RLHF & 70.4 & 15.6 & 37.0 & 61.1 & 53.7 & 31.2 & 28.0 & 32.1 & 81.7 & 12.4  \\
12 & + SIMA & 70.7 & 19.2 & 36.0 & 68.0 & 62.6 & 30.8  & 35.4 & 34.0 & 86.0 & 14.6  \\
13 & + CSR & 74.2 & 17.8 & 35.6 & 68.2 & 62.4&  31.3 & 35.9 & 33.9 &  \textbf{86.8$^{\dagger}$} & 14.9  \\ 
14 & + \name SFT (ours) & 71.4 & 17.4 & 34.7 & 67.2 & 61.5 
 & 30.4  & 34.5 & 33.3 & 83.9 & 13.7  \\
15 & + \name DPO (ours) & \textbf{80.5$^{\dagger}$} & \textbf{21.2} & \textbf{37.3} & \textbf{68.7$^{\dagger}$} & \textbf{64.6$^{\dagger}$} & \textbf{32.3$^{\dagger}$}  & 35.8 & \textbf{35.6$^{\dagger}$} & 86.3 & \textbf{18.2$^{\dagger}$}  \\

\midrule
16 & Qwen2VL-2B & 83.2 & 27.7 & 53.3 & 73.6 & 66.5 &  \textbf{51.0} & 38.7 & 43.4  & 86.5 & 24.3\\
17 & + \name SFT (ours)& 69.1 & 22.1 & 43.3 & 70.3 & 54.5 & 46.4  & 36.6 & 41.5 & 85.5 & 15.1  \\
18 & + \name DPO (ours)& \textbf{88.1} & \textbf{34.8$^{\dagger}$} & \textbf{54.1} & \textbf{74.0$^{\dagger}$} & \textbf{69.0$^{\dagger}$} & 50.9 & \textbf{39.2} & \textbf{43.7} & \textbf{88.3$^{\dagger}$} & \textbf{25.7$^{\dagger}$} \\

\midrule
19 & Qwen2VL-7B & 92.5 & 39.7 & 62.1 & 76.4 & 75.7 & 57.5 & \textbf{50.7} & 56.7 & \textbf{87.3} & 30.8\\
20 & + \name SFT (ours) & 79.6 & 36.6 & 52.6 & 73.9 & 67.4 & 56.3  & 47.7 & 52.9 & 86.2 & 23.7  \\
21 & + \name DPO  (ours) & \textbf{96.2} & \textbf{43.9$^{\dagger}$} & \textbf{65.4} & \textbf{76.8$^{\dagger}$} & \textbf{76.3$^{\dagger}$} & \textbf{58.2}  & 50.0 & \textbf{57.8$^{\dagger}$} & \textbf{87.3} & \textbf{32.5$^{\dagger}$}  \\

\midrule
22 & Qwen2.5VL-3B & 98.1 & 37.2 & 67.3 & 75.0 & 71.5  & 52.5  & \textbf{45.7} & 54.7 & 86.3 & 25.4  \\
23 & + \name SFT (ours) & 79.8 & 33.4 & 50.1 & 70.9 & 60.5 & 48.6  & 43.1 & 48.5 & 84.8 & 21.2   \\
24 & + \name DPO (ours) & 97.1 & \textbf{40.3} & \textbf{67.4} & \textbf{75.5$^{\dagger}$} & \textbf{72.7$^{\dagger}$}  & \textbf{53.4} & 44.9  &  \textbf{56.1$^{\dagger}$} & \textbf{86.4} & \textbf{26.3$^{\dagger}$}  \\

\midrule
25 & Qwen2.5VL-7B & 101.4 & 53.3 & 71.0 & 77.7 & 80.1 & 58.6  & \textbf{50.9} & 61.9 & 86.3 & 32.0  \\
26 & + \name SFT (ours) & 80.2 & 38.9 & 59.0 & 74.9 & 69.4 & 56.8 & 48.1  & 56.1 & 82.8 & 23.8 \\
27 & + \name DPO (ours) & \textbf{101.5} & \textbf{53.4} & \textbf{72.4} & \textbf{77.8} & \textbf{81.1$^{\dagger}$} & \textbf{59.8$^{\dagger}$}  & 50.6 & \textbf{62.5} & \textbf{86.9$^{\dagger}$} & \textbf{32.8$^{\dagger}$} \\

\bottomrule
\end{tabular}
\end{table*}

\subsubsection{Evaluation Benchmarks} We evaluate \name models across ten benchmarks covering diverse skills:  Open-ended QA - LLaVA-in-the-wild (LLaVA\textsuperscript{W})~\citep{llava-v1}, ConTextual (ConT)~\citep{contextual}, \& MM-VET (MMV)~\citep{mm-vet};  Comprehensive vision benchmarks (reasoning \& perception) - SEED Bench (SEED\textsuperscript{I} - image split)~\citep{seedbench} \& MMStar (MMS)~\citep{mmstar}; , Vision-centric reasoning (spatial reasoning, counting) - CV Bench (CV)~\citep{cambrian}; Hallucination \& Adversarial reasoning - Pope~\citep{pope}, NaturalBench (NB)~\citep{naturalbench}; Academic \& Math Reasoning - MathVista (MV)~\citep{mathvista} \& MMMU~\citep{mmmu} (benchmark details provided in Appendix~\S\ref{sec:evaluation_setup_appendix}). 

\subsection{Results}
%\subsubsection{Main Results}
%\zd{may remove the subsubsection name if there's only one}
\label{sec:main_results}

Table~\ref{tab:main_results} presents the core experimental findings. \name models outperform baseline and prior preference-tuned models on 8 out of 10 benchmarks and remain competitive on the rest. LLaVA-\name 7B and 13B achieve average gains of 7\% and 6\%, respectively, while Qwen2VL-\name 2B and 7B improve by 5\% and 3\%. Despite Qwen2.5VL being a strong baseline, preference finetuning on \name yields improvements of 2\% (3B) and 1\% (7B), with gains concentrated in vision-centric benchmarks (CV-Bench, MMStar, SeedBench), and the adversarial benchmark (NaturalBench).

These improvements in vision-centric and adversarial benchmarks are consistent across all \name models, indicating that preference finetuning on \name strengthens visio-linguistic compositionality - particularly in perception (e.g., fine-grained object-attribute identification) and reasoning tasks such as spatial relationships and counting (see Appendix~\S\ref{sec:benchmark_detailed_results}). Notably, improvements also extend to textual and mathematical reasoning benchmarks (ConTextual and MathVista), despite not being explicitly trained on OCR, textual reasoning, or math tasks, VaPR models achieve strong performance in these areas. We attribute this to their enhanced fine-grained perception, spatial reasoning, and counting capabilities, which is consistent with prior work \citep{ocrbench2}. In contrast, all models show limited or no improvement on MMMU, aligning with prior findings \citep{unpacking} that preference optimization primarily enhances alignment and truthfulness, rather than factuality.

Lastly, we observe that While \name demonstrates strong effectiveness under DPO optimization, supervised finetuning (SFT) on the same dataset does not yield comparable gains. Specifically, SFT results in gentle performance degradation for LLaVA, and more for the Qwen2VL and Qwen2.5VL families. This outcome suggests that SFT on a relatively small dataset like \name - particularly in comparison to the original large-scale SFT corpora - may lead to overfitting, especially in models with strong pretrained priors like Qwen2.5VL, thereby limiting their generalization capabilities. In contrast, preference optimization with carefully constructed, length and stylistically similar hard-negative pairs enables more robust and generalizable representation learning. These findings indicate that the gains observed in \name models stem not from the selection of samples, but from the preference signal induced by \name preference pairs.

\subsection{Analysis}

\subsubsection{Preference Dataset Comparison} \label{sec:dataset_cmp}

We compare \name with two alternatives: (1) \textbf{POVID}, which uses GPT-4V to generate rejected responses given an SFT sample, and (2) \textbf{SIMA}, a self-preference method that composes preference pairs from greedy and sampled outputs generated and critiqued by the same LVLM to be preference tuned. \name consistently outperforms both LLaVA and Qwen2VL families, showing 4-6\% average gains over POVID and SIMA.

\begin{figure}[t]
  \centering
  \begin{subfigure}[c]{0.48\textwidth}
    \centering
    \includegraphics[width=\textwidth]{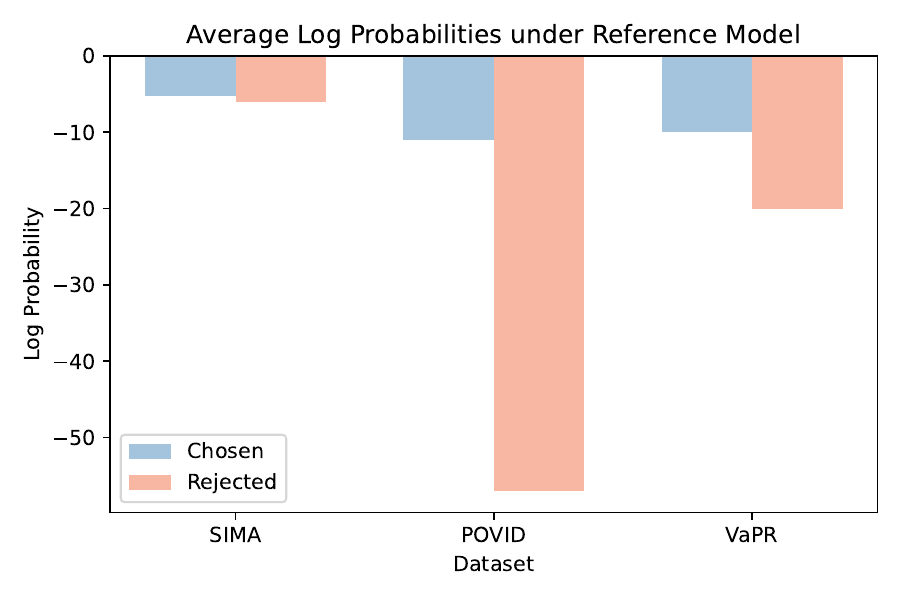}
    \caption{Reference model log-probabilities}
    \label{fig:ref_logp}
  \end{subfigure}
  \hfill
  \begin{subfigure}[c]{0.48\textwidth}
    \centering
    \includegraphics[width=\textwidth]{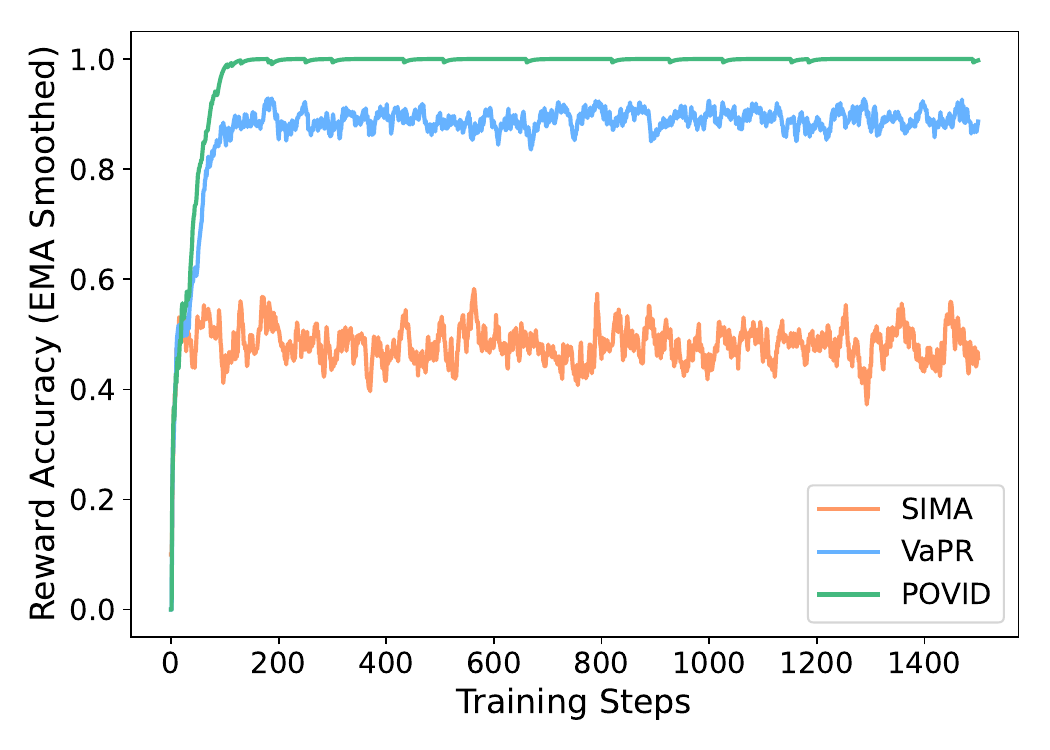}
    \caption{Reward accuracy trends}
    \label{fig:reward_acc}
  \end{subfigure}
  \caption{\small Comparison of preference datasets. (a) Average reference model log-probabilities for chosen vs. rejected responses across \name, SIMA, and POVID - lower values indicate lower reference likelihood. (b) Reward accuracy trends over training steps show that SIMA improves gradually while POVID saturates quickly.}
  \label{fig:logp_and_reward}
\end{figure}

To analyze how the different datasets affect the DPO optimization process, let us rewrite the loss:\(
\mathcal{L}_{\text{DPO}} = -\log \sigma\left( \alpha (\Delta_\theta - \Delta_{\text{ref}}) \right)
\), where \( \Delta_\theta = \log \pi_\theta(y_w \mid x) - \log \pi_\theta(y_l \mid x) \) and \( \Delta_{\text{ref}} = \log \pi_{\text{ref}}(y_w \mid x) - \log \pi_{\text{ref}}(y_l \mid x) \). We observe that POVID exhibits higher \( \Delta_{\text{ref}} \) than \name despite similar chosen log-probabilities (see Fig.~\ref{fig:ref_logp}), indicating that its rejected responses are less likely under the reference model most likely due to greater stylistic and length differences (see Table~\ref{tab:preference_dataset_cmp}). We further observe that the POVID model rapidly attains a reward accuracy of 1 (see Fig.~\ref{fig:reward_acc}), suggesting that the model may be overfitting based on preference signals derived from length and stylistic differences, unlike \name models, which improve more gradually without saturating, indicating reduced overfitting due to exposure to more challenging preference pairs.

SIMA, on the other hand, shows \(\Delta_{ref} \approx 0\) (see Fig.~\ref{fig:ref_logp}), with chosen and rejected responses often near-identical - including ~20\% duplicates. In such cases, the loss depends entirely on \(\Delta_\theta\), removing reference-guided regularization and amplifying noisy or weak signals. This leads to poor reward accuracy (~50\%) and degraded generalization (see Fig.~\ref{fig:logp_and_reward}). While self-preference methods like CSR and RLAIF-V can mitigate some issues via multi-step scoring, like POVID, they can still reward hack due to length and stylistic differences (see Table~\ref{tab:preference_dataset_cmp}). Detailed methodology, results \& analysis is provided in Appendix~\S\ref{sec:dataset_cmp_appendix}.

\name addresses both issues by explicitly generating hard negatives that are stylistically and length-wise similar to positives, ensuring DPO learns from content differences. In low-resource settings where high-quality SFT data is scarce, \name can complement self-preference methods by using confident responses from approaches like CSR or RLAIF-V as positives and generating hard negatives via the VaPR pipeline, allowing the curation of high-quality preference data in an unsupervised manner - a promising future direction

\subsubsection{Scaling analysis}
We investigate the impact of dataset size on model performance by conducting a data scaling analysis using three training budgets: 3K, 10K, and 30K samples from the \name dataset. Detailed results are provided in Appendix~\S\ref{sec:data_scale_appendix}. As shown in Fig.~\ref{fig:data_scale}, all \name models exhibit improved performance with increasing data. Interestingly, LLaVA-\name models achieve substantial gains even at the lowest data budget (3K), with diminishing returns at higher scales. In contrast, Qwen2VL and Qwen2.5VL models- being stronger base instruct models- show more modest improvements at 3K but benefit more as dataset size increases. This trend aligns with their stronger pretrained priors compared to LLaVA-v1.5, which may require more supervision to meaningfully shift under preference tuning.

\subsubsection{\name models do not overconfidently say "Yes" to  "Yes/No" questions}
\label{sec:yn_analysis}

\begin{figure}[t]
  \centering
  \begin{minipage}[c]{0.32\textwidth}
    \centering
    \includegraphics[scale=0.34]{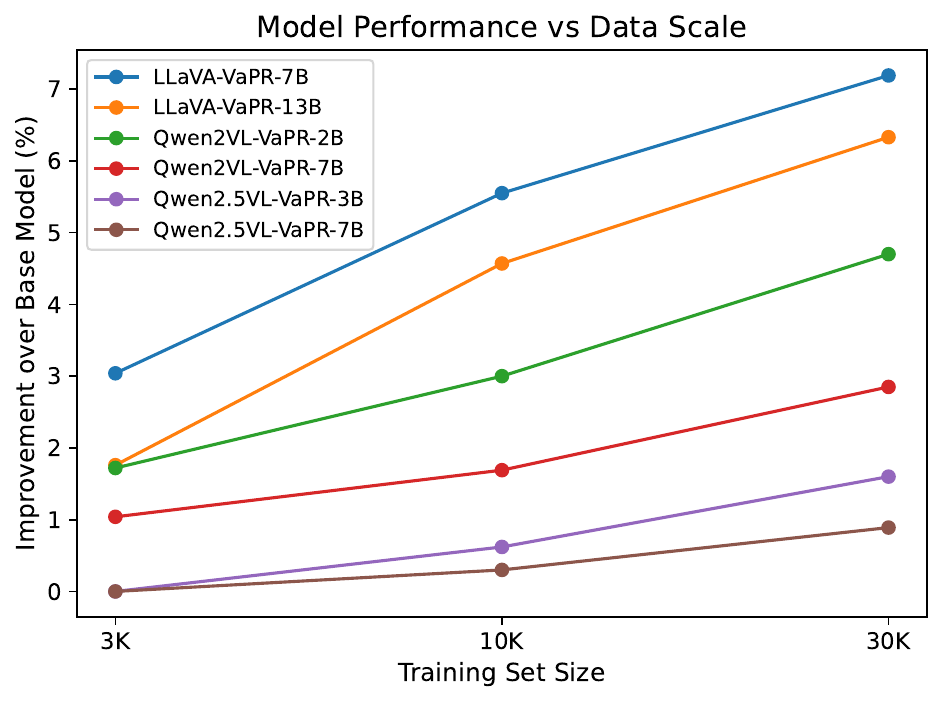} % scaled down
    \caption{\small Performance scaling of \name models with 3K, 10K, and 30K samples, shown as \% improvement over base instruct models. Note: X-axis spacing between 3K, 10K, and 30K is not uniform.}
    \label{fig:data_scale}
  \end{minipage}%
  \hfill
  \begin{minipage}[c]{0.6\textwidth}
    \centering
    \includegraphics[scale=0.23]{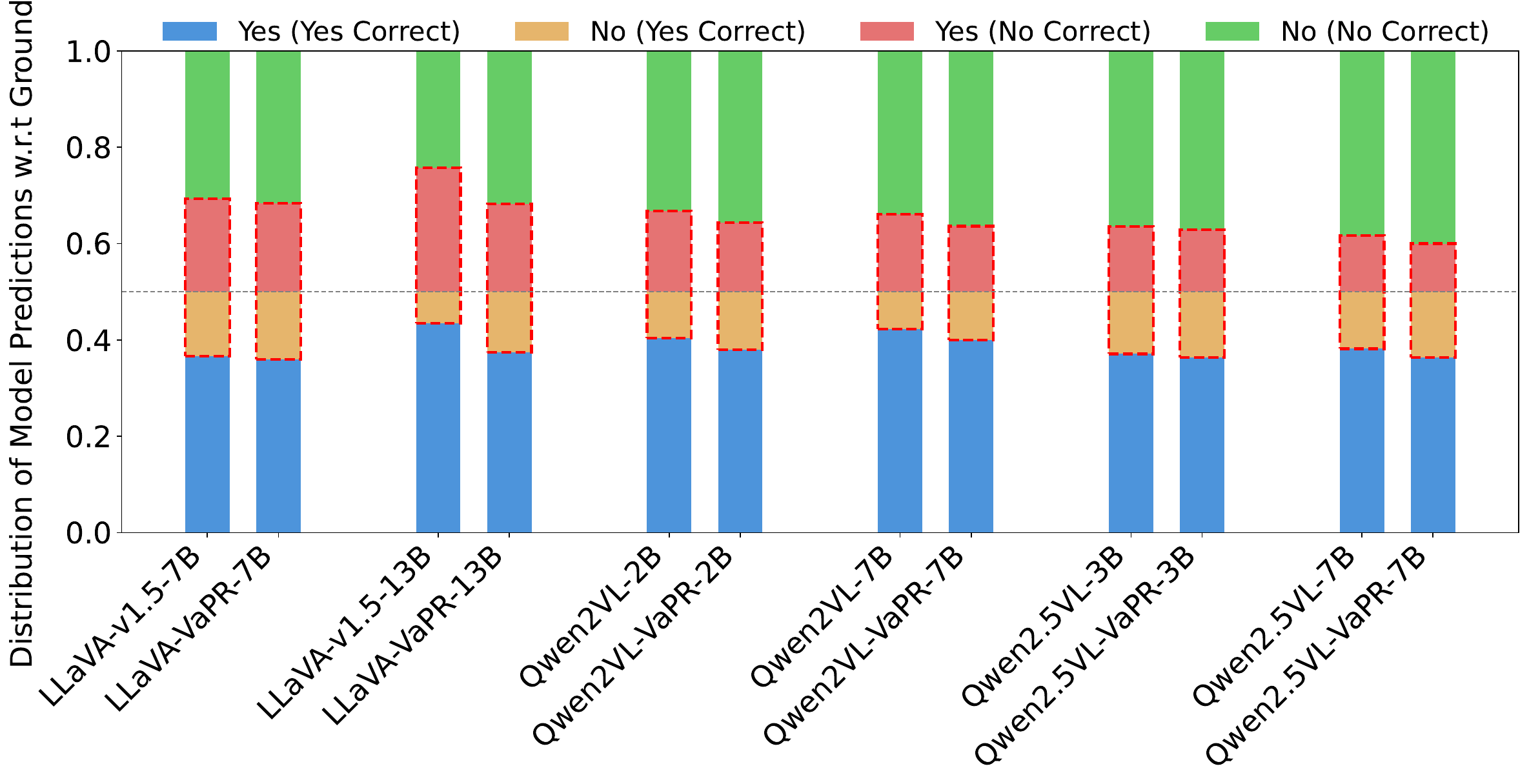} % scaled down
    \caption{\small Comparison of "Yes"/"No" responses on NaturalBench - base vs. \name models. Dotted rectangles highlight incorrect "Yes" (red) and "No" (yellow) predictions, illustrating reduced "Yes" bias after preference finetuning.}
    \label{fig:yes_no_analysis}
  \end{minipage}
\end{figure}

In this section, we examine the response patterns of large vision-language models (LVLMs), which tend to answer "Yes" more frequently than "No" for binary "Yes/No" questions \citep{pope, naturalbench, gavie, hallusion-bench}. Specifically, we analyze model outputs on NaturalBench, an adversarial benchmark comprising paired instructions and paired images. As shown in Fig.~\ref{fig:yes_no_analysis}, base SFT variants of LLaVA-v1.5, Qwen2VL, and Qwen2.5VL exhibit a clear tendency to favor "Yes" responses, even for questions where the correct answer is "No" (highlighted in red). However, the \name models demonstrate a notable reduction in this bias, with an emergent shift towards answering "No" more frequently than "Yes." This behavior is most pronounced in LLaVA-\name 13B. We attribute these improvements to enhanced perception and reasoning capabilities, which manifest as improved visio-linguistic compositionality and a reduced bias towards "Yes" responses. 

\subsubsection{\name-OS: Ablation Study using open-source LLM Editor}
\label{sec:os_analysis}

We conduct an ablation using Qwen3-32B as the open-source LLM editor to evaluate the generalizability of the VaPR pipeline beyond closed models like GPT-4o (see Appendix~\S\ref{sec:os_analysis_appendix}). The resulting dataset, \name-OS, is constructed from the same subset as the 10K version of \name. Analyzing the data, we find that \name-OS exhibits comparable hard-negative characteristics to VaPR, with a token length gap of 6 (vs. 3 in VaPR) and a Levenshtein distance of 10 (vs. 6). We fine-tuned LLaVA-v1.5-Instruct-7B, Qwen2VL-Instruct-2B, and Qwen2.5VL-Instruct-3B on \name-OS and evaluated their performance against models trained on the GPT-4o-based \name (10K subset) across benchmarks (\S\ref{sec:os_analysis_appendix}). Results show that models trained on \name-OS achieve ~99\% of the performance of their GPT-4o-based counterparts (\name), with both consistently outperforming base instruct models. These findings demonstrate that the VaPR pipeline generalizes well to open-source editors, allowing researchers to apply the \name framework to their own SFT datasets without relying on closed-source APIs.

\section{Related Work}
\paragraph{Large Vision Language Models (LVLMs)}
 LVLMs combine visual inputs from pre-trained vision encoders \citep{openai-clip, siglip} with large language models (LLMs) \citep{vicuna-v1.5, llama2, llama3, internlm2} via projection modules \citep{llava-v15, mplugowl2, blip2, cambrian}. They are typically trained in two stages: pretraining on large-scale image-text data to align modalities and instruction tuning or supervised finetuning on vision-language datasets for open-ended tasks. 

\paragraph{Preference Optimization in LVLMs}
Preference optimization, typically the third training stage for LVLMs, improves modality alignment and reduces hallucinations. It relies on reference datasets sourced via human annotation \citep{rlhfv, llava-rlhf}, AI annotation \citep{povid, silkie, ha-dpo}, or self-curation \citep{rlaifv, csr, stic, sima}. While human labels are high-quality but expensive, AI annotations using models like GPT-4V \citep{gpt-4v} are scalable but may introduce stylistic or length inconsistencies and distill hallucinations. Self-curated approaches generate multiple responses, rank them heuristically or with LVLMs, and select preference pairs, but often fail to control for stylistic or length biases - leading to noise, as both chosen and rejected responses may seem equally plausible \citep{vigor}. Our method builds on Direct Preference Optimization (DPO)~\citep{dpo} and introduces a post-SFT data construction pipeline that addresses these challenges. Unlike prior works \citep{povid, rlaifv, csr, stic, sima} that use VLMs for generation or scoring, we employ LLM-guided response editing to inject task-aware semantic errors into rejected responses while explicitly preserving style and length. This reduces the risk of DPO exploiting superficial cues. In contrast to \citet{mme5}, which uses LLMs to generate hard negatives for multilingual embedding training, our approach targets post-SFT preference alignment in English, where fine-grained semantic and stylistic control is crucial.

\section{Conclusion}
In this work, we address challenges in the alignment and reasoning capabilities of LVLMs by introducing \name, a hard-negative preference data generation framework. By constructing a high-quality preference dataset of 30K samples, we mitigate length and stylistic biases prevalent in existing synthetic datasets. We demonstrate significant improvements across benchmarks, particularly excelling in reasoning tasks and adversarial scenarios. Our analysis highlights the effectiveness of \name in improving vision-linguistic compositionality and reducing binary question bias, paving the way for more reliable and generalizable LVLMs.

Future work includes extending the framework to larger and more diverse datasets, applying it to broader reasoning tasks, and refining hard-negative generation to better target nuanced reasoning errors. Another promising direction is combining \name with self-preference methods in low-resource settings - using confident responses as positives and generating hard negatives via the \name pipeline to enable unsupervised preference data creation. We also plan to explore integrating \name with online preference optimization methods such as PPO and GRPO, particularly for complex reasoning tasks like math.

\section*{Acknowledgments}
This research is supported in part by the ECOLE program under Cooperative Agreement
HR00112390060 with the US Defense Advanced Research Projects Agency (DARPA), UCLA-Amazon Science Hub for Humanity and Artificial Intelligence, and OpenAI Researcher Access Program (Grant No. 0000004868).

\bibliography{colm2025_conference}
\bibliographystyle{colm2025_conference}

\clearpage
\appendix
\onecolumn
\section{Appendix Overview}
\label{sec:overview_appendix}

The appendix is organized into the following sections and subsections:

% Updated content with adjusted spacing
\begin{itemize}[itemsep=1em] % Adjust itemsep to control spacing
    \item \textbf{Detailed Methodology of \name Data Generation} (\S\ref{sec:datagen_appendix}):
    \begin{itemize}[itemsep=1em] % Adjust subitem spacing
        \item \textbf{Task Categories (\S\ref{sec:datagen_categories_appendix})}: Definition of the task categories included in the \name dataset. 
        \item \textbf{Prompt Design \& Examples (\S\ref{sec:datagen_prompts_appendix})}: Specific prompts used for data generation for each task category and corresponding qualitative examples illustrating the generated data.
        \item \textbf{Fine-grained length \& Stylistic analysis (\S\ref{sec:fine_grained_data_cmp_appendix})}
        \item \textbf{Human Study (\S\ref{sec:datagen_human_study_appendix})}: Annotation and evaluation setup.

    \end{itemize}
    
    \item \textbf{Experimental Setup} (\S\ref{sec:setup_appendix}):
    \begin{itemize}[itemsep=1em] % 
        \item \textbf{Training Setup (\S\ref{sec:training_setup_appendix})}: Details of the training setup for our models are provided here.
        \item \textbf{Evaluation Setup (\S\ref{sec:evaluation_setup_appendix})}: Implementation of the evaluation benchmarks.
    \end{itemize}
    
    \item \textbf{Extended Results} (\S\ref{sec:detailed_benchmarks_results}):
    \begin{itemize}[itemsep=1em] % Adjust subitem spacing
    
        \item \textbf{Data Scaling (\S\ref{sec:data_scale_appendix})}: Detailed results for \name model with increasing training dataset size.
        \item \textbf{Benchmark detailed results (\S\ref{sec:benchmark_detailed_results})}: Fine-grained evaluation across benchmark categories, for CVBench and MMStar.
        \item \textbf{Preference Dataset Comparison (\S\ref{sec:dataset_cmp_appendix})}: Detailed results for the comparison of base models trained on \name and other preference datasets.
        
    \end{itemize}
\end{itemize}

\clearpage

\section{Detailed Methodology of \name Data Generation}
\label{sec:datagen_appendix}
In this section, we provide the breakdown of \name dataset into task categories, prompt design \& examples for each category, setup for human study and linguistic \& length analysis study.

\subsection{Task Categories}
\label{sec:datagen_categories_appendix}

An essential component of our data generation pipeline is the categorization of filtered samples (\textasciitilde 270K) from the LLaVA-665K SFT set into distinct task categories. This ensures comprehensive coverage of perception, reasoning, and composite tasks (perception \& reasoning) in the final \name dataset. The \name dataset comprises ten task categories: object (like type, material, action), color, size, background (like weather, time of day, surrounding lighting), counting, spatial reasoning, existence, referential VQA (like color of an object on the left, etc), general reasoning (like abstract \& knowledge-based), and image captioning, as outlined in Table~\ref{tab:task_table}. The categorization process is carried out in the following three steps: \\

\begin{itemize}
        \item Task-specific keywords (see Table~\ref{tab:task_table}) are applied to the instructions in the filtered SFT set to assign a task category to each sample containing those keywords. \\
        \item Categorization follows a defined order:
        \begin{itemize}
            \item Samples are initially tagged into the categories: color, size, counting, spatial reasoning, background, existence, captioning, referential VQA (comparative reasoning) and general reasoning (in no specific order). Specifically for the existence category, we check the first word in the instruction.
            \item Remaining samples are assigned to the object category. This approach is necessary because object-related tasks, encompassing types and knowledge, require a broad, non-generalizable keyword list.
    \end{itemize}
    \item \textbf{Resolving Multiple Task Categories for a sample}: Samples tagged with multiple task categories are categorized as referential VQA (qualitative examples shown in \S\ref{sec:refvqa_datagen_prompts_appendix}). Thus, referential VQA consists of samples tagged as comparative reasoning and the above samples.
\end{itemize}

\begin{table}[h!]
\centering
\small
\begin{tabular}{|l|l|l|}
\hline
\textbf{Task} & \textbf{Task Type} & \textbf{Keyword} \\ \hline
Color & Perception & color(s) \\ \hline
Size & Perception & size(s) \\ \hline
Background & Perception & environment, time of, day, year, weather, lighting, \\ \hline
Counting & Reasoning & many, count(s), instance(s), counting \\ \hline
Spatial Reasoning & Reasoning & where, located, placed, positioned, left, right, \\
& & in front of, down, above, below
 \\ \hline
Existence & Perception \& Reasoning & are, is, can, do, does, would, will \\ \hline
General Reasoning & Perception \& Reasoning & could, would, might, purpose, reason, based, should \\ \hline
Referential VQA & Perception \& Reasoning & Samples with keywords from more than one task \\
& & Comparative Reasoning: comparison, difference, \\
& & closer, nearer, bigger \\ \hline
Image Captioning & Perception \& Reasoning & analyze, describe, write, elaborate, description, snapshot \\ \hline
\end{tabular}
\caption{Overview of tasks, task types (perception \& reasoning), and task-specific keywords used for \name dataset categorization. Keyword list is shortened for clarity. Note untagged samples are grouped together into the object category.}
\label{tab:task_table}
\end{table}

% This also aids in the development of task specific prompts, which we had found to be more effective in generating hard negative, specifically presevering stylistic and length similarity better than a generic prompt for all tasks. 

% \begin{itemize}
%     \item \textbf{Color}: 
    
%     \item \textbf{Size}: 
%     \item \textbf{Background}:
%     \item \textbf{Counting}:
%     \item \textbf{Spatial Reasoning}:
%     \item \textbf{Existence}:
%     \item \textbf{General Reasoning}:
%     \item \textbf{Image Captioning}: 
%     \item \textbf{Object}: 
    
% \end{itemize}

\clearpage
\subsection{Prompt Design \& Examples}
\label{sec:datagen_prompts_appendix}
In this section, we provide a detailed walkthrough of hard-negative generation for each task category, accompanied by additional examples from each category - object (\S\ref{sec:object_datagen_prompts_appendix}), color (\S\ref{sec:color_datagen_prompts_appendix}), size (\S\ref{sec:size_datagen_prompts_appendix}), background (\S\ref{sec:background_datagen_prompts_appendix}), counting (\S\ref{sec:counting_datagen_prompts_appendix}), spatial reasoning (\S \ref{sec:spatial_datagen_prompts_appendix}), existence (\S\ref{sec:existence_datagen_prompts_appendix}), 
referential VQA (\S\ref{sec:refvqa_datagen_prompts_appendix}),
image captioning (\S\ref{sec:captioning_datagen_prompts_appendix}),
general reasoning (\S\ref{sec:genreasoning_datagen_prompts_appendix}). In the prompt figures, spans of the chosen responses are highlighted in \textcolor{darkblue}{blue}, while the corresponding perturbed spans in the generated hard-negative responses are highlighted in \textcolor{darkgreen}{green}. Lastly, we also provide the algorithm and additional examples showing generation steps for task categories with penalty list (e.g. color) in \S\ref{sec:penalty_list_examples}.

\subsubsection{Object}
\label{sec:object_datagen_prompts_appendix}
\begin{table*}[h!]
\fontsize{9.0pt}{\baselineskip}\selectfont
\linespread{0.9}\selectfont
%%%%%%%%%%%%%%%%%%%%%%%%%%%%%%%%%%%%%%%%%%%%%%%%%%%%%%%%%
\centering
\includegraphics[scale=0.3]{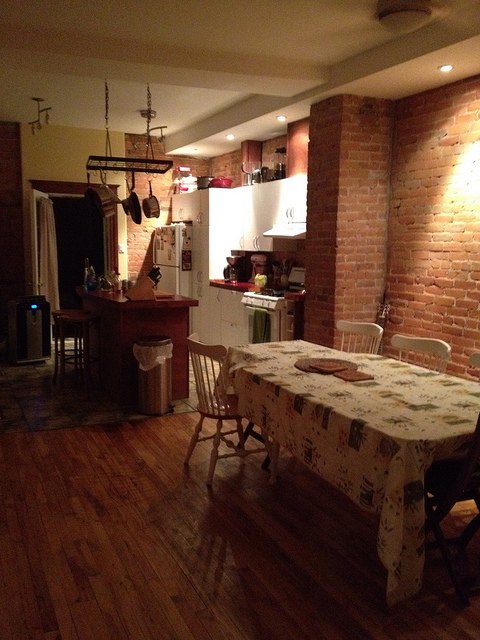}
%%%%%%%%%%%%%%%%%%%%%%%%%%%%%%%%%%%%%%%%%%%%%%%%%%%%%%%%%
\begin{mybody}
\textbf{Prompt:} \\ 
You are a ResponseEditorGPT who is given an instruction and a response generated by a vision-language model. The instruction asks about the objects, or actions. Your task is as follows:  \\

1. Modify the "objects" or "action" to make the "Original Response" about "objects" or "actions" incorrect. \\

2. "New Response" must be linguistically very similar to "Original Response" and must be incorrect.\\

3. You must ensure changes must be realistic given world knowledge. \\

4. You can minimally change other spans of the sentence to grammatical correctness and fluency.\\

5. The output format should be "New Response:"\\

\textbf{Instruction:} What type of flooring is in the room? \\

\textbf{Original Response:} The room has hard wood floors. \\

Your Turn  \\
New Response:
\end{mybody}
%%%%%%%%%%%%%%%%%%%%%%%%%%%%%%%%%%%%%%%%%%%%%%%%%%%%%%%%%
\begin{mybody}
\textbf{Chosen Response:} 
The room has \textcolor{darkblue}{hard wood floors}.
\end{mybody}

\begin{mybody}
\textbf{Hard-negative Rejected Response:} The room has \textcolor{darkgreen}{carpeted floors}.
\end{mybody}
%%%%%%%%%%%%%%%%%%%%%%%%%%%%%%%%%%%%%%%%%%%%%%%%%%%%%%%%%
\captionof{figure}{Example prompt and generated hard-negative for \name sample of object task category.}
\label{fig:object_prompt_ex}
\end{table*}
\clearpage

\begin{figure*}[h!]
\centering
\begin{subfigure}[b]{0.48\textwidth}
    \fontsize{9.0pt}{\baselineskip}\selectfont
    \linespread{0.9}\selectfont
    \begin{mybody}
        \centering
        \includegraphics[width=0.7\textwidth,height=0.9\textwidth]{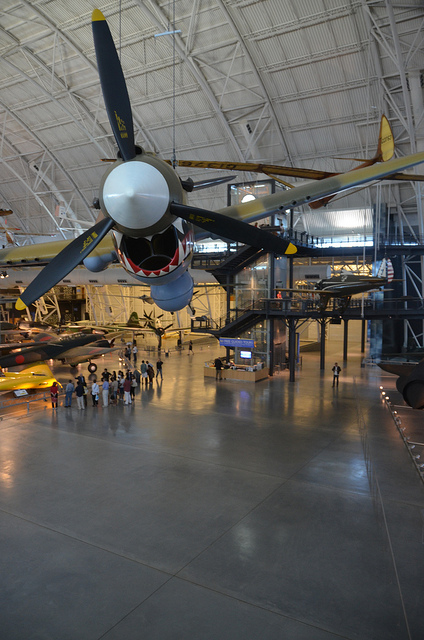}
        \end{mybody}
        \begin{mybody}
        \textbf{Instruction:} What type of airplane is displayed in the image? 
        \end{mybody}
        \begin{mybody}
        \textbf{Chosen Response:} 
        The image displays a World War era airplane with a \textcolor{darkblue}{propeller} hanging in a museum.
        \end{mybody}
        \begin{mybody}
        \textbf{Rejected Response:} The image displays a World War era airplane with \textcolor{darkgreen}{jet engines} hanging in a museum.
        \end{mybody}
    \caption{Task requiring perception of an object's components}
\end{subfigure}
\hfill
\begin{subfigure}[b]{0.48\textwidth}
    \fontsize{9.0pt}{\baselineskip}\selectfont
    \linespread{0.9}\selectfont
        \begin{mybody}
        \centering
        \includegraphics[width=0.7\textwidth,height=0.9\textwidth]{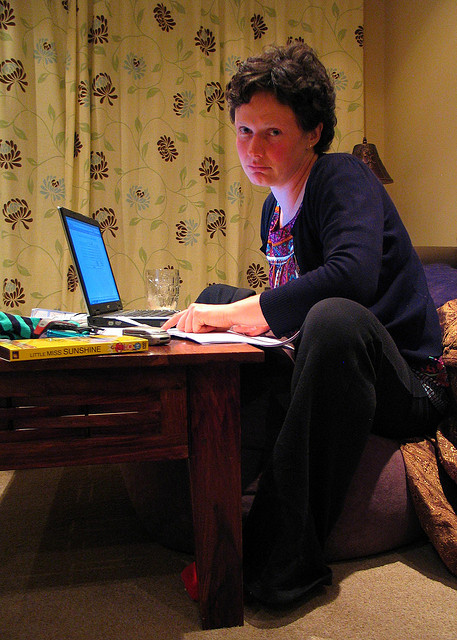}
        \end{mybody}
        \begin{mybody}
        \textbf{Instruction:} What is the woman doing in the image? 
        \end{mybody}
        \begin{mybody}
        \textbf{Chosen Response:} 
        The woman is \textcolor{darkblue}{sitting at} a desk or table, working on a laptop computer.  
        \end{mybody}
        \begin{mybody}
        \textbf{Rejected Response:} The woman is \textcolor{darkgreen}{standing next to} a desk or table, working on a laptop computer.
        \end{mybody}
    \caption{Task requiring perception of an object's action}
\end{subfigure}

\vspace{0.25cm}
\begin{subfigure}[b]{0.48\textwidth}
    \fontsize{9.0pt}{\baselineskip}\selectfont
    \linespread{0.9}\selectfont
        \begin{mybody}
        \centering
        \includegraphics[width=0.8\textwidth,height=0.8\textwidth]{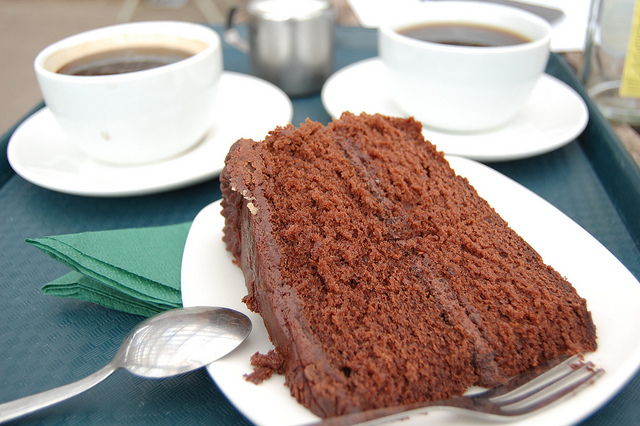}
        \end{mybody}
        \begin{mybody}
        \textbf{Instruction:} What type of cake is on the plate? 
        \end{mybody}
        \begin{mybody}
        \textbf{Chosen Response:} 
        There is a slice of \textcolor{darkblue}{chocolate} cake on the plate. 
        \end{mybody}
        \begin{mybody}
        \textbf{Rejected Response:} There is a slice of \textcolor{darkgreen}{strawberry} cake on the plate. 
        \end{mybody}
    \caption{Task requiring perception of an object's type}
\end{subfigure}
\hfill
\begin{subfigure}[b]{0.48\textwidth}
    \fontsize{9.0pt}{\baselineskip}\selectfont
    \linespread{0.9}\selectfont
        \begin{mybody}
        \centering
        \includegraphics[width=0.8\textwidth,height=0.8\textwidth]{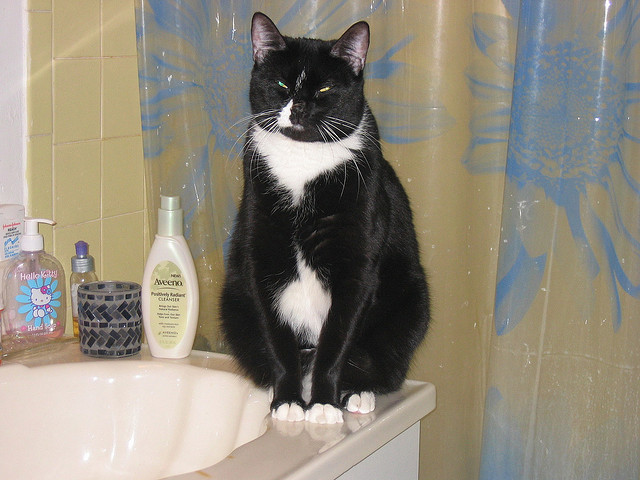}
        \end{mybody}
        \begin{mybody}    
        \textbf{Instruction:} What is the cat doing in the image? 
        \end{mybody}
        \begin{mybody}  
        \textbf{Chosen Response:} The cat is \textcolor{darkblue}{sitting} on the edge of a bathroom sink.
        \end{mybody}
        \begin{mybody}
        \textbf{Rejected Response:} The cat is \textcolor{darkgreen}{sleeping} on the edge of a bathroom sink. 
        \end{mybody}
    \caption{Task requiring perception of an object's action}
\end{subfigure}
\caption{Qualitative examples of the object task category, like object perception, its type, components, and associated actions.}
\label{fig:object_ex}
\end{figure*}

\clearpage

\subsubsection{Color}
\label{sec:color_datagen_prompts_appendix}

\begin{table*}[h!]
\fontsize{9.0pt}{\baselineskip}\selectfont
\linespread{0.9}\selectfont
%%%%%%%%%%%%%%%%%%%%%%%%%%%%%%%%%%%%%%%%%%%%%%%%%%%%%%%%%
\centering
\includegraphics[scale=0.3]{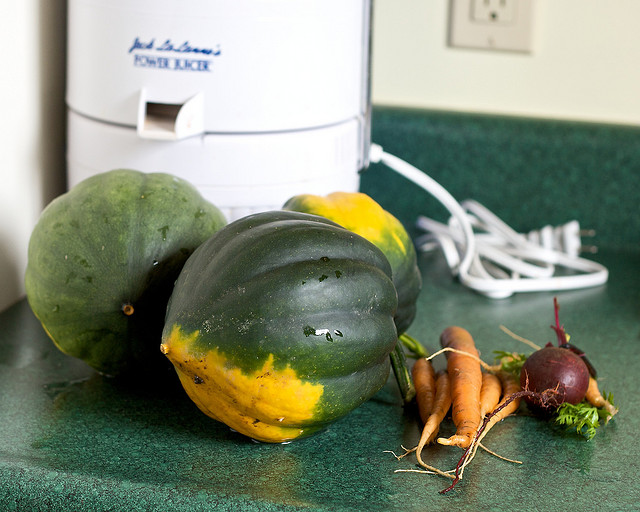}
%%%%%%%%%%%%%%%%%%%%%%%%%%%%%%%%%%%%%%%%%%%%%%%%%%%%%%%%%
\begin{mybody}
\textbf{Prompt:} \\ 
You are a ResponseEditorGPT who is given an instruction and a response generated by a vision-language model. The instruction asks about the colors of objects, environments, or themes. Your task is as follows: \\

1. Modify the "colors" of all objects, environments, or themes in the response to make "Original Response" about "colors" incorrect.  \\

2. You must only change the "colors", so that "New Response" is linguistically very similar to "Original Response" and is incorrect.\\

3. The "New colors" you use to replace original colors must be unique and not be too descriptive. \\

4. The "New colors" must be realistically possible, considering the object they describe. \\

5. You cannot use colors in the penalty list. \\

6. You can minimally change other spans of the sentence to grammatical correctness and fluency.\\

7. List the "New colors" you replace within the response. \\

\textbf{Penalty list:} [white, blue] \\

\textbf{Instruction:} What color is the kitchen counter where the vegetables are placed? \\

\textbf{Original Response:} The kitchen counter where the vegetables are placed is green. \\

Your Turn  \\
New Response: \\
New Colors:
\end{mybody}
%%%%%%%%%%%%%%%%%%%%%%%%%%%%%%%%%%%%%%%%%%%%%%%%%%%%%%%%%
\begin{mybody}
\textbf{Chosen Response:} 
The kitchen counter where the vegetables are placed is \textcolor{darkblue}{green}.
\end{mybody}

\begin{mybody}
\textbf{Hard-negative Rejected Response:} The kitchen counter where the vegetables are placed is \textcolor{darkgreen}{yellow.}
\end{mybody}
\begin{mybody}
\textbf{New Colors: } [yellow]
\end{mybody}
%%%%%%%%%%%%%%%%%%%%%%%%%%%%%%%%%%%%%%%%%%%%%%%%%%%%%%%%%
\captionof{figure}{Example prompt and generated hard-negative for \name sample of color task category. Here, the penalty list is simulated for illustration. New colors represent additional information generated alongside the hard-negative rejected response to maintain the penalty list. We maintain a list of most perturbed color values (size 2), updated after every 10 hard-negative rejected response generations. Note, if more than one color is perturbed, "New Colors" will have more elements in the list.}
\label{fig:color_prompt_ex}
\end{table*}
\clearpage

\begin{figure*}[h!]
\centering
\begin{subfigure}[b]{0.48\textwidth}
    \fontsize{9.0pt}{\baselineskip}\selectfont
    \linespread{0.9}\selectfont
    \begin{mybody}
        \centering
        \includegraphics[width=0.8\textwidth,height=0.8\textwidth]{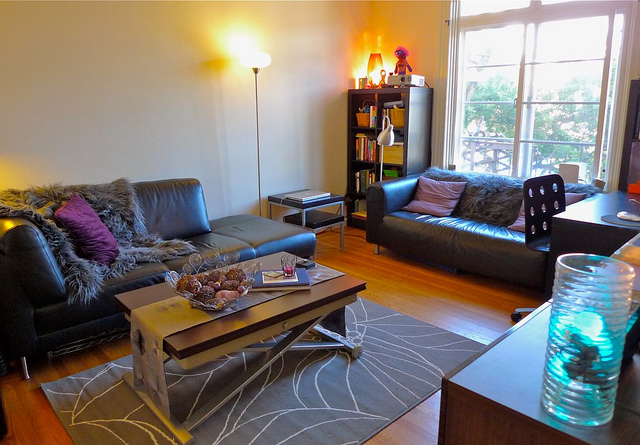}
        \end{mybody}
        \begin{mybody}
        \textbf{Instruction:} What color are the couches in the living room?
        \end{mybody}
        \begin{mybody}
        \textbf{Chosen Response:} 
        The couches in the living room are \textcolor{darkblue}{black}.
        \end{mybody}
        \begin{mybody}
        \textbf{Rejected Response:} The couches in the living room are \textcolor{darkgreen}{orange}.
        \end{mybody}
    \caption{Task requiring color perception of an object.}
\end{subfigure}
\hfill
\begin{subfigure}[b]{0.48\textwidth}
    \fontsize{9.0pt}{\baselineskip}\selectfont
    \linespread{0.9}\selectfont
        \begin{mybody}
        \centering
        \includegraphics[width=0.71\textwidth,height=0.71\textwidth]{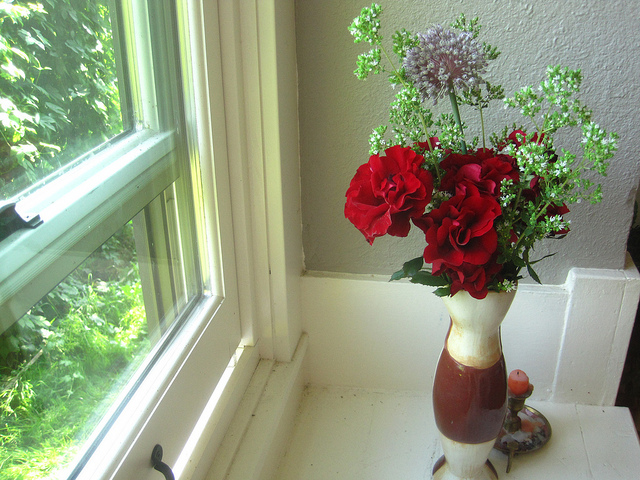}
        \end{mybody}
        \begin{mybody}
        \textbf{Instruction:} What are the colors of the flowers in the vase?
        \end{mybody}
        \begin{mybody}
        \textbf{Chosen Response:} 
        The colors of the flowers in the vase are \textcolor{darkblue}{red, green, and purple}.
        \end{mybody}
        \begin{mybody}
        \textbf{Rejected Response:} The colors of the flowers in the vase are \textcolor{darkgreen}{yellow, blue, and orange}.
        \end{mybody}
    \caption{Task requiring color perception of mulitple objects.}
\end{subfigure}

\vspace{0.25cm}
\begin{subfigure}[b]{0.48\textwidth}
    \fontsize{9.0pt}{\baselineskip}\selectfont
    \linespread{0.9}\selectfont
        \begin{mybody}
        \centering
        \includegraphics[width=0.75\textwidth,height=0.75\textwidth]{images/prompt_images/color/color_ex3.jpg}
        \end{mybody}
        \begin{mybody}
        \textbf{Instruction:} What is the color of the surfboard the dog is sitting on?
        \end{mybody}
        \begin{mybody}
        \textbf{Chosen Response:} 
        The color of the surfboard the dog is sitting on is \textcolor{darkblue}{blue}.
        \end{mybody}
        \begin{mybody}
        \textbf{Rejected Response:} The color of the surfboard the dog is sitting on is \textcolor{darkgreen}{orange}.
        \end{mybody}
    \caption{Task requiring color perception of an object.}
\end{subfigure}
\hfill
\begin{subfigure}[b]{0.48\textwidth}
    \fontsize{9.0pt}{\baselineskip}\selectfont
    \linespread{0.9}\selectfont
        \begin{mybody}
        \centering
        \includegraphics[width=0.7\textwidth,height=0.7\textwidth]{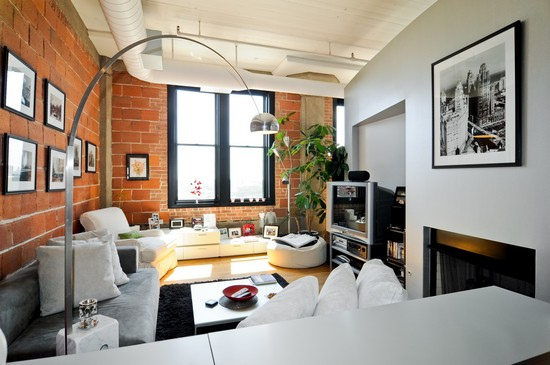}
        \end{mybody}
        \begin{mybody}    
        \textbf{Instruction:} What is the overall color theme of the living room?
        \end{mybody}
        \begin{mybody}  
        \textbf{Chosen Response:} The overall color theme of the living room is predominantly \textcolor{darkblue}{white}, with \textcolor{darkblue}{white} furniture and \textcolor{darkblue}{white} elements on the brick wall.
        \end{mybody}
        \begin{mybody}
        \textbf{Rejected Response:} The overall color theme of the living room is predominantly \textcolor{darkgreen}{turquoise}, with \textcolor{darkgreen}{turquoise} furniture and \textcolor{darkgreen}{turquoise} elements on the brick wall.
        \end{mybody}
    \caption{Task requiring color perception of the theme of a background setting.}
\end{subfigure}
\caption{Qualitative examples of the color task category, like perception of colors of a single object, multiple objects and background theme.}
\label{fig:color_ex}
\end{figure*}

\clearpage

\subsubsection{Size}
\label{sec:size_datagen_prompts_appendix}

\begin{table*}[h!]
\fontsize{9.0pt}{\baselineskip}\selectfont
\linespread{0.9}\selectfont
%%%%%%%%%%%%%%%%%%%%%%%%%%%%%%%%%%%%%%%%%%%%%%%%%%%%%%%%%
\centering
\includegraphics[scale=0.35]{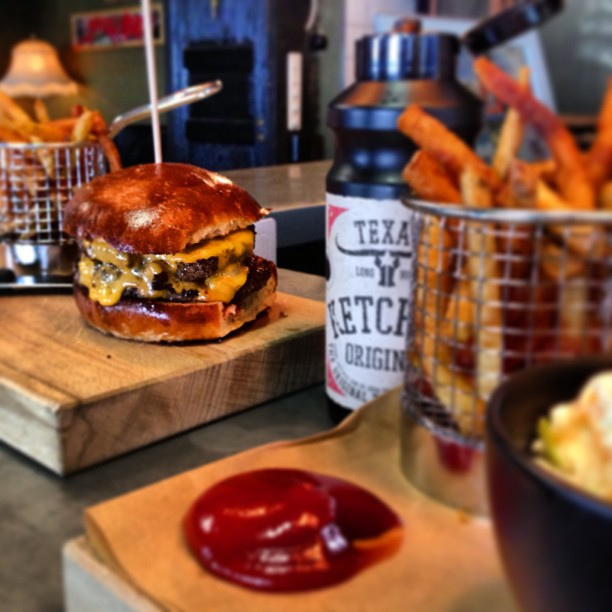}
%%%%%%%%%%%%%%%%%%%%%%%%%%%%%%%%%%%%%%%%%%%%%%%%%%%%%%%%%
\begin{mybody}
\textbf{Prompt:} \\ 
You are a ResponseEditorGPT who is given an instruction and a response generated by a vision-language model. The instruction asks about the "size" of objects or themes. Your task is as follows:  \\

1. Modify the "size" of the objects or themes to make the "Original Response" about the "size" incorrect.  \\

2. "New Response" must be linguistically very similar to "Original Response" and must be incorrect.\\

3. You must ensure changes must be realistic given world knowledge. \\

4. You can minimally change other spans of the sentence to grammatical correctness and fluency.\\

5. The output format should be "New Response:" \\

\textbf{Instruction:} How is the cheeseburger described in terms of size and ingredients? \\

\textbf{Original Response:} The cheeseburger is described as massive and containing double cheese layers. \\

Your Turn  \\
New Response: 
\end{mybody}
%%%%%%%%%%%%%%%%%%%%%%%%%%%%%%%%%%%%%%%%%%%%%%%%%%%%%%%%%
\begin{mybody}
\textbf{Chosen Response:} 
The cheeseburger is described as \textcolor{darkblue}{massive} and containing \textcolor{darkblue}{double cheese layers}.
\end{mybody}

\begin{mybody}
\textbf{Hard-negative Rejected Response:} The cheeseburger is described as \textcolor{darkgreen}{small} and containing \textcolor{darkgreen}{a single cheese layer}.
\end{mybody}
%%%%%%%%%%%%%%%%%%%%%%%%%%%%%%%%%%%%%%%%%%%%%%%%%%%%%%%%%
\captionof{figure}{Example prompt and generated hard-negative for \name sample of size task category.}
\label{fig:size_prompt_ex}
\end{table*}
\clearpage

\begin{figure*}[h!]
\centering
\begin{subfigure}[b]{0.48\textwidth}
    \fontsize{9.0pt}{\baselineskip}\selectfont
    \linespread{0.9}\selectfont
    \begin{mybody}
        \centering
        \includegraphics[width=0.8\textwidth,height=0.8\textwidth]{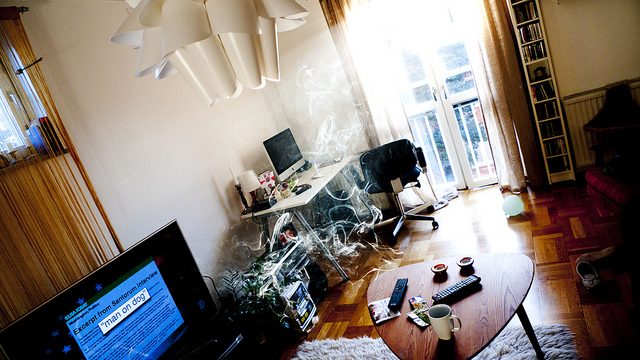}
        \end{mybody}
        \begin{mybody}
        \textbf{Instruction:} What is the general size of the room?
        \end{mybody}
        \begin{mybody}
        \textbf{Chosen Response:} 
        The room is described as a \textcolor{darkblue}{small living area}, which implies that it is \textcolor{darkblue}{not very large or spacious}.
        \end{mybody}
        \begin{mybody}
        \textbf{Rejected Response:} The room is described as an \textcolor{darkgreen}{expansive living area}, which implies that it is \textcolor{darkgreen}{quite large and spacious}.
        \end{mybody}
    \caption{Task requiring thematic size perception, here room.}
\end{subfigure}
\hfill
\begin{subfigure}[b]{0.48\textwidth}
    \fontsize{9.0pt}{\baselineskip}\selectfont
    \linespread{0.9}\selectfont
        \begin{mybody}
        \centering
        \includegraphics[width=0.71\textwidth,height=0.71\textwidth]{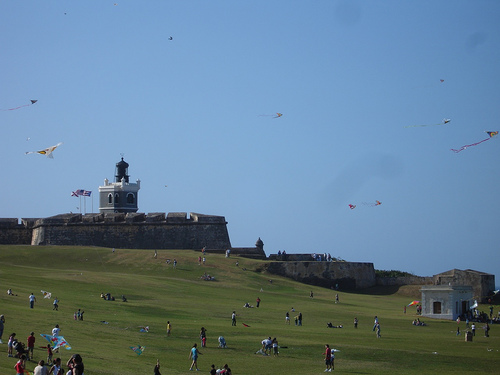}
        \end{mybody}
        \begin{mybody}
        \textbf{Instruction:} How would you describe the size of the group of people flying kites?
        \end{mybody}
        \begin{mybody}
        \textbf{Chosen Response:} 
        The group of people flying kites is \textcolor{darkblue}{large}, indicating that it is a \textcolor{darkblue}{popular and well-attended event or gathering}.
        \end{mybody}
        \begin{mybody}
        \textbf{Rejected Response:} The group of people flying kites is \textcolor{darkgreen}{small}, indicating that it is a \textcolor{darkgreen}{more intimate and possibly less publicized event or gathering}. 
        \end{mybody}
    \caption{Task requiring thematic size perception, here gathering of people.}
\end{subfigure}

\vspace{0.25cm}
\begin{subfigure}[b]{0.48\textwidth}
    \fontsize{9.0pt}{\baselineskip}\selectfont
    \linespread{0.9}\selectfont
        \begin{mybody}
        \centering
        \includegraphics[width=0.6\textwidth,height=0.7\textwidth]{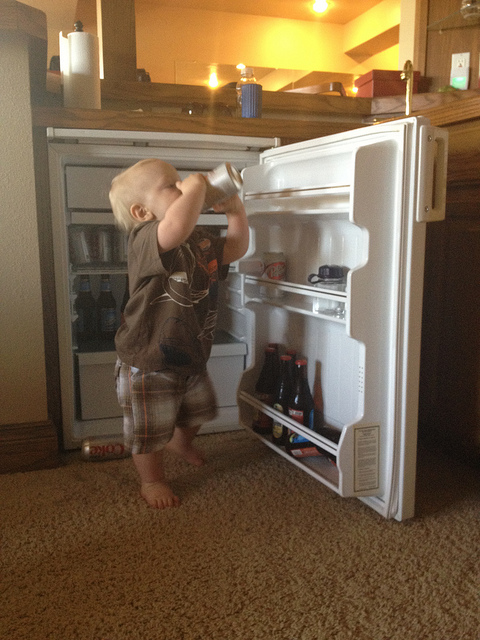}
        \end{mybody}
        \begin{mybody}
        \textbf{Instruction:} What's the size of the refrigerator in the image?
        \end{mybody}
        \begin{mybody}
        \textbf{Chosen Response:} 
        The refrigerator in the image is a \textcolor{darkblue}{small or mini-sized refrigerator}.
        \end{mybody}
        \begin{mybody}
        \textbf{Rejected Response:} The refrigerator in the image is a \textcolor{darkgreen}{large, full-sized refrigerator}.
        \end{mybody}
    \caption{Task requiring size perception of an object.}
\end{subfigure}
\hfill
\begin{subfigure}[b]{0.48\textwidth}
    \fontsize{9.0pt}{\baselineskip}\selectfont
    \linespread{0.9}\selectfont
        \begin{mybody}
        \centering
        \includegraphics[width=0.6\textwidth,height=0.7\textwidth]{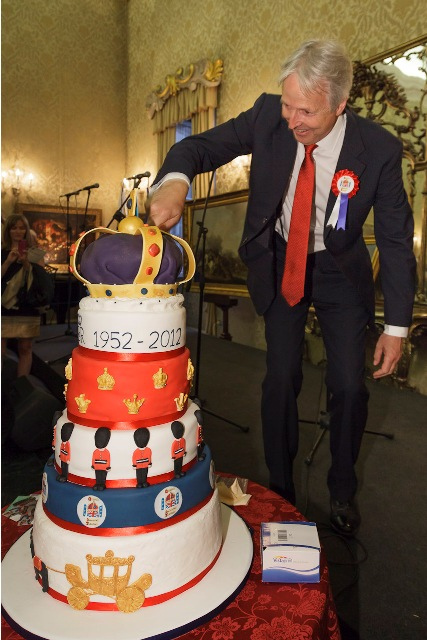}
        \end{mybody}
        \begin{mybody}    
        \textbf{Instruction:} What is the size of the cake in the image?
        \end{mybody}
        \begin{mybody}  
        \textbf{Chosen Response:} The cake in the image is \textcolor{darkblue}{quite large and tall, comprising multiple layers}.
        \end{mybody}
        \begin{mybody}
        \textbf{Rejected Response:} The cake in the image is \textcolor{darkgreen}{small and flat, consisting of a single layer}.
        \end{mybody}
    \caption{Task requiring size perception of an object.}
\end{subfigure}
\caption{Qualitative examples of the size task category, like size perception of an object or broadly a theme (eg. room, gathering of people)}
\label{fig:size_ex}
\end{figure*}

\clearpage

\subsubsection{Background}
\label{sec:background_datagen_prompts_appendix}
\begin{table*}[h!]
\fontsize{9.0pt}{\baselineskip}\selectfont
\linespread{0.9}\selectfont
%%%%%%%%%%%%%%%%%%%%%%%%%%%%%%%%%%%%%%%%%%%%%%%%%%%%%%%%%
\centering
\includegraphics[scale=0.4]{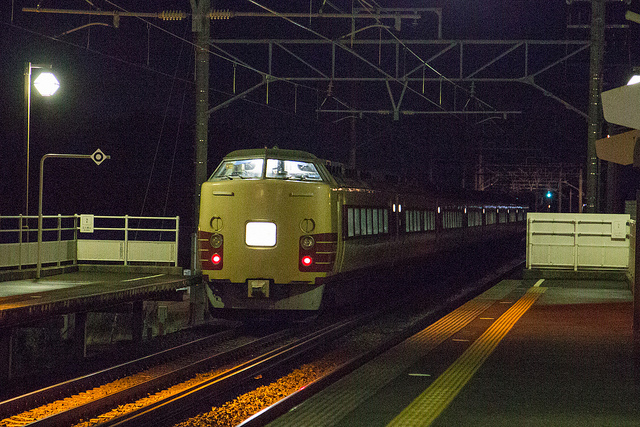}
%%%%%%%%%%%%%%%%%%%%%%%%%%%%%%%%%%%%%%%%%%%%%%%%%%%%%%%%%
\begin{mybody}
\textbf{Prompt:} \\ 
You are a ResponseEditorGPT who is given an instruction and a response generated by a vision-language model. The instruction asks about the "time", "weather", or "environment" of events, surroundings, or themes. Your task is as follows: \\ 

1. Modify the "time", "weather", or "environment" of events, surroundings, or themes to make the "Original Response" about the "time", "weather", or "environment" incorrect. \\

2. "New Response" must be linguistically very similar to "Original Response" and must be incorrect.\\

3. You must ensure changes must be realistic given world knowledge. \\

4. You can minimally change other spans of the sentence to grammatical correctness and fluency.\\

5. The output format should be "New Response:" \\

\textbf{Instruction:} What time of day is it in the image? \\

\textbf{Original Response:} It is nighttime in the image, as evidenced by the dark sky background. \\

Your Turn  \\
New Response: 
\end{mybody}
%%%%%%%%%%%%%%%%%%%%%%%%%%%%%%%%%%%%%%%%%%%%%%%%%%%%%%%%%
\begin{mybody}
\textbf{Chosen Response:} 
It is \textcolor{darkblue}{nighttime} in the image, as evidenced by the \textcolor{darkblue}{dark} sky background.
\end{mybody}

\begin{mybody}
\textbf{Hard-negative Rejected Response:} It is \textcolor{darkgreen}{twilight} in the image, as evidenced by the \textcolor{darkgreen}{darkening} sky background.
\end{mybody}
%%%%%%%%%%%%%%%%%%%%%%%%%%%%%%%%%%%%%%%%%%%%%%%%%%%%%%%%%
\captionof{figure}{Example prompt and generated hard-negative for \name sample of background task category. Notably, the slight linguistic change from "dark" to "darkening" has significant change in semantics.}
\label{fig:background_prompt_ex}
\end{table*}
\clearpage

\begin{figure*}[h!]
\centering
\begin{subfigure}[b]{0.48\textwidth}
    \fontsize{9.0pt}{\baselineskip}\selectfont
    \linespread{0.9}\selectfont
    \begin{mybody}
        \centering
        \includegraphics[width=0.85\textwidth,height=0.85\textwidth]{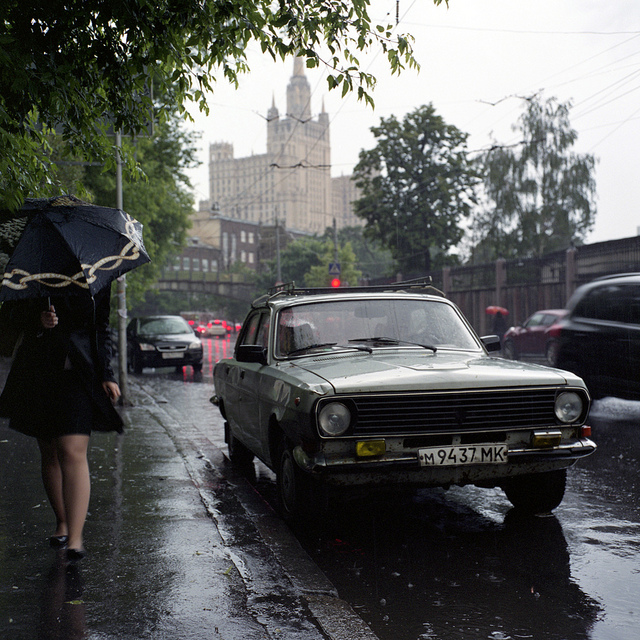}
        \end{mybody}
        \begin{mybody}
        \textbf{Instruction:} What's the weather like in the image?
        \end{mybody}
        \begin{mybody}
        \textbf{Chosen Response:} 
        The weather in the image is \textcolor{darkblue}{rainy}, creating a \textcolor{darkblue}{wet} atmosphere on the street.
        \end{mybody}
        \begin{mybody}
        \textbf{Rejected Response:} The weather in the image is \textcolor{darkgreen}{snowy}, creating a \textcolor{darkgreen}{cold} atmosphere on the street.
        \end{mybody}
    \caption{Task requiring perception of weather.}
\end{subfigure}
\hfill
\begin{subfigure}[b]{0.48\textwidth}
    \fontsize{9.0pt}{\baselineskip}\selectfont
    \linespread{0.9}\selectfont
        \begin{mybody}
        \centering
        % 000000488658
        \includegraphics[width=0.6\textwidth,height=0.66\textwidth]{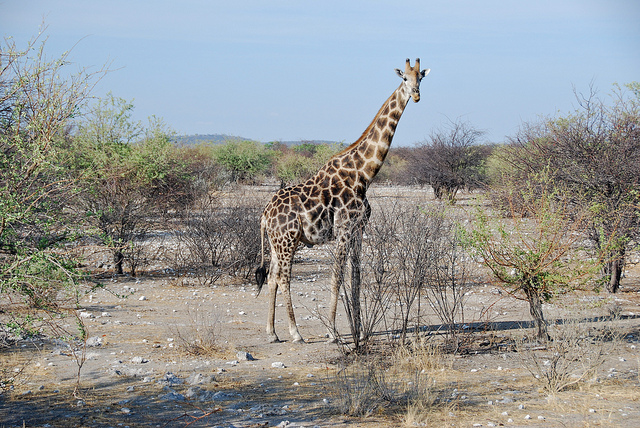}
        \end{mybody}
        \begin{mybody}
        \textbf{Instruction:} What kind of environment is the giraffe in?
        \end{mybody}
        \begin{mybody}
        \textbf{Chosen Response:} 
        The giraffe is in a \textcolor{darkblue}{dry, arid environment with brown vegetation, scrub bushes, and sand in the field}. It appears to be a somewhat \textcolor{darkblue}{harsh}, wild habitat that the giraffe is navigating.
        \end{mybody}
        \begin{mybody}
        \textbf{Rejected Response:} The giraffe is in a \textcolor{darkgreen}{lush, green environment with dense vegetation, leafy bushes, and a grassy field}. It appears to be a somewhat \textcolor{darkgreen}{fertile}, wild habitat that the giraffe is navigating. 
        \end{mybody}
    \caption{Task requiring perception of environment.}
\end{subfigure}

\vspace{0.25cm}
\begin{subfigure}[b]{0.48\textwidth}
    \fontsize{9.0pt}{\baselineskip}\selectfont
    \linespread{0.9}\selectfont
        \begin{mybody}
        \centering
        \includegraphics[width=0.68\textwidth,height=0.7\textwidth]{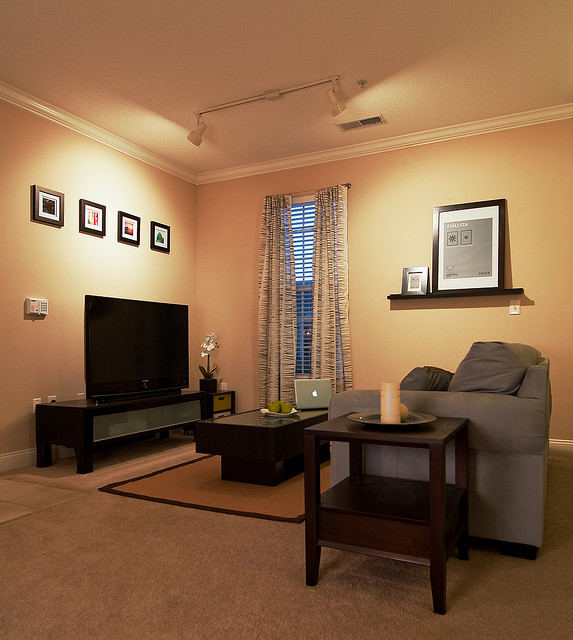}
        \end{mybody}
        \begin{mybody}
        \textbf{Instruction:} What type of lighting is featured in the living room? 
        \end{mybody}
        \begin{mybody}
        \textbf{Chosen Response:} 
        The living room features \textcolor{darkblue}{spot lighting}, which provides a \textcolor{darkblue}{focused illumination on specific areas or objects within the room}.
        \end{mybody}
        \begin{mybody}
        \textbf{Rejected Response:} The living room features \textcolor{darkgreen}{ambient} lighting, which provides a \textcolor{darkgreen}{general, diffused illumination throughout the room}.
        \end{mybody}
    \caption{Task requiring perception of background lighting.}
\end{subfigure}
\hfill
\begin{subfigure}[b]{0.48\textwidth}
    \fontsize{9.0pt}{\baselineskip}\selectfont
    \linespread{0.9}\selectfont
        \begin{mybody}
        \centering
        \includegraphics[width=0.68\textwidth,height=0.68\textwidth]{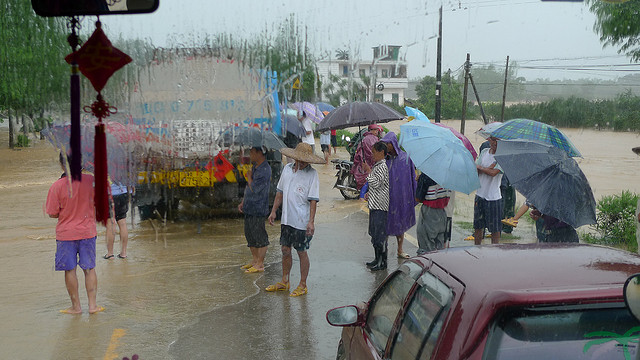}
        \end{mybody}
        \begin{mybody}    
        \textbf{Instruction:} What are the weather conditions in the image?
        \end{mybody}
        \begin{mybody}  
        \textbf{Chosen Response:} The weather conditions in the image are rainy, as evidenced by the group of people holding umbrellas to protect themselves from the \textcolor{darkblue}{pouring rain}.
        \end{mybody}
        \begin{mybody}
        \textbf{Rejected Response:} The weather conditions in the image are rainy, as evidenced by the group of people holding umbrellas to protect themselves from the \textcolor{darkgreen}{light drizzle}.
        \end{mybody}
    \caption{Task requiring perception of weather.}
\end{subfigure}
\caption{Qualitative examples of the background task category, like perception of weather, environment, lighting, and time of the day.}
\label{fig:background_ex}
\end{figure*}

\clearpage

\subsubsection{Counting}
\label{sec:counting_datagen_prompts_appendix}

\begin{table*}[h!]
\fontsize{9.0pt}{\baselineskip}\selectfont
\linespread{0.9}\selectfont
%%%%%%%%%%%%%%%%%%%%%%%%%%%%%%%%%%%%%%%%%%%%%%%%%%%%%%%%%
\centering
\includegraphics[scale=0.26]{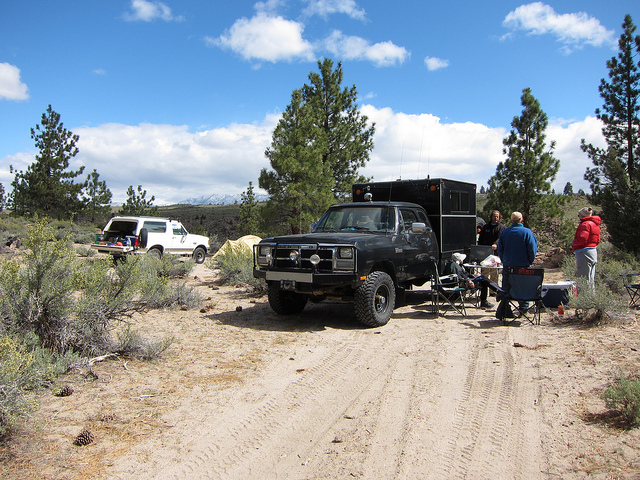}
%%%%%%%%%%%%%%%%%%%%%%%%%%%%%%%%%%%%%%%%%%%%%%%%%%%%%%%%%
\begin{mybody}
\textbf{Prompt:} \\ 
You are a ResponseEditorGPT who is given an instruction and a response generated by a vision-language model. The instruction asks about the counts of objects. Your task is as follows: \\

1. Modify the "counts" of all objects in the response to make "Original Response" about "counts" incorrect.  \\

2. You must only change the "counts", so that "New Response" is linguistically very similar to "Original Response" and is incorrect.\\

3. The "New counts" you use to replace original colors must be unique and not be too descriptive. \\

4. The "New counts" must be realistically possible, considering the object they describe. \\

5. You cannot use counts in the penalty list, neither the word form in the penalty nor its numerical form. \\

6. You can minimally change other spans of the sentence to grammatical correctness and fluency.\\

7. List the "New counts" you replace within the response. \\

\textbf{Penalty list:} [two, seven] \\

\textbf{Instruction:} How many people are visible near the truck in the image? \\

\textbf{Original Response:} There are four people visible near the truck in the image. Three people are standing in front of the truck, while another person is in the background. \\

Your Turn  \\
New Response: \\
New Colors:
\end{mybody}
%%%%%%%%%%%%%%%%%%%%%%%%%%%%%%%%%%%%%%%%%%%%%%%%%%%%%%%%%
\begin{mybody}
\textbf{Chosen Response:} 
There are \textcolor{darkblue}{four} people visible near the truck in the image. \textcolor{darkblue}{Three} people are standing in front of the truck, while another person is in the background. 
\end{mybody}

\begin{mybody}
\textbf{Hard-negative Rejected Response:} There are \textcolor{darkgreen}{six} people visible near the truck in the image. \textcolor{darkgreen}{Five} people are standing in front of the truck, while another person is in the background.
\end{mybody}
\begin{mybody}
\textbf{New Counts: } [six, five]
\end{mybody}
%%%%%%%%%%%%%%%%%%%%%%%%%%%%%%%%%%%%%%%%%%%%%%%%%%%%%%%%%
\captionof{figure}{Example prompt and generated hard-negative for \name sample of counting task category. Here, the penalty list is simulated for illustration. New counts represent additional information generated alongside the hard-negative rejected response to maintain the penalty list. Similar to color task category, We maintain a list of most perturbed count values (size 2), updated after every 10 hard-negative rejected response generations. Note, if more than one count is perturbed, "New Counts" will have more elements in the list.}
\label{fig:counting_prompt_ex}
\end{table*}
\clearpage

\begin{figure*}[h!]
\centering
\begin{subfigure}[b]{0.48\textwidth}
    \fontsize{9.0pt}{\baselineskip}\selectfont
    \linespread{0.9}\selectfont
    \begin{mybody}
        \centering
        \includegraphics[width=0.75\textwidth,height=0.75\textwidth]{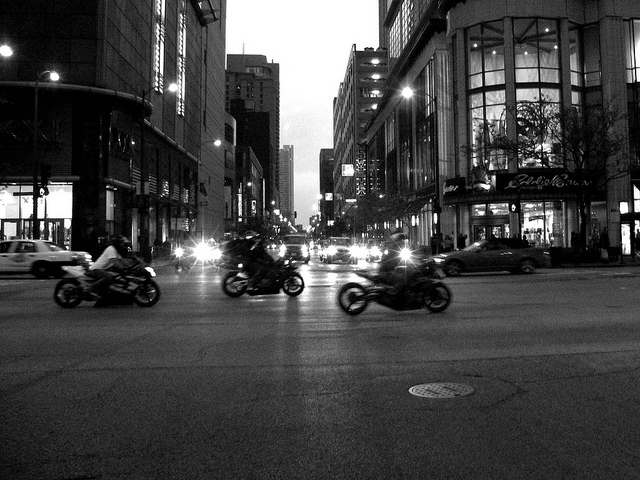}
        \end{mybody}
        \begin{mybody}
        \textbf{Instruction:} How many motorcycles are in the image?
        \end{mybody}
        \begin{mybody}
        \textbf{Chosen Response:} 
        There are \textcolor{darkblue}{three} motorcycles in the image.
        \end{mybody}
        \begin{mybody}
        \textbf{Rejected Response:} There are \textcolor{darkgreen}{six} motorcycles in the image.
        \end{mybody}
    \caption{Task requiring counting capability.}
\end{subfigure}
\hfill
\begin{subfigure}[b]{0.48\textwidth}
    \fontsize{9.0pt}{\baselineskip}\selectfont
    \linespread{0.9}\selectfont
        \begin{mybody}
        \centering
        \includegraphics[width=0.75\textwidth,height=0.75\textwidth]{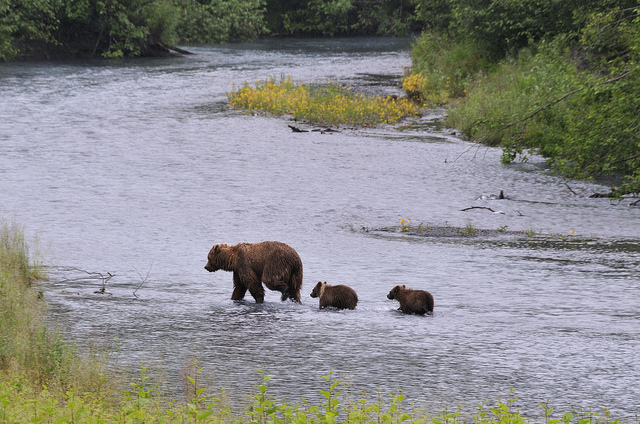}
        \end{mybody}
        \begin{mybody}
        \textbf{Instruction:} How many bears are present in the image?
        \end{mybody}
        \begin{mybody}
        \textbf{Chosen Response:} 
        There are \textcolor{darkblue}{three} bears present in the image - an adult bear and \textcolor{darkblue}{two} bear cubs.
        \end{mybody}
        \begin{mybody}
        \textbf{Rejected Response:} There are \textcolor{darkgreen}{seven} bears present in the image - an adult bear and \textcolor{darkgreen}{six} bear cubs.
        \end{mybody}
    \caption{Task requiring counting capability.}
\end{subfigure}

\vspace{0.25cm}
\begin{subfigure}[b]{0.48\textwidth}
    \fontsize{9.0pt}{\baselineskip}\selectfont
    \linespread{0.9}\selectfont
        \begin{mybody}
        \centering
        \includegraphics[width=0.7\textwidth,height=0.7\textwidth]{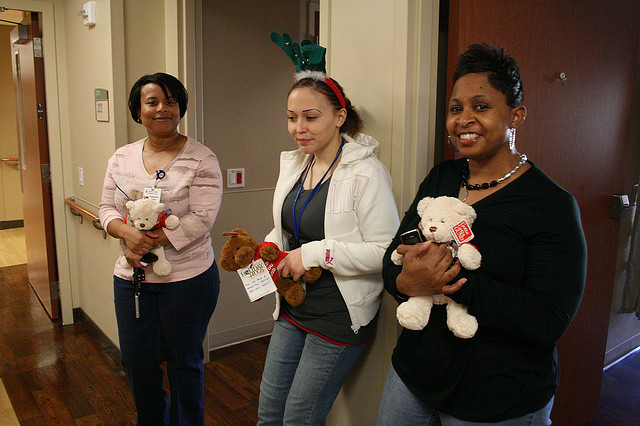}
        \end{mybody}
        \begin{mybody}
        \textbf{Instruction:} How many women are in the image holding teddy bears?
        \end{mybody}
        \begin{mybody}
        \textbf{Chosen Response:} 
        There are \textcolor{darkblue}{three} women in the image holding teddy bears.
        \end{mybody}
        \begin{mybody}
        \textbf{Rejected Response:} There are \textcolor{darkgreen}{two} women in the image holding teddy bears.
        \end{mybody}
    \caption{Task requiring counting ability.}
\end{subfigure}
\hfill
\begin{subfigure}[b]{0.48\textwidth}
    \fontsize{9.0pt}{\baselineskip}\selectfont
    \linespread{0.9}\selectfont
        \begin{mybody}
        \centering
        \includegraphics[width=0.7\textwidth,height=0.73\textwidth]{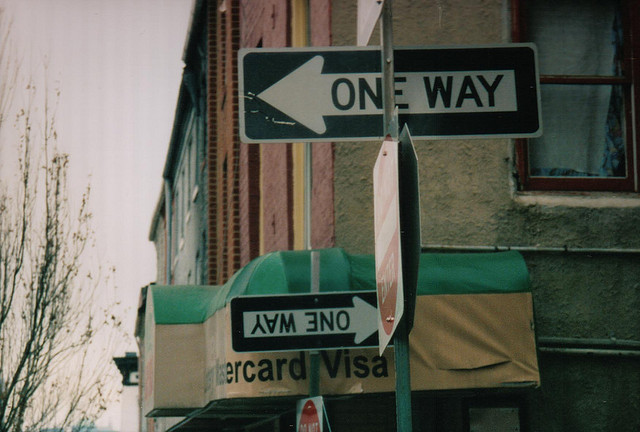}
        \end{mybody}
        \begin{mybody}    
        \textbf{Instruction:} How many road signs are there in the image?
        \end{mybody}
        \begin{mybody}  
        \textbf{Chosen Response:} There are \textcolor{darkblue}{several} road signs in the image, \textcolor{darkblue}{including two One Way signs, one of which is upside down}.
        \end{mybody}
        \begin{mybody}
        \textbf{Rejected Response:} There are \textcolor{darkgreen}{no} road signs in the image, \textcolor{darkgreen}{and the streets are empty of any directives}.
        \end{mybody}
    \caption{Task requiring counting capability.}
\end{subfigure}
\caption{Qualitative examples of the counting task category.}
\label{fig:counting_ex}
\end{figure*}

\clearpage

\subsubsection{Spatial Reasoning}
\label{sec:spatial_datagen_prompts_appendix}

\begin{table*}[h!]
\fontsize{9.0pt}{\baselineskip}\selectfont
\linespread{0.9}\selectfont
%%%%%%%%%%%%%%%%%%%%%%%%%%%%%%%%%%%%%%%%%%%%%%%%%%%%%%%%%
\centering
\includegraphics[scale=0.4]{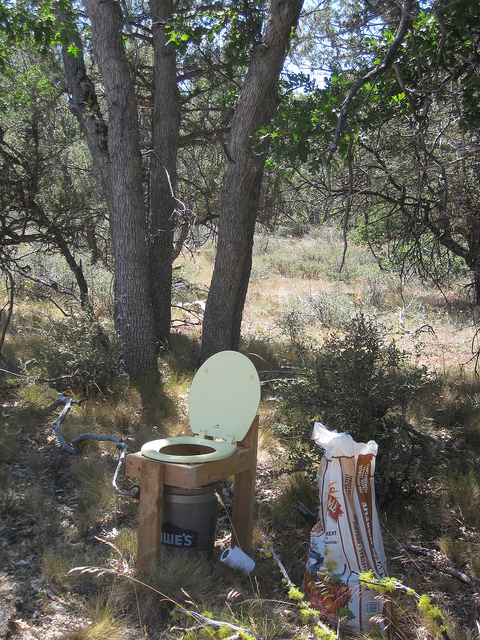}
%%%%%%%%%%%%%%%%%%%%%%%%%%%%%%%%%%%%%%%%%%%%%%%%%%%%%%%%%
\begin{mybody}
\textbf{Prompt:} \\ 
You are a ResponseEditorGPT who is given an instruction and a response generated by a vision-language model. The instruction asks about the "spatial relation", of objects. Your task is as follows: \\ 

1. Modify the "spatial relation" of objects to make the "Original Response" about the "spatial relation" incorrect. \\

2. "New Response" must be linguistically very similar to "Original Response" and must be incorrect.\\

3. You must ensure changes must be realistic given world knowledge. \\

4. You can minimally change other spans of the sentence to grammatical correctness and fluency.\\

5. The output format should be "New Response:" \\

\textbf{Instruction:} Where is the toilet located in the image? \\

\textbf{Original Response:} The toilet is located outdoors, surrounded by a field of grass and trees, in the middle of the woods. \\

Your Turn  \\
New Response: 
\end{mybody}
%%%%%%%%%%%%%%%%%%%%%%%%%%%%%%%%%%%%%%%%%%%%%%%%%%%%%%%%%
\begin{mybody}
\textbf{Chosen Response:} 
The toilet is located \textcolor{darkblue}{outdoors}, surrounded by \textcolor{darkblue}{a field of grass and trees, in the middle of the woods}.
\end{mybody}

\begin{mybody}
\textbf{Hard-negative Rejected Response:} The toilet is located \textcolor{darkgreen}{indoors}, surrounded by \textcolor{darkgreen}{white-tiled walls and a sink, in the corner of a bathroom}.
\end{mybody}
%%%%%%%%%%%%%%%%%%%%%%%%%%%%%%%%%%%%%%%%%%%%%%%%%%%%%%%%%
\captionof{figure}{Example prompt and generated hard-negative for \name sample of spatial reasoning task category.}
\label{fig:spatial_prompt_ex}
\end{table*}
\clearpage

\begin{figure*}[h!]
\centering
\begin{subfigure}[b]{0.48\textwidth}
    \fontsize{9.0pt}{\baselineskip}\selectfont
    \linespread{0.9}\selectfont

    \begin{mybody}
        \centering
        \includegraphics[width=0.73\textwidth,height=0.73\textwidth]{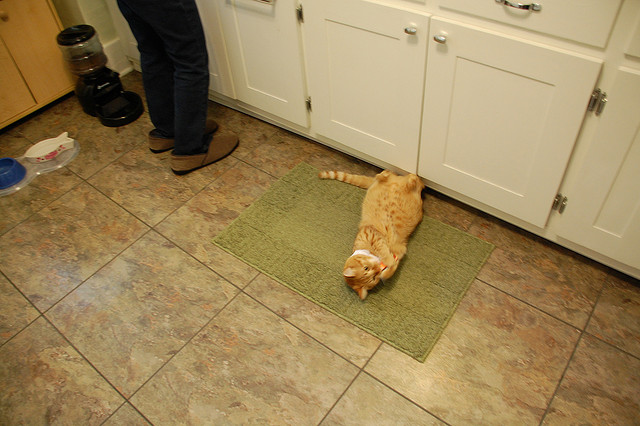}
        \end{mybody}
        \begin{mybody}    
        \textbf{Instruction:} What position is the cat in while laying on the rug?
        \end{mybody}
        \begin{mybody}  
        \textbf{Chosen Response:} The cat is laying on its \textcolor{darkblue}{back} on the rug.
        \end{mybody}
        \begin{mybody}
        \textbf{Rejected Response:} The cat is laying on its \textcolor{darkgreen}{side} on the rug.
        \end{mybody}
    \caption{Task requiring instance location.}

\end{subfigure}
\hfill
\begin{subfigure}[b]{0.48\textwidth}
    \fontsize{9.0pt}{\baselineskip}\selectfont
    \linespread{0.9}\selectfont
        \begin{mybody}
        \centering
        % 000000488658
        \includegraphics[width=0.6\textwidth,height=0.64\textwidth]{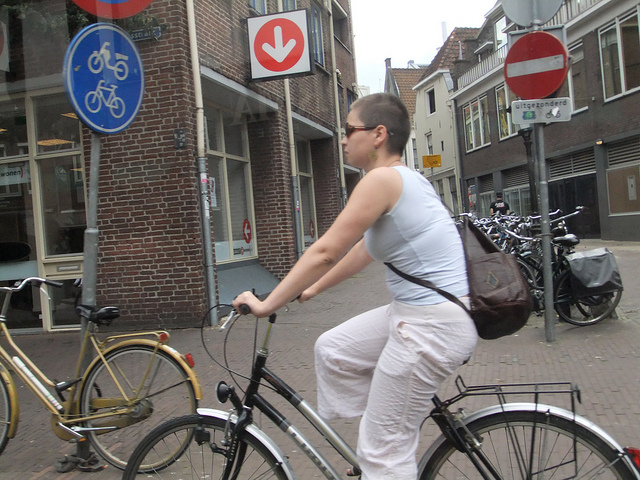}
        \end{mybody}
        \begin{mybody}
        \textbf{Instruction:} Where is the person riding the bicycle in the image?
        \end{mybody}
        \begin{mybody}
        \textbf{Chosen Response:} 
        The person is riding the bicycle on a \textcolor{darkblue}{city street, specifically in a bicycle lane near many} street signs.
        \end{mybody}
        \begin{mybody}
        \textbf{Rejected Response:} The person is riding the bicycle on a \textcolor{darkgreen}{dirt path in a park, far away from any} street signs.
        \end{mybody}
    \caption{Task requiring instance location.}
\end{subfigure}

\vspace{0.25cm}
\begin{subfigure}[b]{0.48\textwidth}
    \fontsize{9.0pt}{\baselineskip}\selectfont
    \linespread{0.9}\selectfont
        \begin{mybody}
        \centering
        \includegraphics[width=0.57\textwidth,height=0.7\textwidth]{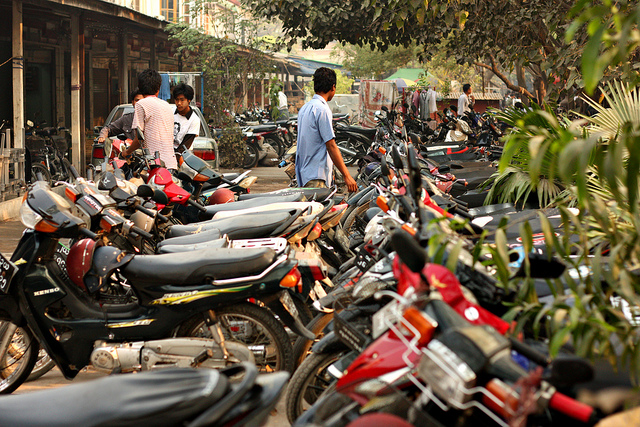}
        \end{mybody}
        \begin{mybody}
        \textbf{Instruction:} How are the motorcycles arranged in the image? 
        \end{mybody}
        \begin{mybody}
        \textbf{Chosen Response:} 
        The motorcycles are arranged in \textcolor{darkblue}{rows or parked together in a row}, which creates an organized and \textcolor{darkblue}{neat} appearance.
        \end{mybody}
        \begin{mybody}
        \textbf{Rejected Response:} The motorcycles are arranged in a \textcolor{darkgreen}{circle, with each facing outwards}, which creates a \textcolor{darkgreen}{symmetrical} and organized appearance.
        \end{mybody}
    \caption{Task requiring relative spatial reasoning.}
\end{subfigure}
\hfill
\begin{subfigure}[b]{0.48\textwidth}
    \fontsize{9.0pt}{\baselineskip}\selectfont
    \linespread{0.9}\selectfont
        \begin{mybody}
        \centering
        \includegraphics[width=0.69\textwidth,height=0.69\textwidth]{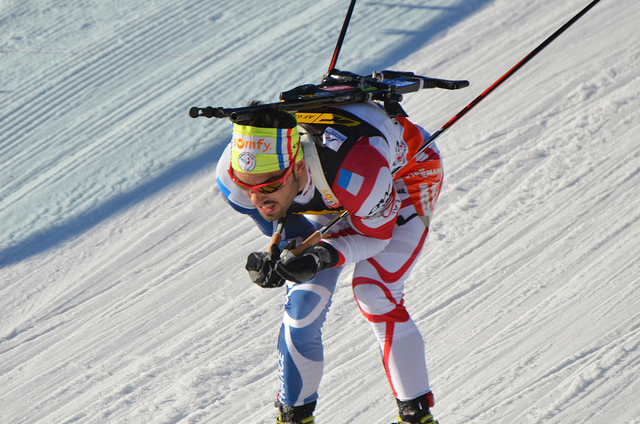}
        \end{mybody}
        \begin{mybody}
        \textbf{Instruction:} What is the position of the skier's ski poles?
        \end{mybody}
        \begin{mybody}
        \textbf{Chosen Response:} 
        The skier \textcolor{darkblue}{has tucked} his ski poles \textcolor{darkblue}{under his arms while racing} through the snow.
        \end{mybody}
        \begin{mybody}
        \textbf{Rejected Response:} The skier \textcolor{darkgreen}{is holding} his ski poles \textcolor{darkgreen}{parallel on either side, with each pole pointing outward from his body as he navigates} through the snow.
        \end{mybody}
    \caption{Task requiring relative spatial reasoning.}
\end{subfigure}
\caption{Qualitative examples of the spatial reasoning task category, involve instance location or relative spatial reasoning.}
\label{fig:spatial_ex}
\end{figure*}

\clearpage

\subsubsection{Existence}
\label{sec:existence_datagen_prompts_appendix}

\begin{table*}[h!]
\fontsize{9.0pt}{\baselineskip}\selectfont
\linespread{0.9}\selectfont
%%%%%%%%%%%%%%%%%%%%%%%%%%%%%%%%%%%%%%%%%%%%%%%%%%%%%%%%%
\centering
\includegraphics[scale=0.35]{images/prompt_images/existence/existence_prompt_ex.jpg}
%%%%%%%%%%%%%%%%%%%%%%%%%%%%%%%%%%%%%%%%%%%%%%%%%%%%%%%%%
\begin{mybody}
\textbf{Prompt:} \\ 

You are a ResponseEditorGPT who is given an instruction and a response generated by a vision-language model. The instruction asks about an "existence" of an object, object attribute, object count, object spatial relation, object comparison, background or theme. Your task is as follows: \\

1. Modify the original response to change the polarity of the response, that is, make "Yes" a "No" and "No" a "Yes". \\

2. Paraphrase both the "Original Response" and the "New Response", such that it says, "Yes" or "No" followed by the ask in the question. \\

3. "New Response" must be linguistically very similar to "Original Response" and must be incorrect.\\

4. You must ensure changes must be realistic given world knowledge. \\

5. You can minimally change other spans of the sentence to grammatical correctness and fluency. \\

6. The output format should be 
"Original Response: "
"New Response \\

\textbf{Instruction:} Would this man score a touchdown? \\

\textbf{Original Response:} No, the man would not score a touchdown. \\

Your Turn  \\
New Response:
\end{mybody}
%%%%%%%%%%%%%%%%%%%%%%%%%%%%%%%%%%%%%%%%%%%%%%%%%%%%%%%%%
\begin{mybody}
\textbf{Chosen Response:} 
\textcolor{darkblue}{No}, the man \textcolor{darkblue}{would not} score a touchdown..
\end{mybody}

\begin{mybody}
\textbf{Hard-negative Rejected Response:} \textcolor{darkgreen}{Yes}, the man \textcolor{darkgreen}{would score} a touchdown.
\end{mybody}
%%%%%%%%%%%%%%%%%%%%%%%%%%%%%%%%%%%%%%%%%%%%%%%%%%%%%%%%%
\captionof{figure}{Example prompt and generated hard-negative for \name sample of object task category.}
\label{fig:existence_prompt_ex}
\end{table*}
\clearpage

\begin{figure*}[h!]
\centering
\begin{subfigure}[b]{0.48\textwidth}
    \fontsize{9.0pt}{\baselineskip}\selectfont
    \linespread{0.9}\selectfont
    \begin{mybody}
        \centering
        \includegraphics[width=0.7\textwidth,height=0.9\textwidth]{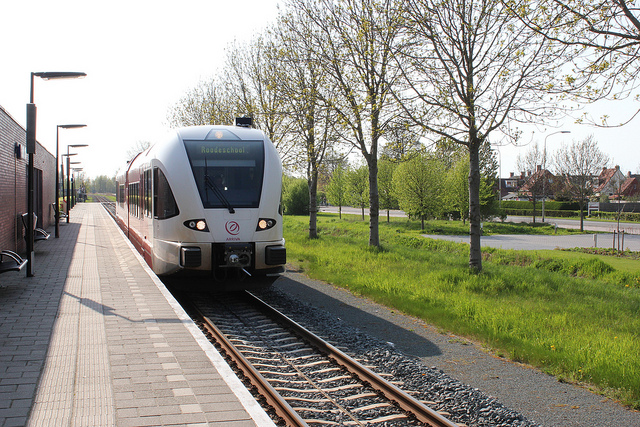}
        \end{mybody}
        \begin{mybody}
        \textbf{Instruction:} Are there any people in the picture? 
        \end{mybody}
        \begin{mybody}
        \textbf{Chosen Response:} 
        \textcolor{darkblue}{No}, there are \textcolor{darkblue}{no people} shown in the picture.  
        \end{mybody}
        \begin{mybody}
        \textbf{Rejected Response:} \textcolor{darkgreen}{Yes}, there are \textcolor{darkgreen}{people visible} in the image.
        \end{mybody}
    \caption{Task requiring object perception}
\end{subfigure}
\hfill
\begin{subfigure}[b]{0.48\textwidth}
    \fontsize{9.0pt}{\baselineskip}\selectfont
    \linespread{0.9}\selectfont
        \begin{mybody}
        \centering
        \includegraphics[width=0.7\textwidth,height=0.9\textwidth]{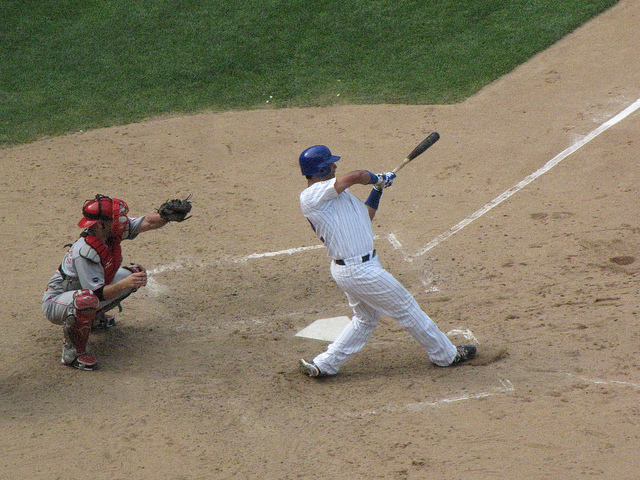}
        \end{mybody}
        \begin{mybody}
        \textbf{Instruction:} Are the batters ankles twisted?
        \end{mybody}
        \begin{mybody}
        \textbf{Chosen Response:} 
        \textcolor{darkblue}{Yes}, the batter's ankles \textcolor{darkblue}{are twisted}.
        \end{mybody}
        \begin{mybody}
        \textbf{Rejected Response:} \textcolor{darkgreen}{No}, the batter's ankles \textcolor{darkgreen}{are not twisted}.
        \end{mybody}
    \caption{Task requiring perception of an object's state and action}
\end{subfigure}

\vspace{0.25cm}
\begin{subfigure}[b]{0.48\textwidth}
    \fontsize{9.0pt}{\baselineskip}\selectfont
    \linespread{0.9}\selectfont
        \begin{mybody}
        \centering
        \includegraphics[width=0.65\textwidth,height=0.65\textwidth]{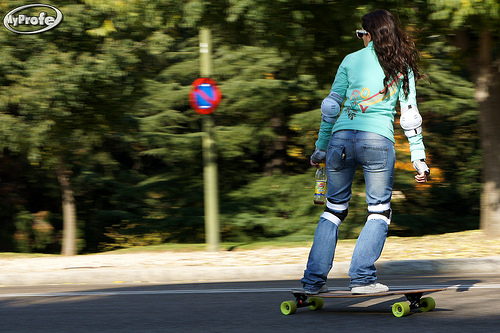}
        \end{mybody}
        \begin{mybody}
        \textbf{Instruction:} Does the elbow pad to the right of the other elbow pad have black color?
        \end{mybody}
        \begin{mybody}
        \textbf{Chosen Response:} 
        \textcolor{darkblue}{No}, the elbow pad to the right of the other elbow pad \textcolor{darkblue}{does not have} a black color.
        \end{mybody}
        \begin{mybody}
        \textbf{Rejected Response:} \textcolor{darkgreen}{Yes}, the elbow pad to the right of the other elbow pad \textcolor{darkgreen}{has a} black color.
        \end{mybody}
    \caption{Task requiring color perception spatial reasoning}
\end{subfigure}
\hfill
\begin{subfigure}[b]{0.48\textwidth}
    \fontsize{9.0pt}{\baselineskip}\selectfont
    \linespread{0.9}\selectfont
        \begin{mybody}
        \centering
        \includegraphics[width=0.8\textwidth,height=0.8\textwidth]{images/prompt_images/existence/existence_ex4.jpg}
        \end{mybody}
        \begin{mybody}    
        \textbf{Instruction:} Does the light say it is ok to walk?
        \end{mybody}
        \begin{mybody}  
        \textbf{Chosen Response:} \textcolor{darkblue}{Yes}, the light \textcolor{darkblue}{indicates it's safe} to walk.
        \end{mybody}
        \begin{mybody}
        \textbf{Rejected Response:} \textcolor{darkgreen}{No}, the light \textcolor{darkgreen}{says it's not safe} to walk.
        \end{mybody}
    \caption{Task requiring object perception and knowledge}
\end{subfigure}
\caption{Qualitative examples of the existence task category.}
\label{fig:existence_ex}
\end{figure*}

\clearpage

\subsubsection{Referential VQA}
\label{sec:refvqa_datagen_prompts_appendix}

\begin{table*}[h!]
\fontsize{9.0pt}{\baselineskip}\selectfont
\linespread{0.9}\selectfont
%%%%%%%%%%%%%%%%%%%%%%%%%%%%%%%%%%%%%%%%%%%%%%%%%%%%%%%%%
\centering
\includegraphics[scale=0.25]{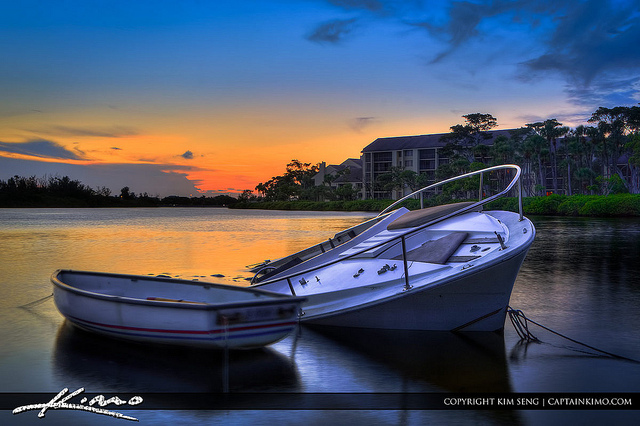}
%%%%%%%%%%%%%%%%%%%%%%%%%%%%%%%%%%%%%%%%%%%%%%%%%%%%%%%%%
\begin{mybody}
\textbf{Prompt:} \\ 

You are a ResponseEditorGPT who is given an instruction and a response generated by a vision-language model. The instruction asks about counts, color, spatial location, comparison or existence of objects. More than one of the tasks can be asked in an instruction. Your task is as follows: \\

1. Identify the different tasks asked in the question. You do not have to output this, only understand the intent. \\

2. Modify the spans in the response which answer the different tasks in the instruction to make "Original Response" incorrect. \\

3. You must only change the "spans", so that "New Response" is linguistically very similar to "Original Response" and is incorrect, while maintaining rest of the response. \\

4. You can minimally change other spans of the sentence to semantic correctness, grammatical correctness and fluency. \\

5. If the task is about colors or counts, ensure that you change the span with wide range of colors and counts respectively. \\

6. The "New colors" or "New Counts" must be realistically possible, considering the object they describe. \\

7. If the response is one word or small phrase, paraphrase both the "Original Response" and the "New Response", such that it says "New Response" is incorrect with respect to the "Original Response" while being semantically sensible. Both "Original Response" and "New Response" must now be full sentences. \\

\textbf{Instruction:} What is the size difference between these two boats? \\

\textbf{Original Response:} There is a noticeable size difference between the two boats, with one being considerably larger than the other smaller boat. \\

Your Turn  \\
New Response:
\end{mybody} 
%%%%%%%%%%%%%%%%%%%%%%%%%%%%%%%%%%%%%%%%%%%%%%%%%%%%%%%%%
\begin{mybody}
\textbf{Chosen Response:} 
There is \textcolor{darkblue}{a noticeable} size difference between the two boats, \textcolor{darkblue}{with one being considerably larger than the other smaller boat}.
\end{mybody}

\begin{mybody}
\textbf{Hard-negative Rejected Response:} There is \textcolor{darkgreen}{no noticeable} size difference between the two boats, \textcolor{darkgreen}{as they appear to be identical in size}.
\end{mybody}
%%%%%%%%%%%%%%%%%%%%%%%%%%%%%%%%%%%%%%%%%%%%%%%%%%%%%%%%%
\captionof{figure}{Example prompt and generated hard-negative for \name sample of referential VQA task category.}
\label{fig:ref_vqa_prompt_ex}
\end{table*}

\clearpage

\begin{figure*}[h!]
\centering
\begin{subfigure}[b]{0.48\textwidth}
    \fontsize{9.0pt}{\baselineskip}\selectfont
    \linespread{0.9}\selectfont
    \begin{mybody}
        \centering
        \includegraphics[width=0.6\textwidth,height=0.6\textwidth]{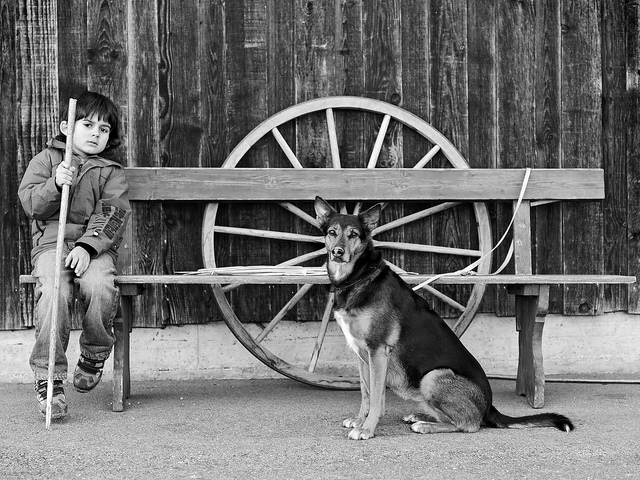}
        \end{mybody}
        \begin{mybody}
        \textbf{Instruction:} What is the size of the dog compared to the child? 
        \end{mybody}
        \begin{mybody}
        \textbf{Chosen Response:} 
        The dog is described as \textcolor{darkblue}{large} in comparison to the child. This implies that the dog may be of a \textcolor{darkblue}{bigger breed or perhaps a fully-grown adult dog}.
        \end{mybody}
        \begin{mybody}
        \textbf{Rejected Response:} The dog is described as \textcolor{darkgreen}{small} in comparison to the child. This implies that \textcolor{darkgreen}{the dog may be a smaller breed or perhaps still a puppy}.
        \end{mybody}
    \caption{Task requiring object perception}
\end{subfigure}
\hfill
\begin{subfigure}[b]{0.48\textwidth}
    \fontsize{9.0pt}{\baselineskip}\selectfont
    \linespread{0.9}\selectfont
        \begin{mybody}
        \centering
        \includegraphics[width=0.7\textwidth,height=0.9\textwidth]{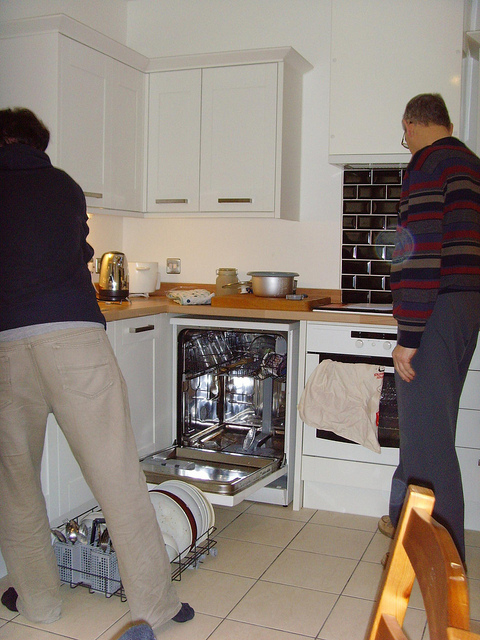}
        \end{mybody}
        \begin{mybody}
        \textbf{Instruction:} How many people are in the image and where are they located?
        \end{mybody}
        \begin{mybody}
        \textbf{Chosen Response:} 
       There are \textcolor{darkblue}{two} people, \textcolor{darkblue}{a} man and \textcolor{darkblue}{a} woman, in the image, and they are located in a \textcolor{darkblue}{kitchen}.
        \end{mybody}
        \begin{mybody}
        \textbf{Rejected Response:} There are \textcolor{darkgreen}{five} people, \textcolor{darkgreen}{consisting of  three} men and \textcolor{darkgreen}{two} women, in the image, and they are located in a \textcolor{darkgreen}{living room}.
        \end{mybody}
    \caption{Task requiring perception of an object's state and action}
\end{subfigure}

\vspace{0.25cm}
\begin{subfigure}[b]{0.48\textwidth}
    \fontsize{9.0pt}{\baselineskip}\selectfont
    \linespread{0.9}\selectfont
        \begin{mybody}
        \centering
        \includegraphics[width=0.65\textwidth,height=0.65\textwidth]{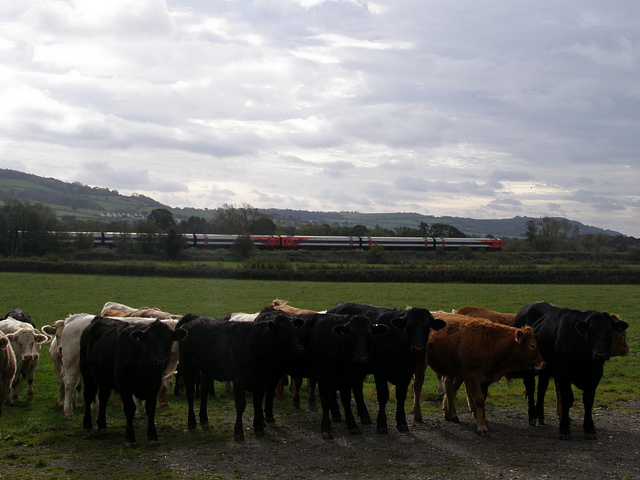}
        \end{mybody}
        \begin{mybody}
        \textbf{Instruction:} How many colors are the cows in the image? 
        \end{mybody}
        \begin{mybody}
        \textbf{Chosen Response:} 
         There are \textcolor{darkblue}{three} main colors of cows in the image: black, \textcolor{darkblue}{brown}, and white.
        \end{mybody}
        \begin{mybody}
        \textbf{Rejected Response:} There are \textcolor{darkgreen}{four} main colors of cows in the image: black, \textcolor{darkgreen}{gray, tan}, and white.
        \end{mybody}
    \caption{Task requiring color perception spatial reasoning}
\end{subfigure}
\hfill
\begin{subfigure}[b]{0.48\textwidth}
    \fontsize{9.0pt}{\baselineskip}\selectfont
    \linespread{0.9}\selectfont
        \begin{mybody}
        \centering
        \includegraphics[width=0.8\textwidth,height=0.8\textwidth]{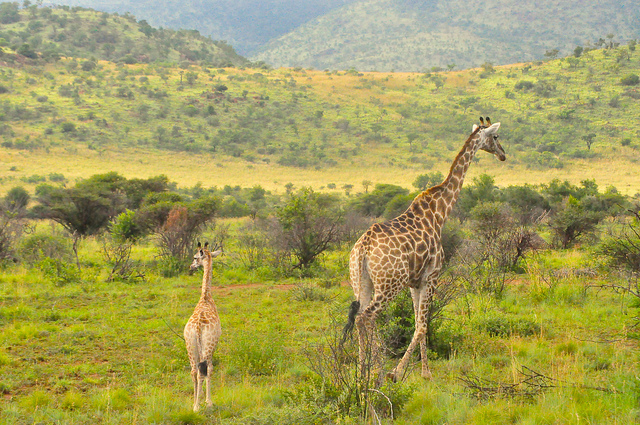}
        \end{mybody}
        \begin{mybody}    
        \textbf{Instruction:} How many giraffes are visible in the image, and what are their relative sizes? %000000018886
        \end{mybody}
        \begin{mybody}  
        \textbf{Chosen Response:} There are \textcolor{darkblue}{two} giraffes visible in the image: a \textcolor{darkblue}{large}  adult giraffe and a \textcolor{darkblue}{smaller kid} giraffe, likely its offspring.
        \end{mybody}
        \begin{mybody}
        \textbf{Rejected Response:} There are \textcolor{darkgreen}{three} giraffes visible in the image: a \textcolor{darkgreen}{medium-sized} adult giraffe \textcolor{darkgreen}{along with two smaller} giraffes, possibly its offspring.
        \end{mybody}
    \caption{Task requiring object perception and knowledge}
\end{subfigure}
\caption{Qualitative examples of the existence task category.}
\label{fig:ref_vqa_ex}
\end{figure*}

\clearpage

\subsubsection{Image Captioning}
\label{sec:captioning_datagen_prompts_appendix}

\begin{table*}[h!]
\fontsize{9.0pt}{\baselineskip}\selectfont
\linespread{0.9}\selectfont
%%%%%%%%%%%%%%%%%%%%%%%%%%%%%%%%%%%%%%%%%%%%%%%%%%%%%%%%%
\centering
\includegraphics[scale=1.]{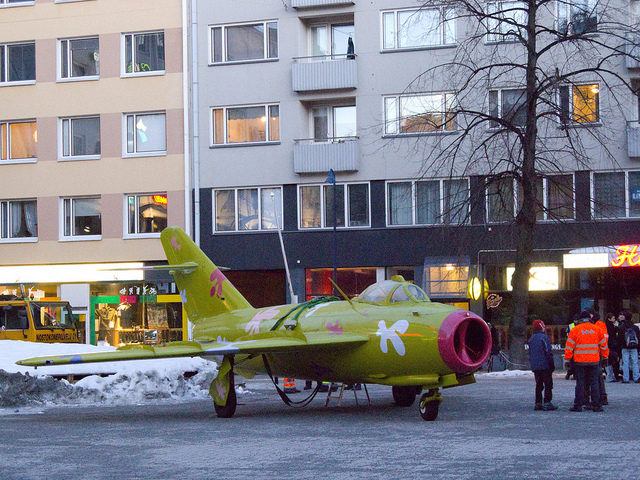}
%%%%%%%%%%%%%%%%%%%%%%%%%%%%%%%%%%%%%%%%%%%%%%%%%%%%%%%%%
\begin{mybody}
\textbf{Prompt:} \\ 

You are a ResponseAnalyzerGPT who is given an instruction and a response generated by a vision-language model. The response will consist of one or more visual elements - objects, object relationships, object attributes, environment information or actions. Each visual element is modified by one or more dimension, where a dimension belongs to the set ("Color", "Number", "Size", "Shape", "Other object physical attribute", "Weather Time", "Background", "Spatial relationship", "Comparative relationship", "Other object relationship")

Your task is to list (visual element, dimension, phrase) triplet, where visual element is an element in the response, dimension modifies the visual element and phrase is a span from the response that shows how dimension modified the visual element. \\

You must follow the guidelines given below: \\

1. Do not repeat the same triplet. \\

2. The dimension must always belong to the set ("Color", "Number", "Size", "Shape", "Other object physical attribute", "Weather Time", "Background", "Spatial relationship", "Comparative relationship", "Other object relationship") \\

3. Output format Must be "Triplet List : []" where "[]" is a list of triplets \\

\textbf{Instruction:} Write a detailed description of the given image. \\

\textbf{Original Response:} The image features a unique scene of a green jetfighter airplane on display in an open area of the city. The airplane has white and pink accents painted on its design, making it visually striking. It is situated in the middle of the road, with tall buildings surrounding the area. \\

There are several people admiring and standing next to the green plane. Some of them can be found on the right side of the image, while another person is seen closer to the plane on the left side. In addition to the airplane, there is a truck parked nearby on the left side of the road.\\

Your Turn  \\
Triplet List: [] 
\end{mybody} 
%%%%%%%%%%%%%%%%%%%%%%%%%%%%%%%%%%%%%%%%%%%%%%%%%%%%%%%%%
\begin{mybody}
\textbf{Triplet List}: [("jetfighter airplane", "color", "green jetfighter airplane"), ("airplane", "color", " white and pink accent"), ("it", "spatial relationship", "in the middle of the road"), ("buildings", "size", "tall buildings"),("people", "counting", "several people"),("some", "spatial relationship", " right side of the
image"),("people", "spatial relationship", "standing next to the green plan"),
("person", "spatial relationship", " closer to the plane on the left side"), ("truck", "spatial relationship", "  truck parked nearby on the left side of the road")]
\end{mybody}
%%%%%%%%%%%%%%%%%%%%%%%%%%%%%%%%%%%%%%%%%%%%%%%%%%%%%%%%%
\captionof{figure}{Example prompt for extracting the different possible dimensions of perturbing an image caption (perception \& reasoning). Note, we simulate the dimension extraction in this example.}
\label{fig:caption_prompt_ex1}
\end{table*}
\clearpage

\begin{table*}[h!]
\fontsize{9.0pt}{\baselineskip}\selectfont
\linespread{0.9}\selectfont
%%%%%%%%%%%%%%%%%%%%%%%%%%%%%%%%%%%%%%%%%%%%%%%%%%%%%%%%%
\centering
%%%%%%%%%%%%%%%%%%%%%%%%%%%%%%%%%%%%%%%%%%%%%%%%%%%%%%%%%
\begin{mybody}
\textbf{Prompt:} \\ 
You are a ResponseEditorGPT who is given an instruction and a response generated by a vision-language model. The response will consist of one or more visual elements - objects, object relationships, object attributes, environment information or actions. Each visual element is modified by one or more dimension, where a dimension must belong to the set ("Color", "Number", "Size", "Shape", "Other object physical attribute", "Weather Time", "Background", "Spatial relationship", "Comparative relationship", "Other object relationship"). You will also be given a list of (visual element, dimension, phrase) triplets, where visual element is an element in the response, dimension modifies the visual element and phrase is a span from the response that shows how dimension modified the visual element. \\

Your task is as follows:  \\

1. For each triplet, modify the phrase in the "Original response" corresponding to each triplet along the dimension mentioned in the triplet to make the "Original response" incorrect. \\

2. "New Response" must be linguistically very similar to "Original response" and must be incorrect. \\

3. You must ensure changes must be realistic given world knowledge. \\

4. You can minimally change other spans of the sentence to grammatical correctness and fluency. \\

5. The output format should be 
"Original Response: "
"New Response \\

\textbf{Instruction:} Write a detailed description of the given image. \\

\textbf{Original Response:} The image features a unique scene of a green jetfighter airplane on display in an open area of the city. The airplane has white and pink accents painted on its design, making it visually striking. It is situated in the middle of the road, with tall buildings surrounding the area. \\

There are several people admiring and standing next to the green plane. Some of them can be found on the right side of the image, while another person is seen closer to the plane on the left side. In addition to the airplane, there is a truck parked nearby on the left side of the road.\\

Triplet List:  [("jetfighter airplane", "color", "green jetfighter airplane"), ("airplane", "color", " white and pink accent"), ("it", "spatial relationship", "in the middle of the road"), ("buildings", "size", "tall buildings"),("people", "spatial relationship", "standing next to the green plan"),
("person", "spatial relationship", " closer to the plane on the left side"), ("truck", "spatial relationship", "  truck parked nearby on the left side of the road")]\\
Your Turn  \\
New Response:
\end{mybody} 
%%%%%%%%%%%%%%%%%%%%%%%%%%%%%%%%%%%%%%%%%%%%%%%%%%%%%%%%%
\begin{mybody}
\textbf{Chosen Response:} 
        The image features a unique scene of a \textcolor{darkblue}{green} jetfighter airplane on display in an open area of the city. The airplane has \textcolor{darkblue}{white} and \textcolor{darkblue}{pink} accents painted on its design, making it visually striking. It is situated \textcolor{darkblue}{in the middle of} the road, with \textcolor{darkblue}{tall} buildings surrounding the area. \\

        There are several people admiring and \textcolor{darkblue}{standing next to the green plane}. Some of them can be found on the right side of the image, while another \textcolor{darkblue}{person is seen closer to the plane on the left side}. In addition to the airplane, there is a truck parked \textcolor{darkblue}{nearby on the left side of the road.} 
\end{mybody}

\begin{mybody}
\textbf{Hard-negative Rejected Response:} The image features a unique scene of a \textcolor{darkgreen}{yellow} jetfighter airplane on display in an open area of the city. The airplane has \textcolor{darkgreen}{purple} and \textcolor{darkgreen}{orange accents} painted on its design, making it visually striking. It is situated \textcolor{darkgreen}{on the side of} the road, with \textcolor{darkgreen}{short} buildings surrounding the area. \\

        There are several people admiring and standing \textcolor{darkgreen}{in the distance from the yellow plane}. Some of them can be found on the right side of the image, while another person is \textcolor{darkgreen}{seen farther from the plane on the left side}. In addition to the airplane, there is a truck \textcolor{darkgreen}{parked far from the right side of the road.} 
\end{mybody}
%%%%%%%%%%%%%%%%%%%%%%%%%%%%%%%%%%%%%%%%%%%%%%%%%%%%%%%%%
\captionof{figure}{Example prompt for generating hard-negative rejected response for image captioning category. We randomly select 50\% to 75\% triplets and feed them to the prompt as shown above. Similar to colors and counting task categories, we maintain a list of most perturbed dimensions (size 2), updated after every 10 hard-negative rejected response generations. If a dimension is used more than once, its frequency is proportionately increased. Note image is the same as the previous prompt (see Fig.~\ref{fig:caption_prompt_ex1}).}
\label{fig:caption_prompt_ex2}
\end{table*}
\clearpage

\begin{figure*}[h!]
\centering
\begin{subfigure}[b]{\textwidth}
    \fontsize{9.0pt}{\baselineskip}\selectfont
    \linespread{0.9}\selectfont
    \begin{mybody}
        \centering
        \includegraphics[width=0.5\textwidth,height=0.5\textwidth]{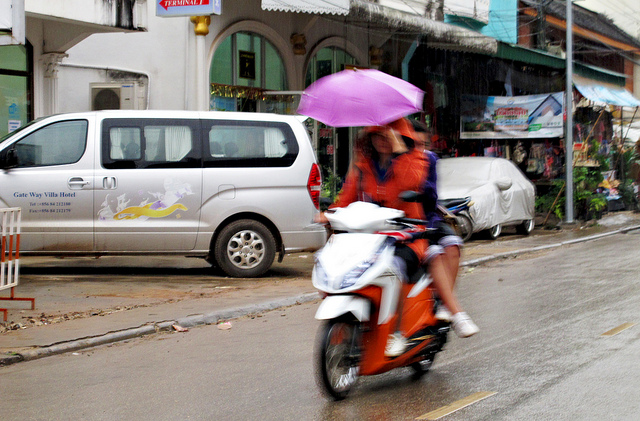}
        \end{mybody}
        \begin{mybody}
        \textbf{Instruction:} Analyze the image in a comprehensive and detailed manner.
        \end{mybody}
        \begin{mybody}
        \textbf{Chosen Response:} 
        The image presents \textcolor{darkblue}{an orange} and white motor scooter with two passengers riding down the middle of a \textcolor{darkblue}{wet} street. The person in the front is driving the scooter, while the person sitting behind them is holding a \textcolor{darkblue}{purple} umbrella over both of them, providing protection from \textcolor{darkblue}{rain}. \\

        There are potted plants on the sidewalk, adding some greenery to the scene. Cars can be seen parked or driving in the background, sharing the road with the scooter. The scene depicts a typical \textcolor{darkblue}{rainy} day on \textcolor{darkblue}{an urban} street with vehicles and pedestrians going about their daily routine. \\
        \end{mybody}
        \begin{mybody}
        \textbf{Rejected Response:} 

        The image presents \textcolor{darkgreen}{a black} and white motor scooter with two passengers riding down the middle of a \textcolor{darkgreen}{dry} street. The person in the front is driving the scooter, while the person sitting behind them is holding a \textcolor{darkgreen}{red} umbrella over both of them, providing protection from \textcolor{darkgreen}{the sun}. \\

        There are potted plants on the sidewalk, adding some greenery to the scene. Cars can be seen parked or driving in the background, sharing the road with the scooter. The scene depicts a typical \textcolor{darkgreen}{sunny} day on \textcolor{darkgreen}{a suburban street} with vehicles and pedestrians going about their daily routine. \\
        \end{mybody}
    \caption{Task requiring color, object, and background perception.}
\end{subfigure}
\caption{Qualitative example of the image captioning task category.}
\label{fig:caption_ex}
\end{figure*}

\clearpage

\subsubsection{General Reasoning}
\label{sec:genreasoning_datagen_prompts_appendix}

\begin{table*}[h!]
\fontsize{9.0pt}{\baselineskip}\selectfont
\linespread{0.9}\selectfont
%%%%%%%%%%%%%%%%%%%%%%%%%%%%%%%%%%%%%%%%%%%%%%%%%%%%%%%%%
\centering
\includegraphics[scale=0.25]{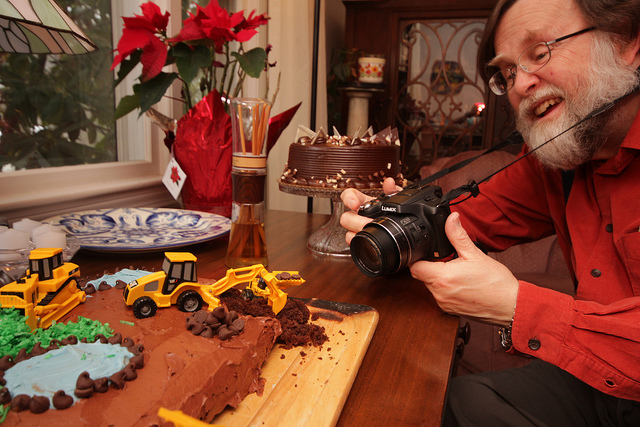}
%%%%%%%%%%%%%%%%%%%%%%%%%%%%%%%%%%%%%%%%%%%%%%%%%%%%%%%%%
\begin{mybody}
\textbf{Prompt:} \\ 

You are a ResponseEditorGPT who is given an instruction and a response generated by a vision-language model. The instruction asks about the "reasoning" of objects, events, environments, or themes. Your task is as follows:  \\

1. Make the reasoning in the original response incorrect.\\

2. You can modify the objects, their attributes, related objected, or action and make the "original response" about "reasoning" is incorrect. \\

3. "New Response" must be linguistically very similar to "Original response" and must be incorrect. \\

4. You must ensure changes must be realistic given world knowledge. \\

5. You can minimally change other spans of the sentence to grammatical correctness and fluency. \\

6. The output format should be 
"Original Response: "
"New Response \\

\textbf{Instruction:} What might be a possible reason for the man taking a picture of the dirt cake? \\

\textbf{Original Response:}  A possible reason for the man taking a picture of the dirt cake could be that the cake is a unique and creative design, which features a construction scene on top. He might want to capture the design and decoration before it is served or share the photo with others to showcase the artistic and aesthetic aspects of the cake. Additionally, the image could be used as a memory of a special occasion or event for which the cake has been prepared.\\

Your Turn  \\
New Response:
\end{mybody} 
%%%%%%%%%%%%%%%%%%%%%%%%%%%%%%%%%%%%%%%%%%%%%%%%%%%%%%%%%
\begin{mybody}
\textbf{Chosen Response:} 
 A possible reason for the man taking a picture of the dirt cake could be that the cake is a \textcolor{darkblue}{unique} and \textcolor{darkblue}{creative design, which features a construction scene on top}. He might want to capture the design and \textcolor{darkblue}{decoration before it is served or share the photo with others to showcase the artistic and aesthetic aspects of the cake}. Additionally, the image could be used as a memory of a \textcolor{darkblue}{special occasion or event for which the cake has been prepared}.
\end{mybody}

\begin{mybody}
\textbf{Hard-negative Rejected Response:} A possible reason for the man taking a picture of the dirt cake could be that the cake is a \textcolor{darkgreen}{rare} and \textcolor{darkgreen}{delicate design featuring an underwater scene on top}. He might want to capture the design and \textcolor{darkgreen}{the aquatic elements before it is served or share the photo with others to highlight the intricate and sea-themed aspects of the cake.} Additionally, the image could be used as a memory of a \textcolor{darkgreen}{beach-themed event or occasion for which the cake has been prepared.}
\end{mybody}
%%%%%%%%%%%%%%%%%%%%%%%%%%%%%%%%%%%%%%%%%%%%%%%%%%%%%%%%%
\captionof{figure}{Example prompt and generated hard-negative for \name sample of general reasoning task category.}
\label{fig:reason_prompt_ex}
\end{table*}

\clearpage 

\subsubsection{Examples Showing Generation Steps for Task Categories with Penalty List}
\label{sec:penalty_list_examples}
In this section, we use the color task as an example to illustrate generation steps involving the penalty list (counting task will have a similar way of prompting). We use a penalty list of size K=10, updated every 10 samples, with one retry generation attempt possible. Note, until we have atleast K (here =10) distinct colors, the penalty list can be < K. 

\textbf{Stochastic Generation:} Since generation is stochastic, the exact penalty list used for a given sample may vary. Therefore, we simulate several cases:
    \begin{itemize}
        \item \textbf{Case-1: Empty list} When the penalty list is empty.
        \item \textbf{Case-2} When the penalty list is non-empty, with sub-cases:
        \begin{itemize}
            \item \textbf{Case-2a: First Attempt} The generation succeeds on the first attempt without conflicting with the penalty list.
            \item \textbf{Case-2b: Second Attempt} The generation initially conflicts but succeeds on a retry attempt.
            \item \textbf{Case-2c: Rerun Script} If both attempts fail due to conflicts, the sample is put back into the pool. This situation happens neglibible number of times and the solution is to re-run the script. When re-running, it typically encounters a new penalty list because:
            \begin{itemize}
                \item Some samples may have already been processed and help seed the penalty list.
                \item The script randomizes sample order, ensuring that failed samples do not repeatedly see the same penalty list.
            \end{itemize}
        \end{itemize}
    \end{itemize}

For the purpose of illustration, we assume no generation failures by the LLM and K=10, when penalty list is not empty. The \cmark rejected response (new response) is actually a sample from the dataset, and hence is kept as the correct response across scenarios. Now, we provide two examples showcasing different cases:

\begin{table*}[h!]
\fontsize{9.0pt}{\baselineskip}\selectfont
\linespread{0.9}\selectfont
%%%%%%%%%%%%%%%%%%%%%%%%%%%%%%%%%%%%%%%%%%%%%%%%%%%%%%%%%
\centering
\includegraphics[scale=0.6]{images/prompt_images/color_step_examples/000000389392.jpg}
%%%%%%%%%%%%%%%%%%%%%%%%%%%%%%%%%%%%%%%%%%%%%%%%%%%%%%%%%
\begin{mybody}
\textbf{Prompt:} \\ 
You are a ResponseEditorGPT who is given an instruction and a response generated by a vision-language model. The instruction asks about the colors of objects, environments, or themes. Your task is as follows: \\

1. Modify the "colors" of all objects, environments, or themes in the response to make "Original Response" about "colors" incorrect.  \\

2. You must only change the "colors", so that "New Response" is linguistically very similar to "Original Response" and is incorrect.\\

3. The "New colors" you use to replace original colors must be unique and not be too descriptive. \\

4. The "New colors" must be realistically possible, considering the object they describe. \\

5. You cannot use colors in the penalty list. \\

6. You can minimally change other spans of the sentence to grammatical correctness and fluency.\\

7. List the "New colors" you replace within the response. \\

\textbf{Penalty list:} {} \\

\textbf{Instruction:} What colors are present on the subway train in the image? \\

\textbf{Original Response:} The subway train in the image is orange, blue, and silver. \\

Your Turn  \\
New Response: \\
New Colors:
\end{mybody}
\end{table*}

\clearpage

\begin{table*}[h!]
\fontsize{9.0pt}{\baselineskip}\selectfont
\linespread{0.9}\selectfont
\textbf{Original Response:} The subway train in the image is orange, blue, and silver. \\ 
\end{table*}

\begin{table*}[h!]
\fontsize{9.0pt}{\baselineskip}\selectfont
\linespread{0.9}\selectfont
\textbf{Case-1: Empty list} \\ 
\begin{mybody}
\textbf{Penalty List:} []
\end{mybody}
\begin{mybody}
\textbf{New Response:} The subway train in the image is pink, turquoise, and white. \cmark

\textbf{New Colors:} ['pink', 'turquoise', 'white'] \cmark
\end{mybody}
%%%%%%%%%%%%%%%%%%%%%%%%%%%%%%%%%%%%%%%%%%%%%%%%%%%%%%%%%
\end{table*}

\begin{table*}[h!]
\fontsize{9.0pt}{\baselineskip}\selectfont
\linespread{0.9}\selectfont
\textbf{Case-2a: First Attempt} \\ 
\begin{mybody}
\textbf{Penalty List:} ['yellow', 'black', 'beige', 'teal', 'green', 'burgundy', 'sepia', 'lavender', 'purple', 'orange']
\end{mybody}
\begin{mybody}
\textbf{New Response:} The subway train in the image is pink, turquoise, and white. \cmark 

\textbf{New Colors:} ['pink', 'turquoise', 'white'] \cmark
\end{mybody}
\begin{mybody}
\textbf{New Response:} The subway train in the image is pink, black, white. \xmark

\textbf{New Colors:} ['pink', 'black', 'white'] \xmark 

\textbf{Reason:} Color \textit{black} conflicts with penalty list, retry.
\end{mybody}
%%%%%%%%%%%%%%%%%%%%%%%%%%%%%%%%%%%%%%%%%%%%%%%%%%%%%%%%%
\end{table*}

\begin{table*}[h!]
\fontsize{9.0pt}{\baselineskip}\selectfont
\linespread{0.9}\selectfont
\textbf{Case-2b: Second Attempt} \\ 
\begin{mybody}
\textbf{Penalty List:} ['yellow', 'black', 'beige', 'teal', 'green', 'burgundy', 'sepia', 'lavender', 'purple', 'orange']
\end{mybody}
\begin{mybody}
\textbf{New Response:} The subway train in the image is pink, turquoise, and white. \cmark

\textbf{New Colors:} ['pink', 'turquoise', 'white'] \cmark
\end{mybody}
\begin{mybody}
\textbf{New Response:} The subway train in the image is yellow, green, white. \xmark

\textbf{New Colors:} ['yellow', 'green', 'white'] \xmark 

\textbf{Reason:} Colors \textit{yellow \& green} conflict with penalty list, put sample back to un-annotated pool and rerun the script with un-annotated samples.
\end{mybody}
%%%%%%%%%%%%%%%%%%%%%%%%%%%%%%%%%%%%%%%%%%%%%%%%%%%%%%%%%
\end{table*}

\begin{table*}[h!]
\fontsize{9.0pt}{\baselineskip}\selectfont
\linespread{0.9}\selectfont
\textbf{Case-2c: Rerun-script (penalty list updates} \\ 
\begin{mybody}
\textbf{Penalty List:} ['yellow', 'black', 'beige', 'blue', 'green', 'red', 'silver', 'lavender', 'purple', 'orange']
\end{mybody}
\begin{mybody}
\textbf{New Response:} The subway train in the image is pink, turquoise, and white. \cmark

\textbf{New Colors:} ['pink', 'turquoise', 'white'] \cmark
\end{mybody}
\begin{mybody}
\textbf{New Response:} The subway train in the image is red, turquoise, white. \xmark

\textbf{New Colors:} ['red', 'turquoise', 'white'] \xmark 

\textbf{Reason:} Color \textit{red} conflicts with penalty list, go to Case-2b.
\end{mybody}
%%%%%%%%%%%%%%%%%%%%%%%%%%%%%%%%%%%%%%%%%%%%%%%%%%%%%%%%%
\end{table*}

\begin{table*}[h!]
\fontsize{9.0pt}{\baselineskip}\selectfont
\linespread{0.9}\selectfont
%%%%%%%%%%%%%%%%%%%%%%%%%%%%%%%%%%%%%%%%%%%%%%%%%%%%%%%%%
\centering
\includegraphics[scale=0.7]{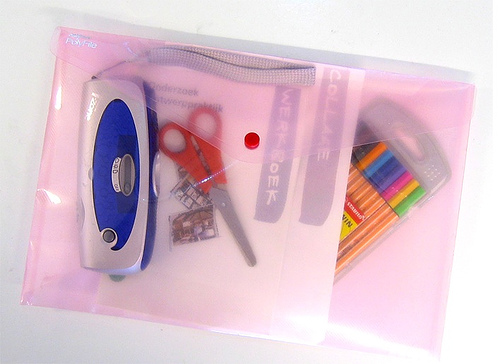}
%%%%%%%%%%%%%%%%%%%%%%%%%%%%%%%%%%%%%%%%%%%%%%%%%%%%%%%%%
\begin{mybody}
\textbf{Prompt:} \\ 
You are a ResponseEditorGPT who is given an instruction and a response generated by a vision-language model. The instruction asks about the colors of objects, environments, or themes. Your task is as follows: \\

1. Modify the "colors" of all objects, environments, or themes in the response to make "Original Response" about "colors" incorrect.  \\

2. You must only change the "colors", so that "New Response" is linguistically very similar to "Original Response" and is incorrect.\\

3. The "New colors" you use to replace original colors must be unique and not be too descriptive. \\

4. The "New colors" must be realistically possible, considering the object they describe. \\

5. You cannot use colors in the penalty list. \\

6. You can minimally change other spans of the sentence to grammatical correctness and fluency.\\

7. List the "New colors" you replace within the response. \\

\textbf{Penalty list:} {} \\

\textbf{Instruction:} What color are the bags or packages containing the office supplies? \\

\textbf{Original Response:} The office supplies are contained in small pink packages or a pink envelope. \\

Your Turn  \\
New Response: \\
New Colors:
\end{mybody}
\end{table*}

\clearpage

\begin{table*}[h!]
\fontsize{9.0pt}{\baselineskip}\selectfont
\linespread{0.9}\selectfont
\textbf{Original Response:} The office supplies are contained in small pink packages or a pink envelope. \\ 
\end{table*}

\begin{table*}[h!]
\fontsize{9.0pt}{\baselineskip}\selectfont
\linespread{0.9}\selectfont
\textbf{Case-1: Empty list} \\ 
\begin{mybody}
\textbf{Penalty List:} []
\end{mybody}
\begin{mybody}
\textbf{New Response:} The office supplies are contained in small teal packages or a teal envelope. \cmark

\textbf{New Colors:} ['teal'] \cmark
\end{mybody}
%%%%%%%%%%%%%%%%%%%%%%%%%%%%%%%%%%%%%%%%%%%%%%%%%%%%%%%%%
\end{table*}

\begin{table*}[h!]
\fontsize{9.0pt}{\baselineskip}\selectfont
\linespread{0.9}\selectfont
\textbf{Case-2a: First Attempt} \\ 
\begin{mybody}
\textbf{Penalty List:} ['purple', 'yellow', 'white', 'orange', 'green', 'brown', 'silver', 'red', 'pink', 'maroon']
\end{mybody}
\begin{mybody}
\textbf{New Response:} The office supplies are contained in small teal packages or a teal envelope. \cmark 

\textbf{New Colors:} ['teal'] \cmark
\end{mybody}
\begin{mybody}
\textbf{New Response:} The office supplies are contained in small white packages or a white envelope.  \xmark

\textbf{New Colors:} ['white'] \xmark 

\textbf{Reason:} Color \textit{white} conflicts with penalty list, retry.
\end{mybody}
%%%%%%%%%%%%%%%%%%%%%%%%%%%%%%%%%%%%%%%%%%%%%%%%%%%%%%%%%
\end{table*}

\begin{table*}[h!]
\fontsize{9.0pt}{\baselineskip}\selectfont
\linespread{0.9}\selectfont
\textbf{Case-2b: Second Attempt} \\ 
\begin{mybody}
\textbf{Penalty List:} ['purple', 'yellow', 'white', 'orange', 'green', 'brown', 'silver', 'red', 'pink', 'maroon']
\end{mybody}
\begin{mybody}
\textbf{New Response:} The office supplies are contained in small teal packages or a teal envelope. \cmark

\textbf{New Colors:} ['teal'] \cmark
\end{mybody}
\begin{mybody}
\textbf{New Response:} The office supplies are contained in small orange packages or a transparent envelope.  \xmark

\textbf{New Colors:} ['orange', 'transparent'] \xmark 

\textbf{Reason:} Color \textit{orange} conflicts with penalty list, even when transparent is reasonable and has no conflict.  Put sample back to un-annotated pool and rerun the script with un-annotated samples.
\end{mybody}
%%%%%%%%%%%%%%%%%%%%%%%%%%%%%%%%%%%%%%%%%%%%%%%%%%%%%%%%%
\end{table*}

\begin{table*}[h!]
\fontsize{9.0pt}{\baselineskip}\selectfont
\linespread{0.9}\selectfont
\textbf{Case-2c: Rerun-script (penalty list updates} \\ 
\begin{mybody}
\textbf{Penalty List:} ['yellow', 'gold', 'blue', 'black', 'brown', 'silver', 'red', 'pink', 'maroon']
\end{mybody}
\begin{mybody}
\textbf{New Response:} The office supplies are contained in small teal packages or a teal envelope. \cmark

\textbf{New Colors:} ['teal'] \cmark
\end{mybody}
\begin{mybody}
\textbf{New Response:} The office supplies are contained in small silver packages or a silver envelope. \xmark

\textbf{New Colors:} ['silver'] \xmark 

\textbf{Reason:} Color \textit{silver} conflicts with penalty list, go to Case-2b.
\end{mybody}
%%%%%%%%%%%%%%%%%%%%%%%%%%%%%%%%%%%%%%%%%%%%%%%%%%%%%%%%%
\end{table*}

\clearpage 
\subsection{Fine-grained length \& Stylistic analysis}
\label{sec:fine_grained_data_cmp_appendix}

\begin{table}[t!]
\centering
\caption{\small{Comparison of the \name Preference Dataset with Related Works: We compare \name to HA-DPO~\citep{ha-dpo}, POVID~\citep{povid}, RLAIF-V~\citep{rlaifv}, and CSR~\citep{csr} based on stylistic and length analysis, using linguistic similarity via word-level Levenshtein distance (LD) and token-level sequence length differences, respectively. We report the percentage of samples where chosen responses are longer \textcolor{darkgreen}{$(chosen>rejected)$} or shorter \textcolor{darkred}{$(rejected>chosen)$} for each overall, short (< 100 tokens), and long response category, with corresponding token differences following the same color notation. A lower LD stands for higher stylistic similarity, and lower token-level sequence length differences stand for higher length similarity. Note CSR only had long responses.}}
\label{tab:preference_dataset_cmp_appendix}
\setlength{\tabcolsep}{0.8mm}
\resizebox{\linewidth}{!}{
\begin{tabular}{lccccc}
\hline
 & \textbf{Ours} & HA-DPO & POVID & RLAIF-V & CSR \\\hline
Overall Samples (\%) & (\textcolor{darkgreen}{21}, \textcolor{darkred}{79}) & (\textcolor{darkgreen}{41}, \textcolor{darkred}{59}) & (\textcolor{darkgreen}{24}, \textcolor{darkred}{76}) & (\textcolor{darkgreen}{45}, \textcolor{darkred}{55}) & (\textcolor{darkgreen}{38}, \textcolor{darkred}{62}) \\
- Linguistic Similarity & 6 & 49 & 30 & 62 & 97 \\
- Avg. Token Length Difference & 3 & 18 & 16 & 15 & 27 \\
- Token Length Difference & (\textcolor{darkgreen}{10}, \textcolor{darkred}{1})& (\textcolor{darkgreen}{15}, \textcolor{darkred}{20}) & (\textcolor{darkgreen}{27}, \textcolor{darkred}{13}) & (\textcolor{darkgreen}{17}, \textcolor{darkred}{14}) & (\textcolor{darkgreen}{23}, \textcolor{darkred}{29})\\\\
Short Response Samples (\%) & (\textcolor{darkgreen}{19}, \textcolor{darkred}{81}) & (\textcolor{darkgreen}{49}, \textcolor{darkred}{51}) & (\textcolor{darkgreen}{10}, \textcolor{darkred}{90}) & (\textcolor{darkgreen}{43}, \textcolor{darkred}{57}) & - \\
- Linguistic Similarity & 3 & 24 & 16 & 17 & - \\
- Avg. Token Length Difference & 4 & 16 & 10 & 10 & - \\
- Token Length Difference & (\textcolor{darkgreen}{7}, \textcolor{darkred}{1}) & (\textcolor{darkgreen}{14}, \textcolor{darkred}{18}) & (\textcolor{darkgreen}{11}, \textcolor{darkred}{10}) & (\textcolor{darkgreen}{14}, \textcolor{darkred}{7}) & - \\\\
Long Response Samples (\%)  & (\textcolor{darkgreen}{33}, \textcolor{darkred}{67}) & (\textcolor{darkgreen}{35}, \textcolor{darkred}{65}) & (\textcolor{darkgreen}{46}, \textcolor{darkred}{54}) & (\textcolor{darkgreen}{45}, \textcolor{darkred}{55}) & (\textcolor{darkgreen}{38}, \textcolor{darkred}{62})\\
- Linguistic Similarity & 19 & 65 & 53 & 76 & 97 \\
- Avg. Token Length Difference & 8 & 20 & 27 & 18 & 27 \\
- Token Length Difference & (\textcolor{darkgreen}{17}, \textcolor{darkred}{4}) & (\textcolor{darkgreen}{19}, \textcolor{darkred}{21}) & (\textcolor{darkgreen}{32}, \textcolor{darkred}{23}) & (\textcolor{darkgreen}{19}, \textcolor{darkred}{17}) & (\textcolor{darkgreen}{23}, \textcolor{darkred}{29}) \\
\hline
\end{tabular}
}
\end{table}
We analyze stylistic and length similarity between chosen and rejected responses, both overall and split by response length (see Table~\ref{tab:preference_dataset_cmp_appendix}). We define responses with <=100 tokens as short and >100 tokens as long. Notably, even for long responses - where other methods exhibit higher dissimilarity and greater length differences - \name maintains higher linguistic similarity and lower token-level differences through its targeted response editing approach. This highlights the hard-negative nature of \name rejections and the nuanced distinctions its models are trained to optimize for.

\clearpage

\subsection{Human Study}
\label{sec:datagen_human_study_appendix}
% \subsection{Stylistic \& Length Similarity Study}
% \label{ls_calrification_appendix}

To evaluate the dataset, we conducted a human annotation study with three annotators per sample, including the authors of this paper. The setup and process are detailed as follows:

\begin{itemize}
    \item \textbf{Sample Selection:}  
    \begin{itemize}
        \item A total of 500 samples were randomly stratified across ten task categories.
        \item To prevent bias from task repetition, consecutive samples were ensured to come from different task categories.
    \end{itemize}
    
    \item \textbf{Annotation Procedure:}  
    \begin{itemize}
        \item The study spanned two days, with 250 annotations collected per day.
        \item Each day consisted of four hours of annotation, divided into one-hour sessions, with a 30-minute break after each session to mitigate annotator fatigue.
        \item Annotators were presented with the following elements for each sample:
        \begin{itemize}
            \item An image.
            \item An instruction.
            \item A chosen response.
            \item A rejected response.
        \end{itemize}
        \item The task required a binary annotation to determine whether the rejected response qualified as a \textit{hard-negative} or was similar to the chosen response.
    \end{itemize}
    
    \item \textbf{Annotation Criteria:}  
    \begin{itemize}
        \item Annotators were instructed to evaluate the content of the responses exclusively.
        \item Factors such as response length or linguistic similarity were explicitly excluded, as these were analyzed in prior studies (see \S\ref{sec:dataset_analysis}) to avoid the confounding factors of human subjectivity in length and linguistic preference.
    \end{itemize}
    
    \item \textbf{Quality Assurance:}  
    \begin{itemize}
        \item Results indicated that 97\% of the samples aligned with the hard-negative response criteria.
        \item Inter-annotator agreement (IAA), calculated using Fleiss' kappa, was 86\%, signifying a high level of consistency among annotators.
    \end{itemize}
\end{itemize}

These findings demonstrate that the dataset achieves high quality and reliability, despite being synthetically generated.

\clearpage

\section{Experimental Setup}
\label{sec:setup_appendix}
% \subsection{Model Setup}
% \label{sec:model_setup_appendix}
% TBA
\subsection{Training Setup}
\label{sec:training_setup_appendix}
This section outlines the training setup and hyperparameters, summarized in Table~\ref{tab:hyperparam_table}. Consistent hyperparameters were applied across all \name models, including LLaVA-\name (7B and 13B), Qwen2VL-\name (2B and 7B) and Qwen2.5VL-\name (3B and 7B). A shallow hyperparameter search was conducted for the learning rate, with 1e-6 yielding optimal results; 1e-5 caused model forgetting, while 1e-7 was insufficient for effective learning. An effective batch size of 32 was selected for training efficiency. The models were trained for 5 epochs using two A100 GPUs.
\begin{table}[h!]
\centering
\small
\begin{tabular}{|l|l|}
\hline
\textbf{Hyperparameter} & \textbf{Value} \\ \hline
Learning rate & 1e-6  \\ \hline
Learning rate Scheduler & Cosine \\ \hline
Warmup Ratio & 0.03 (LLaVA) \& 0.1 (Qwen2VL \& 2.5VL)  \\ \hline
Batch size & 32  \\ \hline
Lora r & 128  \\ \hline
Lora alpha & 256  \\ \hline
DPO Loss & sigmoid  \\ \hline
DPO $\beta$ & 0.1  \\ \hline
Max Sequence Length & 2048  \\ \hline
\end{tabular}
\caption{Overview of training hyperparameters}
\label{tab:hyperparam_table}
\end{table}

\subsection{Evaluation Benchmarks \& Setup}
\label{sec:evaluation_setup_appendix}

\subsubsection{Benchmarks}
\paragraph{Open-ended \& Descriptive Benchmarks}: This includes LLaVA\textsuperscript{W} (LLaVA-in-the-wild)~\citep{llava-v1}, ConTextual (ConT)~\citep{contextual}, \& MM-VET (MMV)~\citep{mm-vet}. LLaVA\textsuperscript{W} assesses open-world visual reasoning and description. ConTextual tests joint reasoning over embedded text and visual elements across diverse text-rich image scenarios. MM-VET evaluates how well LVLMs perform on tasks integrating core-VL capabilities like OCR, spatial reasoning, and math. We report the GPT-4 scores obtained using the respective LLM-as-a-judge prompts of LLaVA\textsuperscript{W}, ConTextual, and MM-VET.

\paragraph{Vision-Centric Benchmarks}: This includes SEED\textsuperscript{I} (SEED Bench image split)~\citep{seedbench}, CV (CV Bench)~\citep{cambrian} and MMStar (MMS)~\citep{mmstar}. SEED \& MMS comprehensively evaluate perception and reasoning. CV assesses counting, spatial reasoning, and comparative reasoning (depth \& distance). We report the overall accuracy for each benchmark.

\paragraph{Academic \& Math Reasoning:} This category includes MathVista (MV)~\citep{mathvista}, testmini subset, which evaluates mathematical reasoning across diverse problem types, and MMMU~\citep{mmmu}, which tests college-level academic reasoning across domains such as physical sciences, social sciences, finance, etc. For MMMU and MathVista we report the overall accuracy.

\paragraph{Hallucination \& Adversarial}: This category includes POPE~\citep{pope} and NaturalBench~\citep{naturalbench}. POPE assesses object hallucination by testing scenarios of object presence and absence. NaturalBench measures visio-linguistic compositionality and vision-blind behavior (bias to provide identical answers regardless of the image) of LVLMs by pairing two questions with two similar images yielding different answers. For POPE, we report the overall F1 score across three categories. For NaturalBench, we provide overall accuracy, per-image accuracy across two questions, per-question accuracy across paired images, and group accuracy, reflecting the model's ability to answer all four image-question combinations correctly.

\subsubsection{Evaluation Setup} 
\paragraph{Model Generation Settings}: To ensure consistency and equitable comparison with prior works \citep{llava-v15, povid, csr, ha-dpo, sima, llava-rlhf, rlaifv}, we set the generation temperature to 0 and the number of beams to 1 for all base, prior work and \name models. 

\paragraph{LLM-as-a-Judge}: For benchmarks such as LLaVA-bench, ConTextual, and MM-Vet, which require OpenAI APIs for evaluation, we utilize GPT-4 to align with prior methodologies. It is important to note that GPT-4 versions evolve due to periodic updates. For instance, prior works may have used GPT-4-0314, which was deprecated in June 2024\footnote{\href{https://platform.openai.com/docs/deprecations}{GPT4 Depracation History}}. To ensure consistency, we fix the version to GPT-4-0613, the current stable release, and calculate scores for these benchmarks across all baseline and \name models. Further, MathVista utilizes GPT-4-Turbo calls for answer extraction. To minimize our costs, we instead used GPT-4o, which is a more performant OpenAI model that is significantly cheaper. This evaluation incurred a total cost of approximately $\$800$ (in addition to $\$300$ for data generation using GPT-4o).

\paragraph{Significance analysis}
For statistical significance analysis, we employ bootstrap resampling \citep{bootstrap_sstest}, which involves randomly sampling 50\% of each benchmark's data and evaluating model performance over 1,000 iterations for comparing two models. We assess each preference-tuned baseline and \name model against the base SFT model, reporting statistical significance when the win rate is $\geq$ 95\% (p = 0.05). Notably, \name models demonstrate improvements over base SFT models, with most results achieving statistical significance. In contrast, while showing improvements, prior works fail to meet the 95\% confidence threshold under the bootstrap test (except CSR-13B on the Pope dataset), as shown in Table~\ref{tab:main_results}.
\clearpage

\section{Extended Results}
\label{sec:detailed_benchmarks_results}
\subsection{Data Scaling}
\label{sec:data_scale_appendix}

\begin{table*}[t]
\caption{
Performance comparison of LLaVA-v1.5-Instruct, Qwen2VL-Instruct, Qwen2.5VL-Instruct, and DPO models finetuned on \name at three data scales 3K, 10K 30K, on ten benchmarks. Higher scores indicate better performance across all benchmarks, with the highest score for each benchmark highlighted in \textbf{bold}. PFT size refers to the preference for fine-tuning dataset
size, where "-" represents no additional dataset. Each model is trained under identical hyperparameter settings.
}
\label{tab:scaling_results}
\centering
\setlength{\tabcolsep}{4pt}
\scriptsize
\begin{tabular}{llccccccccccccc}
\toprule
\textsc{Row} & \textsc{Method} & \textsc{PFT} & \textsc{LLaVA\textsuperscript{W}} & \textsc{ConT} & \textsc{MMV} & \textsc{SEED\textsuperscript{I}} & \textsc{CV} & \textsc{MV} & \textsc{MMMU} & \textsc{MMS} & \textsc{POPE} & \textsc{NB}
 \\
\midrule 
1 & LLaVA-1.5-7B & - & 64.8 & 16.8 & 30.9 & 66.2 & 62.1 & 30.1 & 35.4 & 32.6 & 85.9 & 12.7  \\
2 & + \name DPO & 3K & 69.9 & 19.2 & 31.4 & 66.1 & 61.9 &  30.1 & 35.4 & 32.9 & 85.1 & 13.6 \\
3 & + \name DPO & 10K & 74.4 & 20.2 & 32.3 & 66.4 & 62.3 & 30.4 & 35.6 & 34.0 & 85.2 & 14.0  \\
4 & + \name DPO & 30K & \textbf{76.2} & \textbf{20.6} & \textbf{32.9} & \textbf{66.7} & \textbf{62.9} & \textbf{30.8} &  \textbf{35.7} & \textbf{34.7} & 85.4 & \textbf{14.5}\\
\midrule
5 & LLaVA-1.5-13B & 30.7 & 72.3 & 18.6 & 36.7 & 68.2 & 62.5  & 30.7 & \textbf{36.1} & 33.8 & 86.0 & 14.9\\
6 & + \name DPO & 3K & 75.6 & 19.4 & 36.3 & 68.3 & 63.5 & 31.1 & 35.2 & 34.9 & 86.0 & 15.8 \\
7 & + \name DPO & 10K & 78.9 & 20.3 & 37.0 & 68.4 & 64.2  & 31.8 & 35.6 & 35.3 & 86.2 & 17.4 \\
8 & + \name DPO & 30K & \textbf{80.5} & \textbf{21.2} & \textbf{37.3} & \textbf{68.7} & \textbf{64.6} & \textbf{32.3}  & 35.8 & \textbf{35.6} & 86.3 & \textbf{18.2}  \\

\midrule
9 & Qwen2VL-2B & - & 83.2 & 27.7 & 53.3 & 73.6 & 66.5 &  \textbf{51.0} & 38.7 & 43.4  & 86.5 & 24.3\\
10 & + \name DPO & 3K & 83.8 & 31.6 & 52.6 & 73.6 & 67.5 & 50.2 & 38.7 & 43.4 & 87.6 & 24.9 \\
11 & + \name DPO & 10K & 84.3 & 33.2 & 53.4 & 73.8 & 68.3 & 50.5 & 39.0 & 43.5 & 88.2 & 25.2\\
12 & + \name DPO & 30K & \textbf{88.1} & \textbf{34.8} & \textbf{54.1} & \textbf{74.0} & \textbf{69.0} & 50.9 & \textbf{39.2} & \textbf{43.7} & \textbf{88.3} & \textbf{25.7} \\

\midrule
13 & Qwen2VL-7B & - & 92.5 & 39.7 & 62.1 & 76.4 & 75.7 & 57.5 & \textbf{50.7} & 56.7 & \textbf{87.3} & 30.8\\
14 & + \name DPO & 3K & 92.8 & 41.9 & 63.0 & 76.4 & 75.8 & 57.2 & 50.6 & 57.1 & 87.0 & 31.7 \\
15 & + \name DPO & 10K & 93.6 & 42.3 & 64.4 & 76.5 & 76.0 & 57.4 & 50.5 & 57.5 & 87.2 & 32.0 \\
16 & + \name DPO & 30K & \textbf{96.2} & \textbf{43.9} & \textbf{65.4} & \textbf{76.8} & \textbf{76.3} & \textbf{58.2}  & 50.0 & \textbf{57.8} & \textbf{87.3} & \textbf{32.5}  \\

\midrule
17 & Qwen2.5VL-3B & - & 98.1 & 37.2 & 67.3 & 75.0 & 71.5  & 52.5  & \textbf{45.7} & 54.7 & 86.3 & 25.4  \\
18 & + \name DPO & 3K & 95.1 & 38.0 & 66.8 & 75.0 & 71.6 & 52.5 & 45.3 & 55.0  & 86.0 & 25.5  \\
19 & + \name DPO & 10K & 96.5 & 39.3 & 66.9 & 75.3 & 72.0 & 52.7 &  45.1 & 55.6 & 86.1 & 25.7  \\
20 & + \name DPO & 30K & 97.1 & \textbf{40.3} & \textbf{67.4} & \textbf{75.5} & \textbf{72.7}  & \textbf{53.4} & 44.9  &  \textbf{56.1} & \textbf{86.4} & \textbf{26.3}  \\

\midrule
21 & Qwen2.5VL-7B & - & 101.4 & 53.3 & 71.0 & 77.7 & 80.1 & 58.6  & \textbf{50.9} & 61.9 & 86.3 & 32.0  \\
22 & + \name DPO & 3K & 100.6 & 53.1 & 71.2 & 77.6 & 80.5 & 58.6 & \textbf{50.9} & 62.0  & 86.4 & 32.0 \\
23 & + \name DPO & 10K & 100.8 & 53.2 & 71.8 & 77.7 & 80.7 & 58.8  & 50.8 & 62.2 & 86.7 & 32.2  \\
24 & + \name DPO & 30K & \textbf{101.5} & \textbf{53.4} & \textbf{72.4} & \textbf{77.8} & \textbf{81.1} & \textbf{59.8}  & 50.6 & \textbf{62.5} & \textbf{86.9} & \textbf{32.8}  \\

\bottomrule
\end{tabular}
\end{table*}

\clearpage

\subsection{Benchmark detailed results}
\label{sec:benchmark_detailed_results}
We provide the breakdown of model performance across different task categories for two comprehensive dataset MMStar and CV Bench. 

\begin{table}[ht]
\centering
\caption{
\textbf{MMStar}: Performance is compared across LLaVA-v1.5-Instruct, Qwen2VL-Instruct, Qwen2.5VL-Instruct, and DPO models - preference finetuned on \name (30K subset created with GPT-4o). Higher scores indicate better performance, with the top result for each benchmark shown in bold. All models share the same hyperparameters.
Abbreviations: CP = Coarse Perception, FGP = Fine-grained Perception, IR = Instance Reasoning, LR = Logical Reasoning, S\&T = Science \& Technology.
}
\label{tab:mmstar_results}
\begin{tabular}{@{}l *{7}{c}@{}}
\toprule
Model & Final Score & CP & FGP & IR & LR & S\&T & Math \\
\midrule
LLaVA-v1.5-Instruct-7B & 32.6 & 58.8 & 26.8 & 40.0 & 26.0 & 17.2 & 26.8 \\
VaPR-LLaVA-7B & \textbf{34.7} & \textbf{62.0} & \textbf{27.2} & \textbf{44.0} & \textbf{27.2} & \textbf{18.0} & \textbf{30.0} \\
\midrule 
LLaVA-v1.5-Instruct-13B & 33.8 & 58.0 & 27.2 & 42.4 & 26.4 & 21.2 & \textbf{27.6} \\
VaPR-LLaVA-13B & \textbf{35.6} & \textbf{60.8} & \textbf{28.4} & \textbf{47.6} & \textbf{28.0} & \textbf{23.6} & 25.2 \\
\midrule 
Qwen2VL-Instruct-2B & 43.4 & 52.4 & 41.6 & \textbf{51.6} & \textbf{43.2} & 31.2 & \textbf{40.4} \\
VaPR-Qwen2VL-2B & \textbf{43.7} & \textbf{53.6} & \textbf{45.6} & 50.0 & 42.0 & \textbf{31.6} & 39.6 \\
\midrule 
Qwen2VL-Instruct-7B & 56.7 & 67.2 & 50.4 & 62.8 & 56.4 & 46.4 & 57.2 \\
VaPR-Qwen2VL-7B & \textbf{57.8} & \textbf{67.6} & \textbf{51.2} & \textbf{63.2} & \textbf{57.6} & \textbf{49.2} & \textbf{58.0} \\
\midrule 
Qwen2.5VL-Instruct-3B & 54.7 & 66.4 & 46.4 & 60.8 & 54.8 & \textbf{39.6} & 60.4 \\
VaPR-Qwen2.5VL-3B & \textbf{56.1} & \textbf{68.4} & \textbf{47.2} & \textbf{63.6} & \textbf{55.2} & \textbf{39.6} & \textbf{62.4} \\
\midrule 
Qwen2.5VL-Instruct-7B & 61.9 & \textbf{72.0} & 54.0 & \textbf{70.8} & 63.2 & 44.8 & 66.4 \\
VaPR-Qwen2.5VL-7B & \textbf{62.5} & 71.6 & \textbf{55.2} & 70.0 & \textbf{64.4} & \textbf{45.6} & \textbf{68.4} \\
\bottomrule
\end{tabular}
\end{table}

\begin{table}[ht]
\centering
\caption{
\textbf{CV Bench}: Performance is compared across LLaVA-v1.5-Instruct, Qwen2VL-Instruct, Qwen2.5VL-Instruct, and DPO models - preference finetuned on \name (30K subset created with GPT-4o). Higher scores indicate better performance, with the top result for each benchmark shown in bold. All models share the same hyperparameters. 
Abbreviations:  Overall = Overall Accuracy, Count = Count Accuracy, Spatial = Spatial Relation Accuracy, Depth = Depth (Order) Accuracy, Distance = Relative Distance Accuracy. Depth: Determine which of the two distinct objects is closer to the camera, and Relative Distance: Determine which of the two distinct objects is closer to the anchor object.
}
\label{tab:cvbench_results}

\begin{tabular}{@{}l *{5}{c}@{}}
\toprule
Model & Overall & Count & Spatial & Depth & Distance \\
\midrule
LLaVA-v1.5-Instruct-7B & 62.2 & 54.3 & \textbf{71.2} & 70.0 & \textbf{56.3} \\
VaPR-LLaVA-7B          & \textbf{62.9} & \textbf{57.2} & 70.8 & \textbf{71.7} & 54.3 \\
\midrule 
LLaVA-v1.5-Instruct-13B & 62.5 & 58.8 & 68.2 & 69.7 & 55.8 \\
VaPR-LLaVA-13B         & \textbf{64.6} & \textbf{58.9} & \textbf{69.9} & \textbf{72.7} & \textbf{59.7} \\
\midrule 
Qwen2VL-Instruct-2B    & 66.5 & 66.6 & 67.5 & 65.8 & 67.5 \\
VaPR-Qwen2VL-2B        & \textbf{69.0} & \textbf{68.2} & \textbf{68.3} & \textbf{68.5} & \textbf{72.2} \\
\midrule 
Qwen2VL-Instruct-7B    & 75.7 & \textbf{67.0} & 79.5 & \textbf{85.5} & 72.8 \\
VaPR-Qwen2VL-7B        & \textbf{76.3} & 66.6 & \textbf{81.9} & 82.7 & \textbf{76.3} \\
\midrule 
Qwen2.5VL-Instruct-3B  & 71.5 & 68.5 & 74.8 & 78.2 & 66.3 \\
VaPR-Qwen2.5VL-3B      & \textbf{72.7} & \textbf{68.9} & \textbf{75.1} & \textbf{79.3} & \textbf{69.2} \\
\midrule 
Qwen2.5VL-Instruct-7B  & 80.1 & 68.5 & 90.0 & 86.7 & 78.5 \\
VaPR-Qwen2.5VL-7B      & \textbf{81.1} & \textbf{69.2} & \textbf{90.6} & \textbf{86.8} & \textbf{81.2} \\
\bottomrule
\end{tabular}
\end{table}

From Tables \ref{tab:main_results},\ref{tab:mmstar_results} \& \ref{tab:cvbench_results}, we observe that VaPR models consistently improve on perception tasks  (particularly fine-grained perception), and reasoning tasks, such as spatial relationships (even complex ones like distance and depth) and counting. These gains align with their improved performance on SeedBench and NaturalBench, both of which emphasize visio-linguistic compositionality, perception, and reasoning.

Notably, despite not being explicitly trained on OCR, textual reasoning, or math tasks, VaPR models achieve strong performance in these areas. We attribute this to their enhanced fine-grained perception, spatial reasoning, and counting capabilities, which demonstrates that improvements in these areas support interpretation of embedded text (consistent with prior work \citep{ocrbench2}) and geometric figures. This trend is further corroborated by gains on ConTextual and MathVista benchmarks. Interestingly, VaPR-Qwen2VL-2B achieves the largest gains on Pope, which can be explained by pronounced improvements in fine-grained perception (as evident in MMStar). On the other hand, it shows slight degradation in math, logical reasoning tasks, which can explain why it does not improve on MathVista, where the other models do (see Table~\ref{tab:main_results}).

Lastly, prior work \citep{unpacking} indicates that preference optimization primarily enhances truthfulness and alignment, rather than factuality, which aligns with our observed improvements in perception and reasoning but not in purely knowledge-based tasks. We hypothesize that limited gains in MMMU (see Table~\ref{tab:main_results}) can be attributed to the alignment tax VaPR models possibly pay in knowledge based tasks.

\clearpage

\subsection{Preference Dataset Comparison}
\label{sec:dataset_cmp_appendix}

\subsubsection{Training Setup}

We compare \name with two alternative preference data generation techniques: (1) \textbf{POVID}, which uses GPT-4V to generate rejected responses given the image, instruction, and ground truth response, and (2) \textbf{SIMA}, which follows a self-preference generation paradigm by synthesizing two responses - one via greedy decoding and another via sampling (temperature = 1) - and using the same VLM as a critic to select the preferred response. For SIMA, we apply the generation method to the 10K subset of \name dataset; for POVID, we use the publicly released data (17K). 
For \name we select the 10K subset for training to ensure a fair comparison. We preference finetune LLaVA-v1.5 and Qwen2VL models on the above datasets. To ensure consistency, all models are trained for 50K steps, under identical hyperparameter settings. This setup enables a balanced and computationally efficient comparison, allowing us to evaluate all methods under a fixed training budget without an excessive number of experiments.

\begin{table*}[htbp]
\caption{
Performance comparison of LLaVA-v1.5, Qwen2VL-Instruct, DPO models finetuned on \name, SIMA \& POVID across 2B, 7B, and 13B parameter sizes on ten benchmarks. Higher scores indicate better performance across all benchmarks, with the highest score for each benchmark highlighted in \textbf{bold}. PFT size refers to the preference for fine-tuning dataset size, where \textbf{"-"} represents no additional dataset. 
}
\label{tab:preference_dataset_cmp_results}
\centering
\setlength{\tabcolsep}{4pt}
\scriptsize
\begin{tabular}{llccccccccccccc}
\toprule
\textsc{Row} & \textsc{Method} & \textsc{PFT} & \textsc{LLaVA\textsuperscript{W}} & \textsc{ConT} & \textsc{MMV} & \textsc{SEED\textsuperscript{I}} & \textsc{CV} & \textsc{MV} & \textsc{MMMU} & \textsc{MMS} & \textsc{POPE} & \textsc{NB}
 \\
\midrule 
1 & LLaVA-1.5-7B & - & 64.8 & 16.8 & 30.9 & 66.2 & 62.1 & 30.1 & 35.4 & 32.6 & \textbf{85.9} & 12.7  \\
2 & + \name-SIMA + DPO & 10K & 
68.5 & 17.2 & 31.6 & 66.0 & 60.9 &  29.4 & 35.2 & 32.7 & 85.2 & 12.9 \\
3 & + Povid DPO & 17K & 67.2 & 18.0 & 30.9 & 66.1 & 61.6 & 30.1 & 35.4 & 33.3 & \textbf{85.9} & 13.2 \\
4 & + \name DPO & 10K & \textbf{74.4} & \textbf{20.2} & \textbf{32.3} & \textbf{66.4} & \textbf{62.3} & \textbf{30.4} & \textbf{35.6} & \textbf{34.0} & 85.2 & \textbf{14.0}  \\
\midrule
5 & LLaVA-1.5-13B & - & 72.3 & 18.6 & 36.7 & 68.2 & 62.5  & 30.7 & \textbf{36.1} & 33.8 & 86.0 & 14.9\\
6 & + \name-SIMA + DPO & 10K & 73.3 & 18.2 & 35.5 & 67.7 & 61.0 & 31.2  & 34.5 & 33.6 & 86.0 & 13.1 \\
7 & + Povid DPO & 17K & 74.5 & 19.8 & 34.6 & 68.1 & 63.9 & 31.5  & 34.9 & 34.1 & \textbf{86.2} & 15.3 \\
8 & + \name DPO & 10K & \textbf{78.9} & \textbf{20.3} & \textbf{37.0} & \textbf{68.4} & \textbf{64.2}  & \textbf{31.8} & 35.6 & \textbf{35.3} & \textbf{86.2} & \textbf{17.4} \\

\midrule
9 & Qwen2VL-2B & - & 83.2 & 27.7 & 53.3 & 73.6 & 66.5 &  \textbf{51.0} & 38.7 & 43.4  & 86.5 & 24.3\\
10 & + \name-SIMA + DPO & 10K & 81.6 & 29.2 & 49.1 & 73.4 & 66.7 & 50.0  & 38.8 & 43.1 & 86.8 & 23.4 \\
11 & + Povid DPO & 17K & 82.7 & 30.7 & 49.3 & 73.7 & 67.1 & 50.2 & 38.9 & 43.2 & 87.1 & 23.6 \\
12 & + \name DPO & 10K & \textbf{84.3} & \textbf{33.2} & \textbf{53.4} & \textbf{73.8} & \textbf{68.3} & 50.5 & \textbf{39.0} & \textbf{43.5} & \textbf{88.2} & \textbf{25.2}\\

\midrule
13 & Qwen2VL-7B & - & 92.5 & 39.7 & 62.1 & 76.4 & 75.7 & \textbf{57.5} & \textbf{50.7} & 56.7 & 87.3 & 30.8\\
14 & + \name-SIMA + DPO & 10K & 90.1 & 38.5 & 62.9 & 76.1 & 75.4 & 56.5 & 50.0 & 57.0  & 87.2 & 30.5  \\
15 & + Povid DPO & 17K & 90.9 & 37.4 & 63.5 & 76.2 & 75.8 & 57.0 & 50.4 & 57.2 & \textbf{87.6} & 31.1 \\
16 & + \name DPO & 10K & \textbf{93.6} & \textbf{42.3} & \textbf{64.4} & \textbf{76.5} & \textbf{76.0} & 57.4 & 50.5 & \textbf{57.5} & 87.2 & \textbf{32.0} \\

\bottomrule
\end{tabular}
\end{table*}

\subsubsection{Analysis}
From Table~\ref{tab:preference_dataset_cmp_results}, we observe that \name models consistently outperform both SIMA and POVID across model families. SIMA yields minimal to no improvements over base models and often degrades performance, especially for Qwen2VL. POVID achieves moderate gains for LLaVA but underperforms on Qwen2VL. On average, \name outperforms SIMA by 5-6\% on LLaVA and 3-4\% on Qwen2VL, and POVID by 4-5\% and 2-3\%, respectively. To understand these differences, we analyze DPO optimization using LLaVA-1.5-7B as a representative model. 

The DPO loss can be re-written as:

\begin{multline}
\mathcal{L}_{\text{DPO}} = -\log \sigma \left( \alpha \left[ \log \pi_\theta(y_w \mid x) - \log \pi_\theta(y_l \mid x) - \left( \log \pi_{\text{ref}}(y_w \mid x) - \log \pi_{\text{ref}}(y_l \mid x) \right) \right] \right),
\end{multline}
where \( \pi_\theta \) is the learned policy, \( \pi_{\text{ref}} \) is the fixed reference model, and \( \alpha \) is the temperature scaling factor. Let:
\[
\begin{aligned}
\Delta_\theta &= \log \pi_\theta(y_w \mid x) - \log \pi_\theta(y_l \mid x), \\
\Delta_{\text{ref}} &= \log \pi_{\text{ref}}(y_w \mid x) - \log \pi_{\text{ref}}(y_l \mid x).
\end{aligned}
\]

the loss simplifies to: \(
\mathcal{L}_{\text{DPO}} = -\log \sigma\left( \alpha (\Delta_\theta - \Delta_{\text{ref}}) \right). \)

\begin{figure}[t]
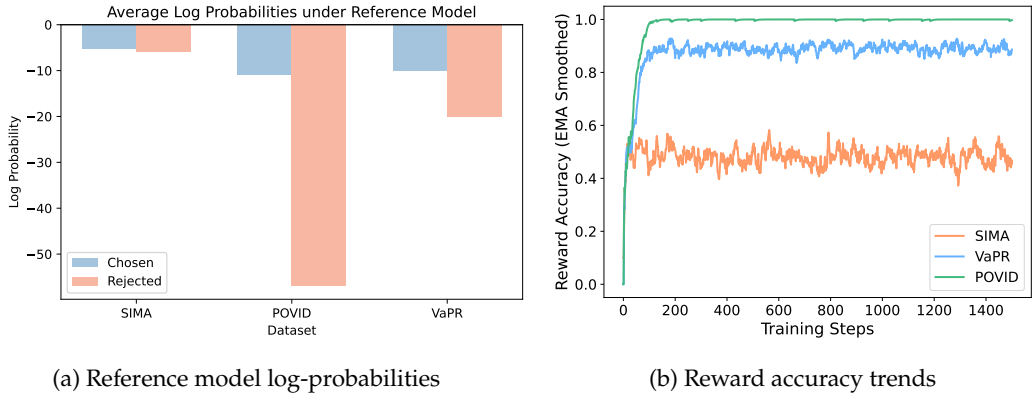

  \centering
  \begin{subfigure}[c]{0.48\textwidth}
    \centering
    \includegraphics[height=4.8cm]{images/ref_logp_comparison.pdf}
    \caption{Reference model log-probabilities}
    \label{fig:ref_logp_appendix}
  \end{subfigure}
  \hfill
  \begin{subfigure}[c]{0.48\textwidth}
    \centering
    \includegraphics[height=4.8cm]{images/reward_accuracy_plot.pdf}
    \caption{Reward accuracy trends}
    \label{fig:reward_acc_appendix}
  \end{subfigure}
  \caption{\small Comparison of preference datasets. (a) Average reference model log-probabilities for chosen vs. rejected responses across \name, SIMA, and POVID - lower values indicate lower reference likelihood. (b) Reward accuracy trends over training steps show that SIMA improves gradually while POVID saturates quickly. Repeating the figure for ease of reference}
  \label{fig:logp_and_reward_appendix}
\end{figure}

For POVID, we observe that \( \Delta_{\text{ref}} \) is already higher than \name while having a similar log probability of chosen response under the reference model (see Fig.~\ref{fig:ref_logp_appendix}), indicating that POVID rejected responses are substantially less likely under the reference model as compared \name's. This can be attributed to the higher linguistic and stylistic differences in POVID preference pairs as compared to \name (see Table~\ref{tab:preference_dataset_cmp}). From Fig.~\ref{fig:reward_acc_appendix}, we observe that the POVID model rapidly attains a reward accuracy of 1, suggesting that the model may be overfitting based on preference signals derived from length and stylistic differences instead of content differences alone. In contrast, \name achieves its highest reward accuracy more gradually and does not converge to 1, indicating reduced overfitting due to exposure to more challenging preference pairs.

In contrast, for SIMA, we observe \( \Delta_{\text{ref}} \approx 0 \) on average (see Fig.~\ref{fig:ref_logp_appendix}), indicating that chosen and rejected responses are often nearly identical. Manual inspection confirms this, with \(\sim20\%\) of pairs being exact duplicates and many others highly similar. When \( \Delta_{\text{ref}} \approx 0 \), the DPO loss is driven entirely by \( \Delta_\theta \), removing the reference model’s regularizing influence. This causes the optimizer to treat weak or noisy preference signals as informative, leading to overconfident updates based on superficial differences. Consequently, the model struggles to distinguish between responses - reflected in ~50\% reward accuracy (see Fig.~\ref{fig:reward_acc_appendix}), potentially learning undesirable behaviors that harm downstream performance. Other self-preference methods, such as RLAIF-V and CSR, which employ multi-step generation and scoring procedures and generate preference pairs with greater stylistic and linguistic variation (see Table~\ref{tab:preference_dataset_cmp}), can mitigate the above issue, but similar to POVID, they may also inadvertently exploit the stylistic and length biases.

On the other hand, \name explicitly controls for stylistic and length similarity when constructing preference pairs, ensuring that DPO learns from content-level differences. However, in low-resource settings where high-quality SFT data may be limited (as required by \name), our method can complement self-preference generation approaches. For instance, one could leverage high-confidence responses from methods like CSR or RLAIF-V as chosen responses, and apply the VaPR pipeline to generate targeted hard negative - creating high-quality preference pairs in an unsupervised manner. We view this as a promising direction for future work and invite the community to explore such hybrid strategies.

\clearpage
\subsection{Open-Source Editor data results}
\label{sec:os_analysis_appendix}
\begin{table*}[t]
\caption{
Performance is compared across LLaVA-v1.5-Instruct, Qwen2VL-Instruct, Qwen2.5VL-Instruct, and DPO models, all preference finetuned on \name (10K subset created with GPT-4o) or \name-OS (8K samples generated by the Open-source model Qwen3-32b). Aside from the model used for generation, the framework and prompts are identical. All models share the same hyperparameters. Higher scores indicate better performance, with the top result for each benchmark shown in \textbf{bold}. In case two models get the top scores, both are \textbf{bolded}, otherwise top score is \textbf{bolded} and the second highest score(s) is \underline{underlined}.
}
\label{tab:os_editor_results}
\centering
\setlength{\tabcolsep}{4pt}
\scriptsize
\begin{tabular}{llccccccccccccc}
\toprule
\textsc{Row} & \textsc{Method} & \textsc{LLaVA\textsuperscript{W}} & \textsc{ConT} & \textsc{MMV} & \textsc{SEED\textsuperscript{I}} & \textsc{CV} & \textsc{MV} & \textsc{MMMU} & \textsc{MMS} & \textsc{POPE} & \textsc{NB}
 \\
\midrule 
1 & LLaVA-1.5-7B-Instruct & 64.8 & 16.8 & 30.9 & 66.2 & 62.1 & 30.1 & 35.4 & 32.6 & \textbf{85.9} & 12.7  \\
2 & + \name-OS DPO & \underline{73.3} & \underline{18.7} & \underline{32.1} & \underline{66.3} & \textbf{62.3} &  \underline{30.2} & \textbf{35.6} & \underline{33.7} & 83.6 & \underline{13.9} \\
3 & + \name DPO & \textbf{74.4} & \textbf{20.2} & \textbf{32.3} & \textbf{66.4} & \textbf{62.3} & \textbf{30.4} & \textbf{35.6} & \textbf{34.0} & \underline{85.2} & \textbf{14.0}  \\

\midrule
4 & Qwen2VL-2B-Instruct & 83.2 & 27.7 & \underline{53.3} & 73.6 & 66.5 &  \textbf{51.0} & 38.7 & 43.4  & 86.5 & 24.3\\
5 & + \name-OS DPO & \underline{84.1} & \underline{32.8} & \underline{53.3} & \underline{73.7} & \underline{67.9} & 50.2 & \underline{38.9} & \textbf{43.5} & \underline{88.0} & \textbf{25.2} \\
6 & + \name DPO & \textbf{84.3} & \textbf{33.2} & \textbf{53.4} & \textbf{73.8} & \textbf{68.3} & \underline{50.5} & \textbf{39.0} & \textbf{43.5} & \textbf{88.2} & \textbf{25.2}\\

\midrule
7 & Qwen2.5VL-3B-Instruct & \textbf{98.1} & 37.2 & \textbf{67.3} & 75.0 & 71.5  & \underline{52.5}  & \textbf{45.7} & 54.7 & \textbf{86.3} & 25.4  \\
8 & + \name-OS DPO & 95.7 & \underline{39.0} & 65.4 & \textbf{75.3} & \textbf{72.0} & 52.4 & \underline{45.5} & \underline{55.4}  & \underline{86.2} & \textbf{25.7}  \\
9 & + \name DPO & \underline{96.5} & \textbf{39.3} & \underline{66.9} & \textbf{75.3} & \textbf{72.0} & \textbf{52.7} &  45.1 & \textbf{55.6} & 86.1 & \textbf{25.7}  \\

\bottomrule
\end{tabular}
\end{table*}

We preference finetuned LLaVA-v1.5-Instruct-7B, Qwen2VL-Instruct-2B and Qwen2.5-VL-Instruct-3B on VaPR-OS (VaPR Open source), where the difference between the models are two aspects: (a) is missing reasoning and captioning samples (1K each), reason shared below (b) uses an open-weight model for generating responses. Key findings include:

\paragraph{Dataset Quality:} VaPR-OS rejected responses exhibit similar hard-negative properties to those generated by GPT-4o, with an average token length difference of 6 (compared to 3 in VaPR) and a Levenshtein distance of 10 (vs. 6 in VaPR).

\paragraph{Model Performance:}
From Table~\ref{tab:os_editor_results}, we observe that models trained on the open-weight dataset achieve ~99\% of the performance of those generated with GPT-4o, with both consistently outperforming the baseline on most benchmarks. This is expected, as VaPR-OS samples overlap with VaPR and exhibit similar linguistic properties and average token length. These results highlight that the VaPR pipeline generalizes effectively and is not limited to closed-weight models.

Interestingly, we find that the contribution of captioning and abstract reasoning tasks to preference learning is limited, which is consistent with prior work \citep{step-dpo} suggesting that complex tasks like reasoning may benefit stepwise decomposition of preference datasets. In our case, captioning and abstract reasoning tasks (e.g. Considering the presence of two clocks on the building, what purpose might this architectural design serve?), can be decomposed into simpler components like fine-grained perception (e.g., attribute recognition) and spatial reasoning (e.g., object location). Training models using these atomic tasks as step-wise preference samples may collectively support learning for more complex tasks, a direction we plan to investigate further. 

\paragraph{Limitations and Future Directions for OS editors:} We observed that data generated using Qwen3 sometimes fails to consistently perturb dependent spans (e.g., object perturbation: “... bathroom … has a large bathtub …” → “... kitchen … has a large bathtub …”, whereas GPT-4o correctly changes it to “... kitchen … has a large oven …”), with this issue being prominent in captioning and abstract reasoning tasks. This could potentially add noise to the dataset and thus we omit these samples in our analysis. To mitigate this issue, we plan to experiment with more open-source models and generation of step-wise preference datasets.

\end{document}